\documentclass[runningheads]{llncs}
\usepackage{graphicx}

\usepackage{tikz}
\usepackage{comment}
\usepackage{amsmath,amssymb} 
\usepackage{color}

\usepackage{graphicx}
\usepackage{bbm}
\usepackage{bm}
\usepackage{xspace}
\usepackage[utf8]{inputenc} 
\usepackage{url}            
\usepackage{booktabs}       
\usepackage{amsfonts}       
\usepackage{nicefrac}       
\usepackage{microtype}      
\usepackage{algorithm}
\usepackage{algorithmic}
\usepackage{array}
\usepackage{multirow}
\usepackage{enumitem}
\usepackage{makecell}
\usepackage{gensymb}         
\usepackage{mathtools}
\usepackage{dblfloatfix}
\usepackage{pifont}
\usepackage[permil]{overpic}

\usepackage{titletoc,tocloft}

\newcommand{\cmark}{\ding{51}}

\newcommand{\rvect}[1]{\begin{bmatrix} #1 \end{bmatrix}}

\makeatletter
\DeclareRobustCommand\onedot{\futurelet\@let@token\@onedot}
\def\@onedot{\ifx\@let@token.\else.\null\fi\xspace}

\def\eg{\emph{e.g}\onedot} 
\def\ie{\emph{i.e}\onedot}

\def\etal{\emph{et al}\onedot}
\makeatother

\usepackage{xcolor}
\usepackage{ifthen}
\usepackage{textcomp}
\definecolor{red}{rgb}{1,0,0}
\definecolor{slateblue}{rgb}{0.7,0.35,0.9}
\definecolor{green}{rgb}{0,1,0}
\definecolor{mahogany}{rgb}{0.75, 0.25, 0.0}
\definecolor{purple}{rgb}{0.6, 0, 0.6}
\definecolor{darkpurple}{rgb}{0.3, 0, 0.3}
\definecolor{darkgreen}{rgb}{0, 0.4, 0}
\definecolor{frenchblue}{rgb}{0.0, 0.45, 0.73}
\definecolor{blue}{rgb}{0,0,1}
\definecolor{goldenrod}{rgb}{0.65, 0.45, 0.03}
\definecolor{gray}{rgb}{0.5,0.5,0.5}
\definecolor{gold}{rgb}{1.0, 0.874, 0}
\definecolor{silver}{rgb}{0.67,0.67,0.67}
\definecolor{brown}{rgb}{0.8, 0.678, 0.4}

\usepackage[accsupp]{axessibility}  

\usepackage{cite}
\usepackage[pagebackref,breaklinks,colorlinks]{hyperref}
\usepackage[capitalize]{cleveref}
\begin{document}
\pagestyle{headings}
\mainmatter
\def\ECCVSubNumber{2312}  

\title{
Data Efficient 3D Learner via Knowledge Transferred from 2D Model
} 

\titlerunning{Data Efficient 3D Learner via Knowledge Transferred from 2D Model}
%
\author{Ping-Chung Yu \and
Cheng Sun \and
Min Sun}
%
%
\institute{National Tsing Hua University \\
\email{\{pingchungyu, chengsun\}@gapp.nthu.edu.tw, sunmin@ee.nthu.edu.tw}}
\maketitle

\begin{abstract}
Collecting and labeling the registered 3D point cloud is costly.
As a result, 3D resources for training are typically limited in quantity compared to the 2D images counterpart.
In this work, we deal with the data scarcity challenge of 3D tasks by transferring knowledge from strong 2D models via RGB-D images.
Specifically, we utilize a strong and well-trained semantic segmentation model for 2D images to augment RGB-D images with pseudo-label.
The augmented dataset can then be used to pre-train 3D models.
Finally, by simply fine-tuning on a few labeled 3D instances, our method already outperforms existing state-of-the-art that is tailored for 3D label efficiency.
We also show that the results of mean-teacher and entropy minimization can be improved by our pre-training, suggesting that the transferred knowledge is helpful in semi-supervised setting.
We verify the effectiveness of our approach on two popular 3D models and three different tasks.
On ScanNet official evaluation, we establish new state-of-the-art semantic segmentation results on the data-efficient track. Code: \url{https://github.com/bryanyu1997/Data-Efficient-3D-Learner}.
\end{abstract}

\section{Introduction}
\label{sec_intro}
Nowadays, 3D sensors are in demand by applications like AR/VR, 3D reconstruction, and autonomous driving.
To have a high-level scene understanding (\eg, recognition, semantic segmentation) on the captured 3D data, deep-learning-based models are typically employed for their outstanding performance.
As 3D sensors become easier accessible, the architecture of deep 3D models~\cite{QiSMG17,QiYSG17,WuQL19,YanZLWC20,ThomasQDMGG19,ChoyGS19,GrahamEM18} also progresses steadily for better result qualities.
In this work, we investigate an orthogonal direction to model architecture design---we present a novel 3D models pre-training strategy to improve performance in a model agnostic manner.

In 2D vision tasks, model pre-training on ImageNet~\cite{DengDSLL009} has become a commonly applied strategy for achieving better performances across different downstream tasks.
However, there is no standard large-scale dataset like ImageNet to pre-train 3D models due to the considerable effort to acquire and label a diverse set of point cloud data compared to the 2D counterpart.
As a result, 3D models are typically trained from scratch, which hinders the performance, especially under the data scarcity scenario of the registered point cloud.

\begin{figure}[t]
	\centering
	\includegraphics[width=\textwidth,trim={100 75 100 120},clip]{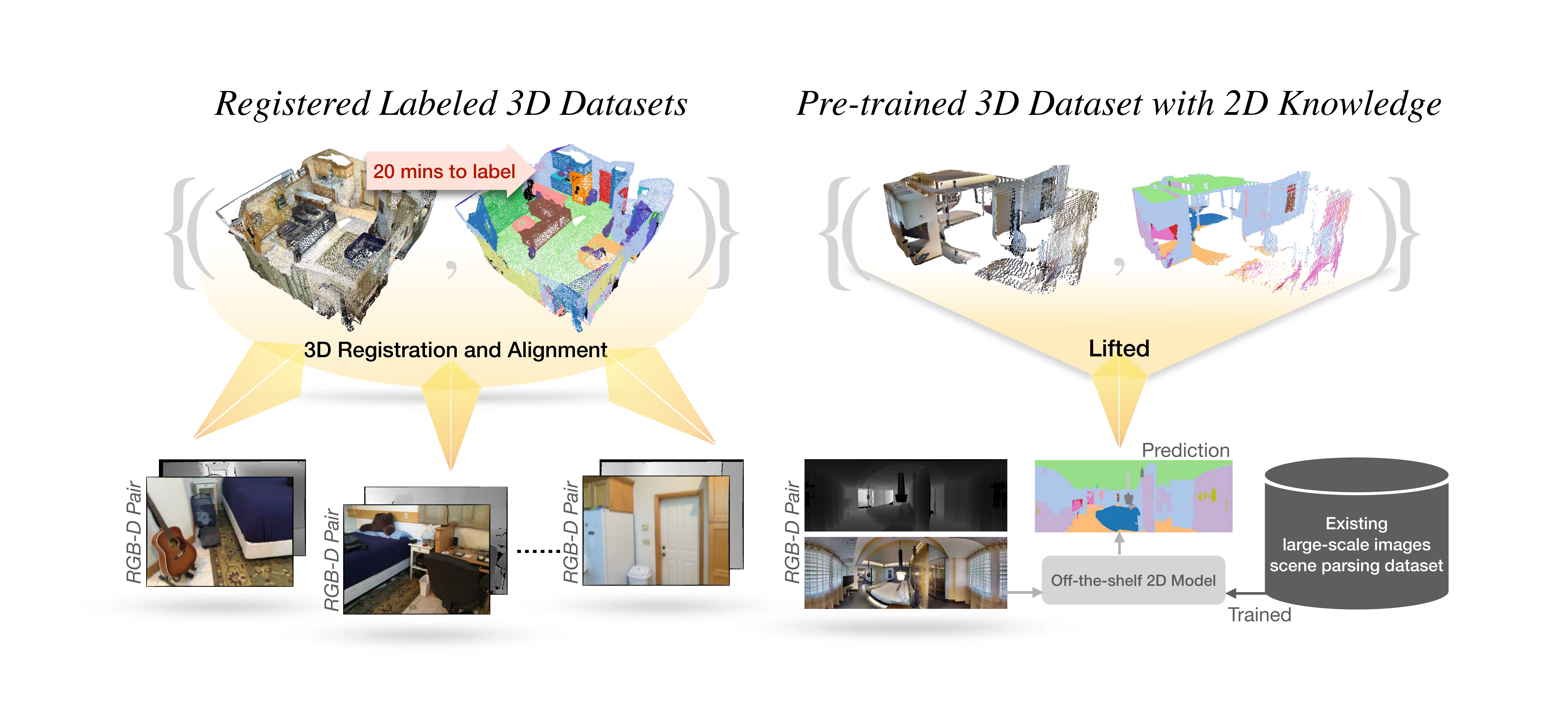}
	\caption{ 
        Left-panel: collecting labeled 3D data is challenging due to the need for (a) robust 3D registration and alignment processes, and (b) a time-consuming human labeling process.
Right-panel: a large amount of RGB-D data serve as the bridge between 2D and 3D knowledge. First, pseudo semantic labels are generated by applying an off-the-shelf 2D model on RBG images. Next, the pseudo-labeled 2D data are lifted to 3D using the associated depth map. Finally, we pre-train a 3D model with the large amount of pseudo-labeled 3D data.
	}
	\label{fig:teaser}
\end{figure}

To avoid the burden of data labeling, self-supervised learning has emerged as an alternative for pre-training 2D models without labels~\cite{ChenK0H20,CaronMMGBJ20,He0WXG20,ChenH21,GrillSATRBDPGAP20,HeCXLDG21}.
To reproduce this trend in 3D data, PointContrast~\cite{XieGGQGL20} uses contrastive loss to learn the correspondence between two point clouds with visual overlap, improving results on the downstream 3D tasks.
However, the diversity and scale of 3D datasets, even releasing the dependency on labeling, are still not comparable to the 2D datasets.
For instance, ScanNet~\cite{DaiCSHFN17}, which is used in PointContrast's pre-training, has only about a thousand indoor scenes, while ImageNet has more than a million images covering a thousand different classes.
As a result, the accuracy improvement by self-supervised learning on point clouds is still limited for 3D tasks.

To address the issue of limited resources of point clouds, we present a novel 3D model pre-training approach via transferring the learned knowledge of a 2D model via RGB-D datasets (see \cref{fig:teaser}).
The single view depth sensor is much cheaper than ever and could be widely popular as a built-in function of phones to capture various scenes.
Using RGB-D data as the bridge to transfer knowledge from strong 2D models to 3D models is thus a valuable direction to explore.
Specifically, we employ a 2D semantic segmentation model, which is trained on a large and diverse scene parsing dataset, to augment the RGB-D images with pseudo-labels.
We then train a 3D model to take the 3D point cloud lifted from RGB-D as input and reproduce the pseudo-labels.
By doing so, the 3D models can learn from the strong 2D teacher model and also see a large variety of scenes captured by the RGB-D data.

We demonstrate the effectiveness of our pre-training on  semantic segmentation for scene point cloud of the popular ScanNet dataset.
Annotating 3D scene point clouds is a demanding task, which takes more than 20 minutes to label a single scene.
Some recent approaches emerges to learn from fewer 3D labels to reduce the labeling cost.
However, the lack of large-scale pre-training hinders their performances.
We show that our pre-training with simple fine-tuning on the scarce label can already outperform existing results tailored for 3D data efficiency~\cite{XieGGQGL20,LiuQF21,HouGNX21}.
We also show our pre-training can boost the performance of the widely used semi-supervised techniques, which suggests that our pre-training provides an opportunity for future 3D research on both data efficiency and semi-supervised learning to build upon our results for better quality.
Finally, despite pre-training on a scene level, we also evaluate our models on object-level tasks and observe improved performances on 3D object classification and shape part segmentation.
This suggests that our pre-training is also well transferred to different tasks and input 3D scales.

We summarize our contributions as follows:
\begin{itemize}[topsep=0pt]
    \item We introduce a pre-training strategy to transfer knowledge from a strong 2D scene parsing model via RGB-D images to 3D models.
    \item We demonstrate the effectiveness of our pre-training under limited data scenario across two models (\ie, O-CNN~\cite{WangLGST17}, SparseConv~\cite{GrahamEM18}) and three different tasks (\ie, 3D object classification, point-cloud part segmentation, and indoor point-cloud segmentation).
    \item By simply finetuning our pre-trained model on a few labels, we establish new state-of-the-art results on ScanNet~\cite{DaiCSHFN17} official evaluation on data-efficient setups, verifying that our pre-training results in data-efficient 3D learners.
\end{itemize}

\section{Related work}
\paragraph{Deep 3D models for point cloud understanding.}
As the 3D analysis studies flourish, several deep models are proposed to extract point cloud features for a high-level understanding.
Existing approaches can be classified into point-based and voxel-based methods.
PointNet~\cite{QiSMG17} and PointNet++~\cite{QiYSG17} are the pioneering point-based methods to apply multi-layer perceptron layers directly on point clouds.
After that, several convolution-based models~\cite{LiBSWDC18,WuQL19,ThomasQDMGG19,Boulch20} are proposed, achieving better quality.
Recently, attention-based models~\cite{YanZLWC20} have become a new effective way for point cloud processing.
On the other hand, the voxel-based method is attractive for its computational-friendly 3D data representations.
A discretized step is typically applied on the point cloud before employing the voxel-based models~\cite{WuSKYZTX15,MaturanaS15}.

As the cubic memory complexity limits the resolution of dense voxel grids, 3D sparse CNNs~\cite{Graham15,GrahamEM18,ChoyGS19} emerges to achieve a feasible space-time complexity for scene-level point clouds, where the CNNs are working on occupied voxels only.
OctNet~\cite{RieglerUG17} and O-CNN~\cite{WangLGST17} further use the octree data structure to achieve a higher grid resolution efficiently.

\newpage
As the deep architectures for 3D point clouds progress steadily, all of them are still trained from scratch due to the lack of a large-scale point cloud dataset for pre-training.
To sidestep the issue of 3D resources, this work presents a model agnostic pre-training strategy using 2D resources to improve performances.

\paragraph{Data-efficient 3D.}
Data-efficient learning restricts the amount and the variety of labeled data of the target task, which is helpful for tasks on scene-level point clouds.
Registered scene point clouds are hard to acquire and time-consuming to label (\eg, 20 more minutes for a ScanNet scene).
Thus, data-efficient solutions are always welcome.
Existing works have explored self-supervised pre-training and semi-supervised learning using unlabeled scene point clouds to improve performance.
PointContrast~\cite{XieGGQGL20} sub-samples partial scans and use contrastive learning to pre-train 3D models to identify point correspondents.
CSC~\cite{HouGNX21} further improves the pre-training by incorporating spatial scene contexts into the objective.
As negative pairs sampling in contrastive learning could be ambiguous, ViewPointBN~\cite{LuoTZZ21} proposed to use correlation as the objective instead.
The aforementioned self-supervised pre-training achieves good label efficiency in a target task agnostic manner.
In OTOC~\cite{LiuQF21}, pseudo-label-based semi-supervised learning is employed to simultaneously learn from both labeled and unlabeled data, achieving superior label efficiency.
We use the abundant 2D resources to pre-train instead of 3D, achieving state-of-the-art label efficiency on scene point clouds. 

\paragraph{Knowledge transferred from 2D.}
Most recently, DepthContrast~\cite{ZhangGJM21} and Contrastive Pixel-to-point~\cite{liu2021learning} are also proposed to pre-train 3D models using 2D RGB-D datasets.
DepthContrast trains 3D models to discriminate 3D point clouds projected from different RGB-D images.
Contrastive Pixel-to-point treat each 3D point as instance and apply contrastive loss to learn from pixel features extracted by a trained 2D model.

One challenge of these contrastive-based methods is the ambiguity of the selected negative pairs.
Negative pairs sampling is a crucial problem for contrastive pre-training to avoid collapsing solutions~\cite{ChenH21,ZbontarJMLD21}.
The sampling strategy requires careful designs~\cite{He0WXG20,ChenK0H20} in the contrastive pre-training, even for images.
As the diversity of the employed indoor pre-training data is still limited, the model could have difficulty differentiate two room with similar setup or two 3D points belonging to the same stuff (\eg, walls).
Conversely, we directly train the 3D models to reproduce the pseudo labels generated by a strong and well-trained image scene parser, which is more straightforward but non-trivial to learn as we have clear and informative supervision for each 3D point.

2D3DNet~\cite{GenovaYKPCSBSF21} trains 3D models with only 2D supervision, which requires the 2D teacher to be a specialist for the downstream task and the 3D model has to co-work with the 2D teacher to achieve good results.
Our pre-trained 3D model does not work directly in the downstream 3D task, but it is generally benefit to different downstream task and our 3D model works independently after pre-training.

\section{Approach}
\begin{figure}[t]
	\centering
	\includegraphics[width=1\textwidth]{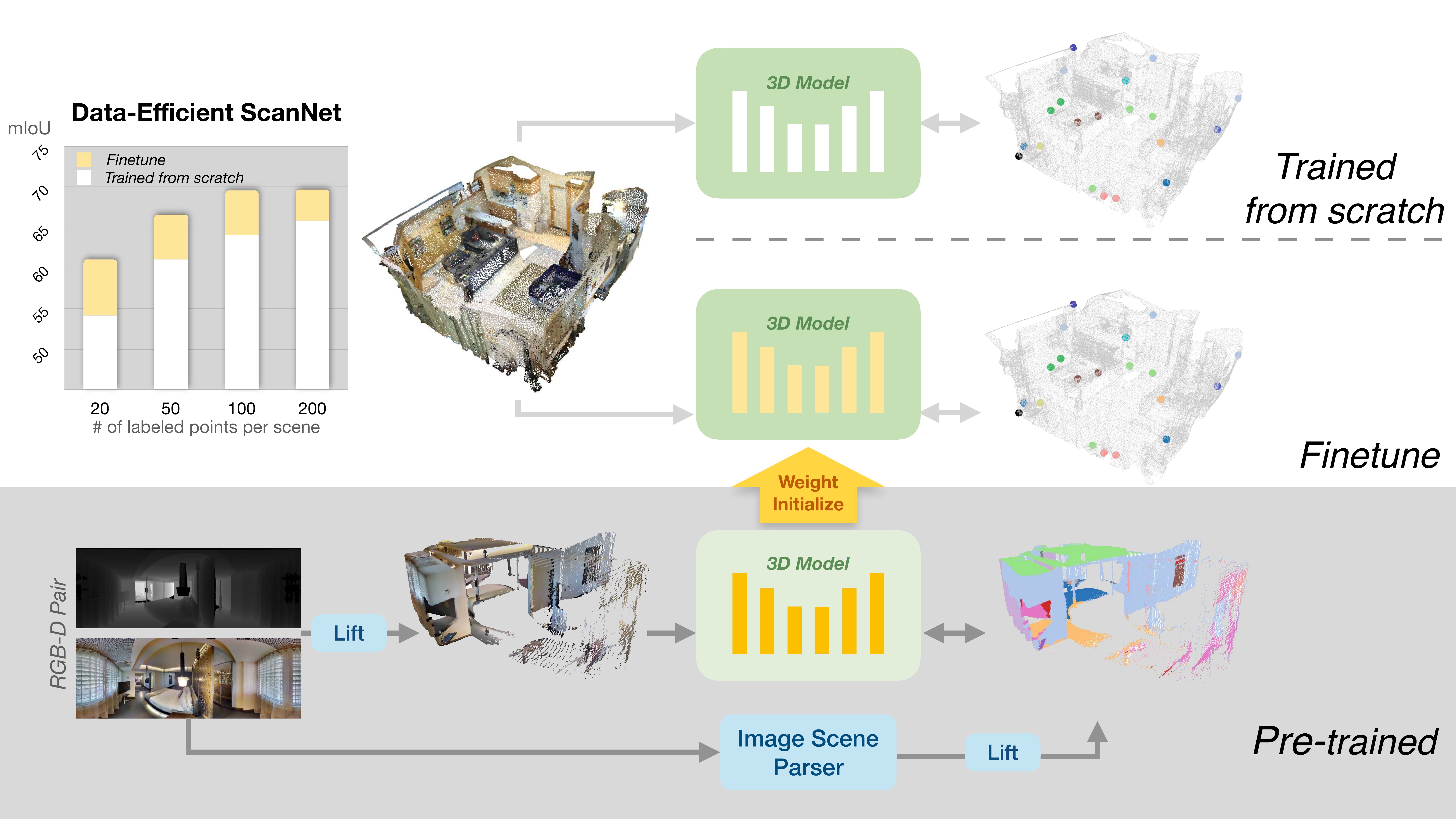}
	\caption{ 
        Overview of our approach. {\bf (Below)} We use a strong and well-trained image scene parser to augment single view RGB-D datasets with pseudo-labels, which is used to pre-train a 3D model in an architecture agnostic manner. {\bf (Top)} Our pre-training improves the results of the limited annotation training.
	}
	\label{fig:flowchart}
\end{figure}

Our goal is to pre-train 3D models in an architecture agnostic manner.
An overview of the proposed approach is illustrated in 
\cref{fig:flowchart}.
Below, we first give a general introduction to 3D models in \cref{ssec:3d_model}.
In \cref{ssec:pretrain}, we detail the proposed pre-training approach.
We also apply our pre-trained model on some semi-supervised techniques in \cref{subsec:semi-supervised}.

\subsection{3D model} \label{ssec:3d_model}
Our pre-training approach is model agnostic and does not require specific 3D architecture designs.
Here, we describe a general 3D encoder-decoder to extract deep features from point clouds.
A 3D point cloud $\bm{x}^{\text{(pts)}}$ represents a 3D scene or object by a set of 3D coordinates (with additional color or normal features).
For voxel-based models, a pre-process input layer before the first model layer is needed to discretize the coordinates into a regular 3D grid.
The encoder $E$, consisting of a sequence of 3D convolution layers, batch normalizations, and nonlinear activation (\eg, ReLU), is a bottom-up way to map the point clouds into a down-sampled latent features $\bm{z}^{\text{(e)}} = E(\bm{x}^{\text{(pts)}})$, which has the high-level understanding of the input scene or object.
The decoder $D$ then upsamples and incorporates the low-level feature into $\bm{z}^{\text{(e)}}$ to have a holistic point-wise deep feature $\bm{z}^{\text{(d)}} = D(\bm{z}^{\text{(e)}})$.

\paragraph{Classification.}
\label{para:classification}
To classify a whole input point cloud into a pre-defined set of classes, we discard the decoder $D$ as we do not need point-level features.
The high-level features $\bm{z}^{\text{(e)}}$ are first aggregated into a single latent vector via global max pooling, which is then followed by a classification head $\operatorname{ClsHead}$ to map the latent vector into a categorical distribution: $\bm{y}^{\text{(cls)}} = \operatorname{ClsHead}(\operatorname{GlobalMaxPool}(\bm{z}^{\text{(e)}}))$, where the $\operatorname{ClsHead}$ consists of a linear layer and a $\operatorname{Softmax}$ layer.

\paragraph{Point-level semantic segmentation.}
The semantic segmentation head simply maps the point features into per-point class distribution by a linear and a $\operatorname{Softmax}$ layer: $\bm{y}^{\text{(ss)}} = \operatorname{SegHead}(\bm{z}^{\text{(d)}})$.

\subsection{Knowledge transferred from 2D} \label{ssec:pretrain}
Our approach aims to pre-train a 3D model by transferring the knowledge of a strong 2D model learned from a large-scale 2D dataset.
However, the 2D model $F^{\text{(2D)}}$ takes images $\bm{x}^{\text{(img)}}$ as input while the 3D model takes point clouds $\bm{x}^{\text{(pts)}}$.
To learn from 2D models, we use RGB-D dataset $\left\{\left(\bm{x}^{\text{(img)}}, \bm{d}^{\text{(img)}}\right)\right\}$ as the bridge, where $\bm{d}^{\text{(img)}}$ is the depth map of a image.
Our idea is to use a well-trained 2D model to generate pseudo-label for the images in the RGB-D dataset
\begin{equation}
\bm{t} = F^{\text{(2D)}}\left(\bm{x}^{\text{(img)}}\right)~,
\end{equation}
and then we lift the image to point cloud using their depth maps
\begin{equation}
\bm{x}^{\text{(img2pts)}} = \operatorname{Lift}\left(\bm{x}^{\text{(img)}}, \bm{d}^{\text{(img)}}\right)
\end{equation}
so that we can train the 3D model using the augmented RGB-D dataset $\left\{\left(\bm{x}^{\text{(img2pts)}}, \bm{t}\right)\right\}$.
Below, we introduce the explored two different sources of RGB-D data and detail our pre-training strategy.

\paragraph{Lifting perspective images.} 
With the provided depth maps and the camera intrinsic $K$, we can lift a 2D image coordinate $\rvect{u, v}$ to the 3D camera coordinate $\rvect{x, y, z}$ by:
\begin{align}
\begin{bmatrix}
x \\ y \\ z 
\end{bmatrix}
= d\cdot K^{-1} \begin{bmatrix}
 u \\ v \\ 1 \end{bmatrix} ~,
\end{align}
where $d$ is the depth value of a pixel and $\rvect{u, v, 1}^\intercal$ is the homogeneous coordinate.

\paragraph{Lifting panoramic images.}
Panoramic images cover more information in one shot attributed to the omnidirectional field of view compared to perspective images.
Unlike perspective depth maps where z-values are recorded, panoramic depth maps directly record the distance between the observed 3D points to camera.
We can lift a panoramic image to 3D by:
\begin{align}
\left\{\begin{array}{llr}
 x &= d \cdot \cos(v) \cdot \cos(u) \ ; & \\
 y &= d \cdot \cos(v) \cdot \sin(u) \ ; & \\ 
 z &= - d \cdot \sin(v) ~, & 
\end{array}\right.
\end{align}
where $d$ is the recorded distance of a pixel in the panoramic depth maps, and $u \in \left[-\pi, \pi\right], v \in \left[-\frac{\pi}{2}, \frac{\pi}{2} \right]$ are the panoramic image coordinate in UV space.

\paragraph{Learning from 2D scene parser via soft pseudo-label.}
\label{subsec:predictor}
Image scene parsing aims to classify each pixel of images, which provides a thorough and detailed understanding of the captured scenes.
Besides, existing training corpora for image scene parsing~\cite{ZhouZPFB017,CaesarUF18} is abundant, with more than 10k images covering a large variety of scenes and classes.
Scene parsing models~\cite{ZhaoSQWJ17,ChenPKMY18,ZhuXBHB19,YuanCW20,HsiaoSCS21,RanftlBK21} have also progressed steadily and achieved strong performances.
In this work, we adopt DPT~\cite{RanftlBK21} for its outstanding performance, and we find it generalized well on both perspective and panoramic images.

We first attach a point semantic segmentation head $\operatorname{SegHead}^{\text{(pre)}}$ on the 3D encoder decoder to classify $C$ classes of the trained DPT
\begin{equation}
    \bm{y}^{\text{(img2pts)}} = \operatorname{SegHead}^{\text{(pre)}}\left(D\left(E\left(\bm{x}^{\text{(img2pts)}}\right)\right)\right) ~.
\end{equation}
Instead of the one-hot decision, soft labels are learned as the target during pre-training.
Soft-label is first introduced in a knowledge distillation method~\cite{HintonVD15}, where each pixel or point is labeled by the categorical distribution over classes.
We use cross-entropy loss as the training objective for the 3D model to learn from the soft-label:
\begin{equation}
    \mathcal{L}^{\text{(pretrain)}} = \frac{1}{N} \sum_{i=1}^{N} \sum_{c=1}^{C} \left(- \bm{t}_i[c] \log\left(\bm{y}^{\text{(img2pts)}}_i[c]\right)\right) ~,
\end{equation}
where $i$ is the index to the $N$ 3D points, and $t_i$ is the soft-label probability of the pixel corresponding to the $i$-th 3D point.

\paragraph{Downstream tasks.}
After pre-training, we discard the pre-training head $\operatorname{SegHead}^{\text{(pre)}}$ and directly used the trained encoder $E$ and decoder $D$ in the downstream tasks.

\subsection{Semi-Supervised}
\label{subsec:semi-supervised}
Semi-supervised learning is helpful to learn from scarce labels, where the improvement by our pre-training can be additive to it to achieve an even better result.
We use two simple and commonly used semi-supervised techniques in this work.
First, entropy minimization~\cite{GrandvaletB05} encourages concentration of the predicted probability distribution on the unlabeled data:
\begin{equation}
\mathcal{L}^{\text{(mini-entropy)}} = \frac{1}{N} \sum_{i=1}^{N} \sum_{c=1}^{C} \left(- \bm{y}_i[c] \log\left(\bm{y}_i[c]\right)\right) ~.
\end{equation}
Second, we use mean-teacher~\cite{TarvainenV17} to guide our model, where the weights of teacher model $\theta_{t}{}'$ is obtained by exponential moving averaging (EMA) of the weight of student model $\theta_t$ across training step $t$:
\begin{equation}
\theta_{t} {}' = \alpha\theta_{t-1} {}' + (1-\alpha)\theta_{t} ~,
\end{equation}
where $\alpha$ is a smoothing hyperparameter.
We use mean squared error to encourage consistency between our predictions $\bm{y}$ and teacher predictions $\bm{y}'$
\begin{equation}
\mathcal{L}^{\text{(consistency)}} = \frac{1}{N} \sum_{i=1}^N \left\|\bm{y}_i - \bm{y}_i'\right\|^{2} ~.
\end{equation}
Different from the 2d teacher model we mention in \cref{sec_intro}, the mean-teacher here is a typical semi-supervised technique.

\section{Experiments}
In this section, we evaluate our pre-training strategy under the data scarcity scenario.
The implementation details are provided in \cref{pretraining detail}.
To demonstrate the benefits of our approach, we conduct a series of experiments on scene understanding task in \cref{scene} and the shape analysis tasks in \cref{shape}.
We show the results of our pre-training on different 3D models in \cref{3dmodel} and present the ablation experiments in \cref{ablation}.

\subsection{Implement details}
\label{pretraining detail}
\paragraph{Pre-training images dataset.}
For perspective pre-training, we utilize SUNRGB-D~\cite{SongLX15}, which consists of more than 10{,}000 RGB-D indoor images and the corresponding pixel-wise semantic labels from 37 categories. In addition to the depth maps, SUNRGB-D also provides the intrinsic which allow us to lift the 3D scenes. We split the dataset into 5{,}285 training sets and 5{,}050 validation sets in the pre-training procedure.
We only used the provided ground-truth semantic maps in ablation experiments and used the generated pseudo-label by our approach in all other experiments.
For panoramic pre-training, Matterport3D~\cite{ChangDFHNSSZZ17} provides various indoor scenes captured by panoramic images, each of which covers a omnidirectional field of view.
Matterport3D contains 10{,}800 panoramic views with corresponding depth maps from 90 building-scale scenes, including 61 scenes for training, 11 for validation, and 18 for testing following official data splits. 

\paragraph{Image scene parser.}
We employ DPT~\cite{RanftlBK21} as a teacher model in our pre-training.
The DPT is first trained on the ADE20K~\cite{ZhouZPFB017} dataset, which has 20k images with ground-truth semantic segmentation maps covering 150 different classes.
In pre-training, the DPT is fixed and used to generate soft pseudo-labels to transfer its knowledge to the 3D models.

\paragraph{3D models.}
SparseConv~\cite{GrahamEM18} is trained by Adam optimizer~\cite{KingmaB14} for scene semantic segmentation. The models are trained on NVIDIA GTX 1080Ti GPUs for 200 epochs with batch size 4 and learning rate 0.001. O-CNN~\cite{WangLGST17} is implemented for shape analysis and indoor scene semantic segmentation. During lifted scenes pre-training, the projected-points are formed as $512^{3}$ resolution of leaf octants as the same as indoor scene semantic segmentation. For part segmentation, we train networks for 600 epochs with batch size 32, leaning rate 0.025 and weight decay 0.0001. For scene semantic segmentation, we train models for 500 epochs with batch size 4 and leaning rate 0.05. Both tasks adopt the SGD optimizer with a momentum 0.9 and use polynomial schedulers powered by 0.9. The models are trained on several NVIDIA Tesla V100-SXM2 GPUs. As the setup for part segmentation, the classification model is trained for 300 epochs with batch size 32 and learning rate 0.05. The O-CNNs are adopted with $32^{3}$, $64^{3}$ and $512^{3}$ resolution of leaf octants for object classification, shape part segmentation and scene semantic segmentation, respectively. The point-wise predictions are interpolated by linear for part segmentation and nearest for scene semantic segmentation.  


\paragraph{Data augmentation.}
We apply random rotation, random scaling, random elastic distortion, random color contrast, and random color jittering to the input point cloud as data augmentation.

\subsection{Data efficient scene semantic segmentation}
\label{scene}
We validate the effectiveness of our pre-training on ScanNet Data Efficient Benchmark.
ScanNetV2~\cite{DaiCSHFN17} consists of various indoor scenes formed as 3D point clouds and corresponding semantic annotations. The data is taken from 707 distinct spaces and covers 20 semantic classes. We follow the official data scarcity scenarios: {\it i)} Limited Annotations (LA) considers only a few labeled points in each scene, {\it ii)} and Limited Reconstructions(LR) considers the a few number of labeled scenes. 
We use O-CNN pre-trained by our approach on Matterport3D dataset for all the results submitted to official ScanNet.

\paragraph{Limited annotations.}
\label{para:LA}
Following the official configuration in the \emph{3D Semantic label with Limited Annotations benchmark}, there are four different scales, including $\left \{20, 50, 100, 200 \right \}$ labeled points per training scene. We show the quantitative comparison on the testing split in \cref{tab:scannet_LA}.
We achieve state-of-the-art results on purely supervised fine-tuning setup and semi-supervised setup.
It is worth noting that our results without using unlabeled data already outperforms the semi-supervised OTOC~\cite{LiuQF21} on 50, 100, and 200 labeled points per scene, which further highlights the effectiveness of our pre-training approach.


 
\begin{table}[t]
  \setlength{\tabcolsep}{10pt}
  \centering
  \caption{Quantitative comparisons on official ScanNet Limited Annotations(LA) track. For a fair comparisons, the results are separated based on whether the unlabeled data points are used in training.}
  \label{tab:scannet_LA}
  \begin{tabular}{@{}lc|cccc@{}}
    \toprule
    \multirow{2}{*}{Method} & \multirow{2}{*}{semi} & \multicolumn{4}{c}{\# of labeled points per scene} \\ 
     & & 20 & 50 & 100 & 200 \\
    \midrule
    PointContrast~\cite{XieGGQGL20} & - & 55.0 & 61.4 & 63.6 & 65.3\\
    CSC~\cite{HouGNX21}             & - & 53.1 &  61.2 &  64.4 & 66.5\\
    ViewPointBN~\cite{LuoTZZ21}     & - & 54.8 &  62.3 &  65.0 & 66.9\\
    Ours                            & - & \underline{57.9} & \underline{65.8} & \underline{\bf 71.4} & \underline{\bf 71.1}\\
    \midrule
    OTOC~\cite{LiuQF21}             & \cmark & 59.4 & 64.2 & 67.0 & 69.4\\
    Ours                            & \cmark & \underline{\bf 63.9} & \underline{\bf 69.5} & \underline{70.4} & \underline{70.9}\\
    \bottomrule
  \end{tabular}
\end{table}

\begin{table}[t]
  \setlength{\tabcolsep}{10pt}
  \centering
  \caption{Quantitative comparisons on offical ScanNet Limited Reconstructions(LR) track. }
  \label{tab:scannet_LR}
  \begin{tabular}{@{}lc|cccc@{}}
    \toprule
    \multirow{2}{*}{Method} & \multirow{2}{*}{semi} & \multicolumn{4}{c}{percentage of labeled scene} \\ 
     & & 1\% & 5\% & 10\% & 20\% \\
    \midrule
    PointContrast~\cite{XieGGQGL20} & - & 25.3 & 43.8 & 55.5 & 60.3\\
    CSC~\cite{HouGNX21}             & - & \underline{\bf 27.0} & 46.0 & 57.5 & 61.2\\
    ViewPointBN~\cite{LuoTZZ21}     & - & 25.6 & 45.2 & 56.6 & 62.5\\
    Ours                            & - & 26.6 & \underline{46.7} & \underline{\bf{61.2}} & \underline{64.0}\\
    \midrule
    Ours                            & \cmark & 26.3 & \bf{50.8} & 60.8 & \bf{66.3}\\
    \bottomrule
  \end{tabular}
\end{table}

\paragraph{Limited reconstruction.}
\label{para:LR}
Limited Reconstruction (LR) is constructed by limiting the number of scenes.
The official subset is randomly sampled from 1201 scenes, where $\left \{1\%, 5\%, 10\%, 20\% \right \}$ of training data is subsampled (corresponding to 12, 60, 120, and 240 scenes). 
We compare our results on \emph{3D Semantic label with Limited Reconstructions benchmark} in \cref{tab:scannet_LR}.
When only $1\%$ of the training data is available, all methods achieve similar performance.
When $5\%, 10\%, 20\%$ of the training data is given, we achieve a superior mIoU compared to previous methods.
Additionally, we adopt semi-supervised learning and examine the capability under the limited reconstruction scenario. 
We first trained our model on the given labeled data, and then we randomly sample a subset of unlabeled scenes to produce pseudo labels.
The pseudo labels is generated by selecting the top-most 20\% confident predictions of each class.
We can then combined the labeled, pseudo-labeled, and unlabeled data points to train our models.

From the comparisons in \cref{tab:scannet_LA} and \cref{tab:scannet_LR}, we find our pre-trained models with simple supervised fine-tuning already outperform existing methods tailored for 3D data-efficiency.
Combining our pre-training with semi-supervised sometimes can further improve our results.
In cases that semi-supervised does not improve, we still achieve similar performance.
As we only use the simplest semi-supervised techniques, a more tailored one could be more helpful.
In sum, our pre-training provides a good starting point for future work to develop 3D data-efficient approach.

\subsection{Pre-training on different 3D models}
\label{3dmodel}
To verify the effectiveness of our pre-training strategy on different 3D models, we implement O-CNN~\cite{WangLGST17} and SparseConv~\cite{GrahamEM18} for 3D scene semantic segmentation. 
SparseConv uses a sparse voxelized input representation and keeps the same level of sparsity throughout the model.
O-CNN builds up the octree structure of 3D points representation. As the octree depth increases, the octree-based CNN layers extract the information with higher resolution.
Both models work on occupied point sets only to ensure computational efficiency, and the networks are constructed by U-Net~\cite{RonnebergerFB15,GrahamEM18} architecture to produce point-wise predictions. 
The overall results are summarized in \cref{tab:image_modality}.
On both SparseConv and O-CNN, all different variations of our pre-training approach improved mIoU significantly over training from scratch.

 \begin{table}[t]
  \setlength\tabcolsep{5pt} 
  \centering
  \caption{mIoU of different 3D segmentation model and different pre-training strategy. The results are evaluated on ScanNet validation set. The oracle pre-training supervision is the ground-truth semantic segmentation maps provided in the pre-training dataset.}
  \label{tab:image_modality}
  \begin{tabular}{@{}llc|cc@{}}
    \toprule
    \multirow{2}{*}{Based model} & \multirow{2}{*}{Pre-training strategy} &\multirow{2}{*}{\makecell[c]{Oracle pre-training\\supervision}} & \multicolumn{2}{c}{LA (points)} \\ & & & 20 & 200 \\
    \midrule
    \multirow{4}{*}{SparseConv} 
    & trained from scratch  & - & 51.6 & 65.4 \\
    \cline{2-5} & perspective pre-training & \cmark & 55.4 & 66.0 \\ 
    & perspective pre-training & -  & 56.2 & 65.7  \\ 
    & panoramic pre-training & - & 58.5 & 67.3\\ 
    \midrule
    \multirow{4}{*}{O-CNN} 
    & trained from scratch  & - & 52.9 & 65.0 \\
    \cline{2-5} & perspective pre-training & \cmark & 56.9 & 66.2 \\ 
    & perspective pre-training & - & 56.2 & 65.3 \\
    & panoramic pre-training & - & 55.6 & 67.8\\ 

    \bottomrule
  \end{tabular}
\end{table}

\subsection{Shape analysis under limited data scenario}
Our approach discussed above presents a positive outcome on scene-level semantic segmentation under the limited data scenario.
In this section, we apply our pre-trained model on object-level 3D tasks such as object classification and shape part segmentation.
With the data scarcity scenario for shape analysis, the implementation details and results are presented in the following paragraph.
For all the experiments in shape analysis, we use O-CNN pre-trained by our approach on Matterport3D, which can also show the generalizibility from scene-level to object-level of our pre-training.

\label{shape}
\paragraph{Object classification}
Object classification is the fundamental task of shape analysis. Given the 3D shape represented as point clouds, the classification model intends to assign the category of the input objects. With our approaches, the 3D models are required to extract features by limited annotations.

\emph{Dataset}. ModelNet40~\cite{WuSKYZTX15} dataset contains 12{,}311 shapes from 40 object categories, and is split into 9{,}843 objects for training and 2{,}468 objects for testing. For the limited data  scenario, we randomly sample the subset by ratio $\left \{ 1\%, 5\%, 20 \% \right \}$ in each category from the training set, and evaluate on the original testing set.
For semi-supervised learning, the unlabeled batches is sampled randomly from the remained data of the training set.

\emph{Results}. We finetune the 3D models with panoramic images pre-trained for object classification.
The results are shown in \cref{tab:limited scene variety of cls.}, which is divided by whether the semi-supervised training is used.
Considering the model trained without unlabeled data, our pre-training strategy can directly improve the accuracy compared with the model trained from scratch.
In the cases that the models are trained from scratch, the improvements of using the unlabeled data are limited.
Conversely, our pre-training with only supervised fine-tuning already outperform the results of semi-supervised with random weight initialization.
Besides, the improvements by semi-supervised learning is much significant than the improvement from random initialization.

\begin{table}[t]
  \renewcommand{\arraystretch}{0.9}
  \setlength\tabcolsep{5pt} 
  \centering
  \caption{Accuracy of 3D object classification in ModelNet40~\cite{WuSKYZTX15} dataset under limited training data. The training data is random sampled by 1\%, 5\%, 20\% and the results are evaluated on the same testing set. }
  \label{tab:limited scene variety of cls.}
  \begin{tabular}{@{}lc|ccc@{}}
    \toprule
    Pre-training strategy & Semi-supervised & 1\% & 5\% & 20\%   \\
    \midrule
    \multirow{2}{*}{Trained from scratch}  & - & 60.2 & 79.6 & 86.5 \\
      & \cmark & \underline{61.1} & \underline{80.4} & \underline{87.6} \\
    \midrule
    \multirow{2}{*}{Our pre-training}  & - & 65.5 & 80.6 & 87.5 \\
     & \cmark & \underline{\bf 69.0} & \underline{\bf 82.9} & \underline{\bf 89.4} \\
    \bottomrule
  \end{tabular}
\end{table}

\begin{table}[t]
  \renewcommand{\arraystretch}{1.1}
  \setlength\tabcolsep{8pt} 
  \centering
  \caption{mIoU of shape part segmentation in ShapeNet~\cite{WuSKYZTX15} dataset under sampled training data. We use category mIoU across all categories and instance mIoU across all shape instance as the evaluation metrics. The results are evaluated on ShapeNet testing set which split by~\cite{YiKCSYSLHSG16}. }
  \label{tab:limited scene variety of part seg.}
  \begin{tabular}{lccc|ccc}
    \toprule
    \multirow{2}{*}{Pre-training strategy} & \multicolumn{3}{c|}{Category mIoU} & \multicolumn{3}{c}{Instance mIoU} \\ \cline{2-7} & 1\% & 5\% & 20\%  & 1\% & 5\% & 20\% \\
    \midrule
    Trained from scratch     &  63.3 & 67.5 & 73.9 & 67.9 & 72.2 & 76.7 \\
    Our pretraining &  {\bf 64.1} & {\bf 74.1} & {\bf 76.3} & {\bf 68.5} & {\bf 76.8} & {\bf 78.9} \\
    \bottomrule
  \end{tabular}
\end{table}

\paragraph{Shape part segmentation}
Shape part segmentation is more complicated than classification for shape analysis. Part segmentation is expected to generate the dense prediction and assign the part category to each point in objects.  

\emph{Dataset}. Yi~\etal~\cite{YiKCSYSLHSG16} annotates a subset of ShapeNet~\cite{WuSKYZTX15} 3D models with semantic part labels. The annotated subset of ShapeNet contains 2 to 6 parts per category, and 50 distinct parts in total among 16 shape categories. For training on the limited scenario, we build the limited subset by random sampling as ratio $\left \{ 1\%, 5\%, 20\% \right \}$ from each shape category, and then evaluate on the original validation set. According to the setup of pre-processing in~\cite{WangLGST17}, the input points are condensed by triangle faces and built in octree structure. 

\emph{Results}
We finetune the segmentation models with panoramic images pre-training in each category separately, and evaluate the performance of part segmentation by mean IoU across all categories and mIoU across all object instances. The results are shown in \cref{tab:limited scene variety of part seg.}.
As the data split in\textcolor{red}~\cite{ZhaoBDT19, HassaniH19, WangYZWL021} are not provided, we only compare the results on our own split.
With the knowledge distilled from off-the-shelf image scene parser, the pre-trained model performs better category mIoU and instance mIoU across all limited scales (1\%, 5\% and 20\%).

\subsection{Ablation study}
\label{ablation}
\paragraph{Image modality}
To observe the effect of the distillation via different image modalities, we pre-train O-CNN~\cite{WangLGST17} on perspective images and panoramic images, and the supervised fine-tune the O-CNN on ScanNet with limited annotations.
The results are demonstrated in \cref{tab:image_modality}.
For all the variations of our pre-training, we achieve significant better results than training from scratch.
Note that the panoramic images pre-training rises the performance by 2.7\% and 2.8\% mIoU compared with baseline on 20, 200 labeled points of scene(55.6\% vs 52.9\% and 67.8\% vs 65.0\%).
The results imply that models can learn non-trivial and informative representation from the strong and well-trained image scene parser via RGB-D images.
The learned representation boosts the capability of scene understanding under the limited annotations.

This work can not conclude whether perspective pre-training or panoramic pre-training is better, as the number of their training data is different.
The panoramic data used in this work leads us to better results, so we use panoramic pre-training in all the other experiments.

\paragraph{Pre-training by 2D annotations and pseudo labels}
\label{para: 2D annotation using}
In our pre-training strategy, we supervise 3D models by the pseudo-labels introduced in \cref{subsec:predictor}.
To examine the difference between the information distilled from pseudo labels and ground-truth labels, we conduct our pre-training strategy on perspective images with ground-truth semantic labels and generated pseudo-labels.
The ground-truth semantic labels are provided by SUNRGB-D~\cite{SongLX15}, while the pseudo labels are predicted by a strong and well-trained image scene parser.
With the 3D model trained on limited annotations scenario, the results are presented in \cref{tab:image_modality}.
The perspective images attain higher performance than the model trained from scratch regardless of whether the ground-truth labels are provided.
The results imply that the pseudo labels can also improve the knowledge distillation during pre-training. Additionally, with provided annotations of images, our pre-training strategy achieves better mIoU than the baseline by 4\% mIoU on 20 points and 1\% mIoU on 200 points. 
Overall speaking, the less constrained pseudo-labels pre-training can achieve comparable results to the ground-truth labels pre-training, which suggests that ground-truth semantic maps are not necessary in our pre-training approach.

\begin{table}[t]
  \setlength\tabcolsep{10pt} 
  \centering
  \caption{Ablation studies of our pre-training with and without semi-supervised learning. We follow the limited training data from official. The results are presented as mIoU and compared on ScanNet validation set. }
  \label{tab:LA semi supervised}
  \begin{tabular}{@{}lc|cccc@{}}
    \toprule
    \multirow{2}{*}{Method} & \multirow{2}{*}{semi} & \multicolumn{4}{c}{\# of labeled points per scene} \\ 
     & & 20 & 50 & 100 & 200 \\
    \midrule
    Trained from scratch    & - & 54.2 & 61.1 &  64.1 & 65.9 \\
    Our pre-training            & - & {\bf 55.9} & {\bf 63.4} & {\bf 70.2} & {\bf 69.4} \\
    \midrule
    Trained from scratch    & \cmark & 56.6 & 62.5 & 65.3 & 66.0 \\
    Our pre-training            & \cmark & {\bf 61.1} & {\bf 66.6} & {\bf 69.6} & {\bf 69.7} \\
    \bottomrule
  \end{tabular}
\end{table}

\begin{table}[t]
  \setlength\tabcolsep{8pt} 
  \centering
  \caption{mIoU of limited scene variety on ScanNet validation set.}
  \label{tab:LR semi supervised}
  \begin{tabular}{@{}ll|cccc@{}}
    \toprule
    \multirow{2}{*}{Model} & \multirow{2}{*}{Pre-training strategy}  & \multicolumn{4}{c}{LR(\%)} \\ & & 1 & 5 & 10 & 20 \\
    \midrule
    \multirow{3}{*}{O-CNN} 
    & Trained from scratch          & 18.7 & 39.7 & 52.4 & 59.6  \\
    \cline{2-6}
    & Our pre-training                  & 26.3 & 44.9 & 56.5 & 62.7 \\ 
    & Our pre-training + Semi-supervised  & {\bf 27.5} & {\bf 49.4} & {\bf 59.2} & {\bf 64.8} \\

    \bottomrule
  \end{tabular}
\end{table}

\paragraph{Combining pre-training with semi-supervised learning}
We study the effectiveness of our pre-training when combining with semi-supervised learning to simultaneously learn from labeled and unlabeled data.
We run semi-supervised learning with random initialized model weights and our pre-trained model weights, and then evaluate the performance on ScanNetv2 validation sets. The approach is examined on both limited annotation(LA) and limit reconstruction(LR).

For LA, the results are summarized in \cref{tab:LA semi supervised}. In ``trained from scratch" column, the semi-supervised learning slightly improves mIoU when unlabeled data is used during training. In contrast, our pre-training strategy increases the performance significantly when combining with the semi-supervised learning, particularly in the 20, 50 of labeled points per scene. Using unlabeled data with our pre-training strategy outperforms the models trained from scratch by 4.5\% and 4.1\%.(61.1\% vs 56.6\% and 66.6\% vs 62.5\%).

For LR, we follow \cref{para:LR} for semi-supervised training. As a result, \cref{tab:LR semi supervised} shows that our pre-training strategy achieves greater mIoU by 27.5\%, 49.4\%, 59.2\% and 64.8\% on $\left \{ 1\%, 5\%, 10\%, 20\% \right \}$. Consequently, unlabeled data enhances the ability of scene understanding on both limited annotations and limited reconstruction.



\section{Conclusion}
This work presents a new 3D deep models pre-training strategy.
We use RGB-D images as the bridge to transfer the knowledge from a strong and well-trained 2D scene parsing network to 3D models.
Our pre-training strategy is model agnostic, and we show its effectiveness on two popular 3D models architectures.
On the official ScanNet data-efficient track, we establish new state-of-the-art results.
Besides, we also show that the improvement by our pre-training is additive to other label-efficient techniques.
We hope our pre-trained weights can serve as a stepping stone for future 3D approaches and encourage more exploration on how to make good use of 2D resources in 3D tasks.

\subsubsection{Acknowledgements} This work is supported in part by Ministry of Science and
Technology of Taiwan (MOST 110-2634-F-002-051). We would like to thank National
Center for High-performance Computing (NCHC) for computational
and storage resource.

\clearpage

\title{
Data Efficient 3D Learner via Knowledge Transferred from 2D Model
} 
\subtitle{Supplementary Material}

\author{}
\institute{}

\maketitle

\setcounter{section}{0}
\renewcommand\thesection{\Alph{section}}
\setcounter{table}{0}
\renewcommand\thetable{\Alph{table}}
\setcounter{figure}{0}
\renewcommand\thefigure{\Alph{figure}}

\vspace{-0.2in}
In this supplementary material, we show the per-category breakdown comparisons in \cref{sec:quantitative}, and then the qualitative results are showed in \cref{sec:qualitative}. In \cref{sec:depthcontrast}, we compare with DepthContrast~\cite{ZhangGJM21} on the efficient ScanNet setup. In \cref{sec:imple detail}, we provide more implement details. Finally, we discuss the effectiveness of image scene parser in \cref{sec:softlabel}. 

\section{Quantitative results of per-category mIoU}
\label{sec:quantitative}
We use mean IoU to validate the effectiveness of our approach on ScanNet~\cite{DaiCSHFN17} Data Efficient Benchmark in our main paper.
Below, we expand the results presented in the main paper with per-category breakdown.

\paragraph{Limited annotations.}
Following the official configuration in the \emph{3D Semantic
label with Limited Annotations benchmark}, we demonstrate the state-of-the-art results of Table 1 in the main paper. Here we show the detailed performance of each category on 20, 50, 100 and 200 labeled points per scene in \cref{tab:LA_20}, \cref{tab:LA_50}, \cref{tab:LA_100}, and \cref{tab:LA_200} respectively. 

\begin{table}
  \setlength\tabcolsep{2.5pt} 
  \centering
  \caption{20-points LA}
  \label{tab:LA_20}
  \scalebox{0.5}{
  \begin{tabular}{l||c|cccccccccccccccccccc}
    \toprule
    points & avg.&bathtub&bed&book.&cabinet&	chair&counter&curtain&desk&door&floor& otherf. &picture&frige&shower&	sink&sofa&table&toilet&wall&window \\
    \toprule
    PointContrast & 0.550 &0.735 &0.676 &0.601 &0.475 &0.794 &0.288 &0.621 &0.378 &0.430 &0.940 &0.303 &0.089 &0.379 &0.580 &0.531 &0.689 &0.422 &0.852 &0.758 &0.468 \\
    CSC & 0.531 &0.659 &0.638 &0.578 &0.417 &0.775 &0.254 &0.537 &0.396 &0.439 &0.939 &0.284 &0.083 &0.414 &0.599 &0.488 &0.698 &0.444 &0.785 &0.747 &0.440 \\
    ViewPointBN & 0.548 &0.747 &0.574 &0.631 &0.456 &0.762 &0.355 &0.639 &0.412 &0.404 &0.940 &0.335 &0.107 &0.277 &\bf0.645 &0.495 &0.666 &0.517 &0.818 &0.740 &0.431 \\
    OTOC & 0.594 &0.756 &0.722 &0.494 &0.546 &0.795 &0.371 &0.725 &\bf0.559 &0.488 &\bf0.957 &0.367 &\bf0.261 &\bf0.547 &0.575 &0.225 &0.671 &0.543 &0.904 &\bf0.826 &0.557 \\
    Ours & \bf 0.639&\bf 0.839&\bf0.723&\bf0.681&\bf0.629&\bf0.839&\bf0.424&\bf0.728&0.538&\bf0.526&0.945&\bf0.427&0.12&0.511&0.643&\bf0.547&\bf0.781&\bf0.566&\bf0.905&0.809&\bf0.607 \\
    
    \bottomrule
  \end{tabular}
  }
  \vspace{-0.5in}
\end{table}

\begin{table}
  \setlength\tabcolsep{2.5pt} 
  \centering
  \caption{50-points LA}
  \label{tab:LA_50}
  \scalebox{0.5}{
  \begin{tabular}{l||c|cccccccccccccccccccc}
    \toprule
    points & avg.&bathtub&bed&book.&cabinet&	chair&counter&curtain&desk&door&floor& otherf. &picture&frige&shower&	sink&sofa&table&toilet&wall&window \\
    \toprule
    PointContrast &0.614 &0.844 &0.731 &0.681 &0.590 &0.791 &0.348 &0.689 &0.503 &0.502 &0.942 &0.361 &0.154 &0.484 &0.624 &0.591 &0.708 &0.524 &0.874 &0.793 &0.536 \\
    CSC & 0.612 &0.747 &0.731 &0.679 &0.603 &0.815 &0.400&	0.648 &0.453 &0.481 &0.944 &0.421 &0.173 &0.504 &0.623 &0.588 &0.690 &0.545 &0.877 &0.778 &0.541 \\
    ViewPointBN & 0.623 &0.812 &0.743 &0.654 &0.579 &0.800 &\bf0.462 &0.713 &0.533 &0.516 &0.944 &0.434 &0.215 &0.437 &0.521 &0.601 &0.720 &0.563 &0.884 &0.800 &0.534 \\
    OTOC & 0.642 &0.725 &0.735 &0.717 &0.635 &0.829 &0.457 &0.639 &0.421 &0.552 &\bf0.967 &0.460 &\bf0.240 &0.558 &\bf0.788 &0.621 &0.720 &0.477 &\bf0.915 &\bf0.842 &0.539 \\
    Ours & \bf0.695&\bf0.897&\bf0.784&\bf0.728&\bf0.697&\bf0.846&0.441&\bf0.77&\bf0.615&\bf0.585&0.951&\bf0.504&0.232&\bf0.672&0.76&\bf0.655&\bf0.772&\bf0.599&0.877&0.834&\bf0.678 \\

    \bottomrule
  \end{tabular}
  }
  \vspace{-0.5in}
\end{table}

\begin{table}
  \setlength\tabcolsep{2.5pt} 
  \centering
  \caption{100-points LA}
  \label{tab:LA_100}
  \scalebox{0.5}{
  \begin{tabular}{l||c|cccccccccccccccccccc}
    \toprule
    points & avg.&bathtub&bed&book.&cabinet&	chair&counter&curtain&desk&door&floor& otherf. &picture&frige&shower&	sink&sofa&table&toilet&wall&window \\
    \toprule
    PointContrast & 0.636 &0.694 &0.738 &0.731 &0.653 &0.817 &0.467 &0.651 &0.517 &0.522 &0.946 &0.479 &0.198 &0.575 &0.526 &0.649 &0.747 &0.569 &0.845 &0.803 &0.600 \\
    CSC & 0.644 &0.761 &0.707 &0.703 &0.642 &0.813 &0.436 &0.659 &0.502 &0.516 &0.945 &\bf0.487 &0.238 &0.538 &0.678 &0.659 &0.739 &0.568 &0.915 &0.811 &0.566 \\
    ViewPointBN & 0.650 &\bf0.778 &0.731 &0.688 &0.617 &0.812 &0.446 &0.739 &0.618 &0.540 &0.945 &0.415 &0.204 &0.623 &0.676 &0.594 &0.744 &0.576 &0.868 &0.811 &0.582\\
    OTOC & 0.670 &0.734 &\bf0.815 &0.661 &0.644 &\bf0.841 &\bf0.509 &0.741 &0.479 &0.548 &\bf0.968 &0.461 &\bf0.251 &0.664 &0.754 &0.656 &0.744 &0.541 &\bf0.917 &0.844 &0.625 \\
    Ours & \bf0.704&0.774&0.766&\bf0.764&\bf0.687&0.832&0.413&\bf0.79&\bf0.639&\bf0.599&0.952&0.478&0.222&\bf0.746&\bf0.859&\bf0.678&\bf0.806&\bf0.607&0.915&\bf0.847&\bf0.703 \\

    \bottomrule
  \end{tabular}
  }
  \vspace{-0.5in}
\end{table}

\begin{table}
  \setlength\tabcolsep{2.5pt} 
  \centering
  \caption{200-points LA}
  \label{tab:LA_200}
  \scalebox{0.5}{
  \begin{tabular}{l||c|cccccccccccccccccccc}
    \toprule
    points & avg.&bathtub&bed&book.&cabinet&	chair&counter&curtain&desk&door&floor& otherf. &picture&frige&shower&	sink&sofa&table&toilet&wall&window \\
    \toprule
    PointContrast & 0.653 &0.717 &0.775 &0.754 &0.626 &0.804 &0.391 &0.689 &0.485 &0.572 &0.945 &0.448 &0.232 &0.603 &0.813 &0.591 &0.775 &0.537 &0.885 &0.816 &0.608 \\
    CSC & 0.665 &0.857 &0.756 &\bf0.763 &0.647 &\bf0.852 &0.432 &0.684 &0.543 &0.514 &0.948 &0.469 &0.179 &0.599 &0.702 &0.620 &0.789 &\bf0.614 &0.911 &0.815 &0.607 \\
    ViewPointBN & 0.669 &0.847 &0.732 &0.724 &0.613 &0.827 &0.443 &0.742 &0.562 &0.551 &0.947 &0.441 &0.218 &0.650 &0.753 &0.621 &0.765 &0.601 &0.905 &0.814 &	0.618\\
    OTOC & 0.694 &0.760 &\bf0.815 &0.706 &0.684 &0.840 &\bf0.492 &0.701 &0.557 &0.596 &\bf0.972 &\bf0.497 &\bf0.281 &\bf0.709 &0.757 &0.689 &0.789 &0.600 &0.907 &\bf0.864 &0.6\\
    Ours &\bf0.709& \bf0.877&0.772&0.744&\bf0.694&0.836&0.453&\bf0.787&\bf0.623&\bf0.598&0.953&0.49&0.216&0.682&\bf0.879&\bf0.727&\bf0.802&0.604&\bf0.922&0.845&\bf0.676 \\

    \bottomrule
  \end{tabular}
  }
  \vspace{-0.5in}
\end{table}

\newpage

\begin{table}
  \setlength\tabcolsep{2.5pt} 
  \centering
  \caption{1\% LR}
  \label{tab:LR_1}
  \scalebox{0.5}{
  \begin{tabular}{l||c|ccccccccccccccccccccc}
    \toprule
    rate & avg.&bathtub&bed&book.&cabinet&	chair&counter&curtain&desk&door&floor& otherf. &picture&frige&shower&	sink&sofa&table&toilet&wall&window \\
    \toprule
    PointContrast & 0.253 &0.000 &0.412 &0.347 &0.137 &0.564 &0.140 &\bf0.361 &0.187 &0.249 &0.914 &0.092 &0.055 &0.102 &0.000 &0.048 &0.392 &0.302 &0.000 &0.697 &0.056 \\
    CSC & \bf0.270 &0.000 &0.528 &0.331 &0.139 &0.535 &0.118 &0.326 &0.222 &0.292 &0.921 &0.089 &0.163 &\bf0.129 &0.000 &\bf0.131 &\bf0.463 &0.278 &0.000 &\bf0.699 &0.033 \\
    ViewPointBN & 0.256 &0.000 &0.479 &\bf0.377 &\bf0.204 &0.551 &\bf0.205 &0.219 &\bf0.235 &0.224 &0.903 &\bf0.092 &0.088 &0.122 &0.000 &0.003 &0.354 &0.354 &0.000 &0.676 &0.034 \\
    Ours & 0.263&0&\bf0.547&0.235&0.184&\bf0.566&0.165&0.249&0.196&\bf0.309&\bf0.938&0.07&\bf0.186&0.069&0&0&0.368&\bf0.356&0&0.698&\bf0.118\\
    
    \bottomrule
  \end{tabular}
  }
  \vspace{-0.4in}
\end{table}

\begin{table}
  \setlength\tabcolsep{2.5pt} 
  \centering
  \caption{5\% LR}
  \label{tab:LR_5}
  \scalebox{0.5}{
  \begin{tabular}{l||c|ccccccccccccccccccccc}
    \toprule
    rate & avg.&bathtub&bed&book.&cabinet&	chair&counter&curtain&desk&door&floor& otherf. &picture&frige&shower&	sink&sofa&table&toilet&wall&window \\
    \toprule
    PointContrast & 0.438 & 0.517 & 0.659 & 0.251 & 0.332 & 0.783 & 0.244 & 0.408 & 0.411 & 0.409 & 0.935 & 0.206 & 0.119 & 0.200 & 0.048 & 0.355 & 0.682 & 0.414 & 0.647 & 0.743 & 0.391 \\
    CSC & 0.460 & 0.472 & \bf0.731 & \bf0.465 & 0.398 &\bf0.817 &0.292 &0.442 & 0.311 & 0.387 & 0.939 & 0.218 &0.181 &0.302 &\bf0.076 &\bf0.449 &\bf0.743 & 0.430 & 0.444 & 0.737 & 0.368 \\
    ViewPointBN & 0.452 & 0.587 & 0.569 & 0.172 & 0.391 & 0.769 & 0.290 & \bf0.512 & 0.501 & 0.373 & 0.935 & 0.251 & 0.173 & 0.201 & 0.003 & 0.352 & 0.619 & 0.454 & \bf0.783 & 0.719 & 0.390 \\
    Ours &\bf0.508& \bf0.824&0.53&0.314&\bf0.479&0.746&\bf0.334&0.49&\bf0.508&\bf0.477&\bf0.95&\bf0.269&\bf0.221&\bf0.324&0.029&0.421&0.626&\bf0.49&0.727&\bf0.782&\bf0.62\\
    
    \bottomrule
  \end{tabular}
  }
  \vspace{-0.4in}
\end{table}

\begin{table}
  \setlength\tabcolsep{2.5pt} 
  \centering
  \caption{10\% LR}
  \label{tab:LR_10}
  \scalebox{0.5}{
  \begin{tabular}{l||c|ccccccccccccccccccccc}
    \toprule
    rate & avg.&bathtub&bed&book.&cabinet&	chair&counter&curtain&desk&door&floor& otherf. &picture&frige&shower&	sink&sofa&table&toilet&wall&window \\
    \toprule
    PointContrast &  0.555 & 0.711 & 0.668 & 0.622 & 0.425 & 0.830 & 0.433 & 0.552 & 0.273 & 0.440 & 0.938 & 0.287 & 0.096 & \bf0.470 & 0.576 & \bf0.612 & 0.687 & 0.438 & 0.781 & 0.785 & 0.474 \\
    CSC &  0.575 & 0.671 & \bf0.740 & \bf0.727 & 0.445 & \bf0.847 & 0.380 & 0.602 & 0.512 & 0.447 & 0.942 & \bf0.291 & 0.184 & 0.353 & 0.468 & 0.508 & 0.745 & \bf0.602 & \bf0.855 & 0.765 & 0.420\\
    ViewPointBN & 0.566 & 0.780 & 0.659 & 0.677 & \bf0.484 & 0.799 & 0.419 & 0.636 & 0.480 & 0.432 & 0.940 & 0.238 & 0.124 & 0.396 & \bf0.609 & 0.432 & 0.735 & 0.527 & 0.787 & 0.752 & 0.423 \\
    Ours &\bf0.608& \bf0.853&0.689&0.593&0.483&0.83&\bf0.466&\bf0.652&\bf0.528&\bf0.482&\bf0.954&0.288&\bf0.25&0.448&0.595&0.532&\bf0.748&0.503&0.822&\bf0.806&\bf0.647\\
    
    \bottomrule
  \end{tabular}
  }
  \vspace{-0.4in}
\end{table}

\begin{table}
  \setlength\tabcolsep{2.5pt} 
  \centering
  \caption{20\% LR}
  \label{tab:LR_20}
  \scalebox{0.5}{
  \begin{tabular}{l||c|ccccccccccccccccccccc}
    \toprule
    rate & avg.&bathtub&bed&book.&cabinet&	chair&counter&curtain&desk&door&floor& otherf. &picture&frige&shower&	sink&sofa&table&toilet&wall&window \\
    \toprule
    PointContrast & 0.603 &0.740 & 0.700 &0.700 &0.546 &0.843 &0.419 &0.592 &0.462 &0.513 &0.946 & 0.374 &0.104 &0.530 &0.687 &0.571 &0.694 &0.519 &0.850 &0.781 & 0.484  \\
    CSC &  0.612 &0.739 &\bf0.794 & 0.687 &0.564 &\bf0.850 &0.347 &0.590 &\bf0.587 &0.521 &0.945 &0.358 & 0.140 &0.522 &0.496 &\bf0.627 &0.725 &\bf0.598 &0.850 &0.792 &0.508 \\
    ViewPointBN & 0.625 &\bf0.873 &0.727 &0.709 &0.535 &0.820 &0.402 &0.643 &0.540 &0.501 &0.946 &0.352 &\bf0.181 &0.535 &0.594 &0.596 &0.685 &0.543 &\bf0.927 &0.792 &0.592\\
    Ours &\bf0.663&0.851&0.77&\bf0.76&\bf0.615&0.83&\bf0.439&\bf0.67&0.546&\bf0.587&\bf0.955&\bf0.406&0.177&\bf0.627&\bf0.758&0.606&\bf0.74&0.549&0.888&\bf0.84&\bf0.652 \\
    
    \bottomrule
  \end{tabular}
  }
\end{table}

\paragraph{Limited reconstruction}
We expand the Table 2 in our main paper and show the per-category results on \emph{3D Semantic label with
Limited Reconstructions benchmark} in \cref{tab:LR_1}, \cref{tab:LR_5}, \cref{tab:LR_10} and \cref{tab:LR_20} on 1\%, 5\%, 10\% and 20\% percentage of labeled scene. When only 1\% of the training data is available, all methods achieve similar performance. When 5\%,10\%,20\%
of the training data is given, we outperform the previous state-of-the-art.

\section{Comparison with DepthContrast~\cite{ZhangGJM21}}
\label{sec:depthcontrast}
We use the released code and the pre-trained weight by DepthContrast to train on the efficient ScanNet setup.
We note that DepthContrast uses Minkowski while we use O-CNN as the base model, so we also run the Minkowski without pre-training for a fair comparison.
We present the results of the LR setup in \cref{tab:depth}.
\begin{table}
    \setlength\tabcolsep{2.5pt}
    \centering
    \caption{Comparison our pre-training with DepthContrast~\cite{ZhangGJM21}. The results are presented as mIoU and compared on ScanNet validation set.}
    \label{tab:depth}
    \begin{tabular}{cc|cccc}
    \toprule
    base model & pre-training & LR 1\% & LR 5\% & LR 10\% & LR 20\% \\
    \midrule
    Minkowski & none & 24.8 & 40.5 & 53.5 & 59.8 \\
    Minkowski & \cite{ZhangGJM21} & 27.7 & 43.0 & 56.8 & 61.8 \\
    \hline
    O-CNN & none & 18.7 & 39.7 & 52.4 & 59.6 \\
    O-CNN & ours & 26.3 & 44.9 & 56.5 & 62.7 \\
    \bottomrule
    \end{tabular}
    \vspace{-0.5in}
\end{table}

Without any pre-training, Minkowski outperforms O-CNN.
However, O-CNN with our pre-training can overtake Minkowski with DepthContrast pre-training in LR 5\% and LR 20\%, and achieve similar mIoU in LR 10\%.
The results show that the proposed pre-training can boost better than the previous work in the LR setup.

\section{Additional implement details}
\label{sec:imple detail}
In semi-supervised learning, the loss of unlabeled data are weighted sum by different coefficients across downstream tasks. For object classification, the weights of mini-entropy loss and consistency loss referenced in the main paper are 0.01 and 10. For scene semantic segmentation, the weights of mini-entropy loss and consistency loss are 0.25 and 10. Among Mean-Teacher training, the weights of the teacher model were updated each training step by EMA with smoothing hyperparameter $\alpha$ = 0.999, and the coefficient of consistency cost will ramp up during the first 30 epochs by a sigmoid-shaped function $e^{-5(1-x)^{2}}$, where $x \in \left[0, 1\right]$. 

\section{Statistic of the pre-training pseudo-labels}
\label{sec:softlabel}
\begin{figure}
	\centering
	\caption{The histogram for the summed probability of the pre-training pseudo-label. The x-axis are the 150 different classes of a image scene parse, which is DPT in this work, well-trained on the ADE20k dataset.}
	\includegraphics[width=\textwidth,trim={0 75 0 175},clip]{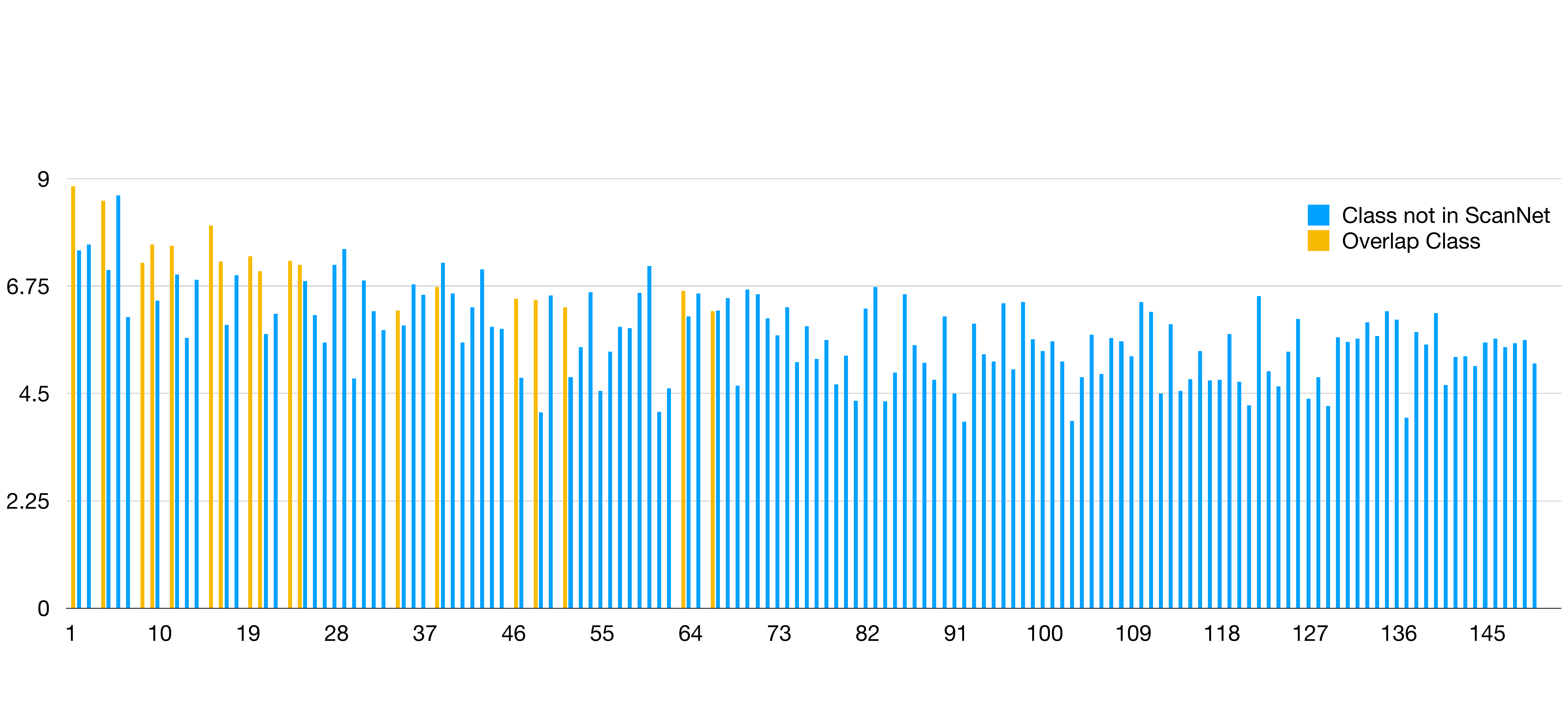}
	\label{fig:probability}
	\vspace{-0.3in}
\end{figure}

We adopt DPT~\cite{RanftlBK21}, which has been trained on the ADE20k~\cite{ZhouZPFB017} dataset, to predict the categorical probability distribution over 150 classes for all pixels.
We sum the predicted probability of pixels in all images in Matterport3D~\cite{ChangDFHNSSZZ17} dataset for each class.
The histogram of the summed pseudo-label probability are presented in \cref{fig:probability}.
The horizontal axis represents the 150 different classes in ADE20k and the vertical axis shows the log of summed probabilities, then the yellow bars are the overlap classes for ADE20k and ScanNet.
As shown in \cref{fig:probability}, the pseudo-labels for pre-training provide extensive knowledge beyond the target ScanNet dataset and is non-trivial to learn. 

\newpage

\section{Qualitative results of data efficient on Scannet}
\label{sec:qualitative}
\begin{figure}
  \centering
  \caption{Qualitative results for ScanNet LA}
  \label{fig:scene1}
  \renewcommand{\arraystretch}{0.4}
  \scalebox{1}{
  \begin{tabular}{l|cccc}
    \toprule
    & \multirow{2}{*}{20} & \multirow{2}{*}{50} & \multirow{2}{*}{100} & \multirow{2}{*}{200} \\
    & \\
    \midrule
    \midrule
    \includegraphics[width=0.19\linewidth, height=0.11\textheight]{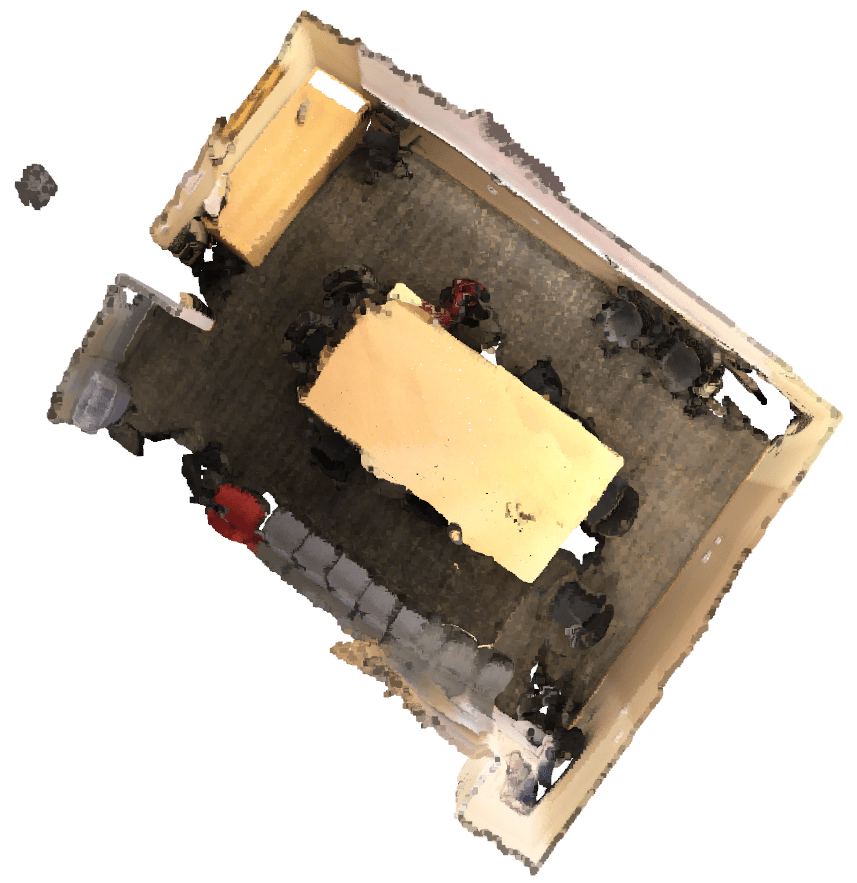}
    &\includegraphics[width=0.19\linewidth, height=0.11\textheight]{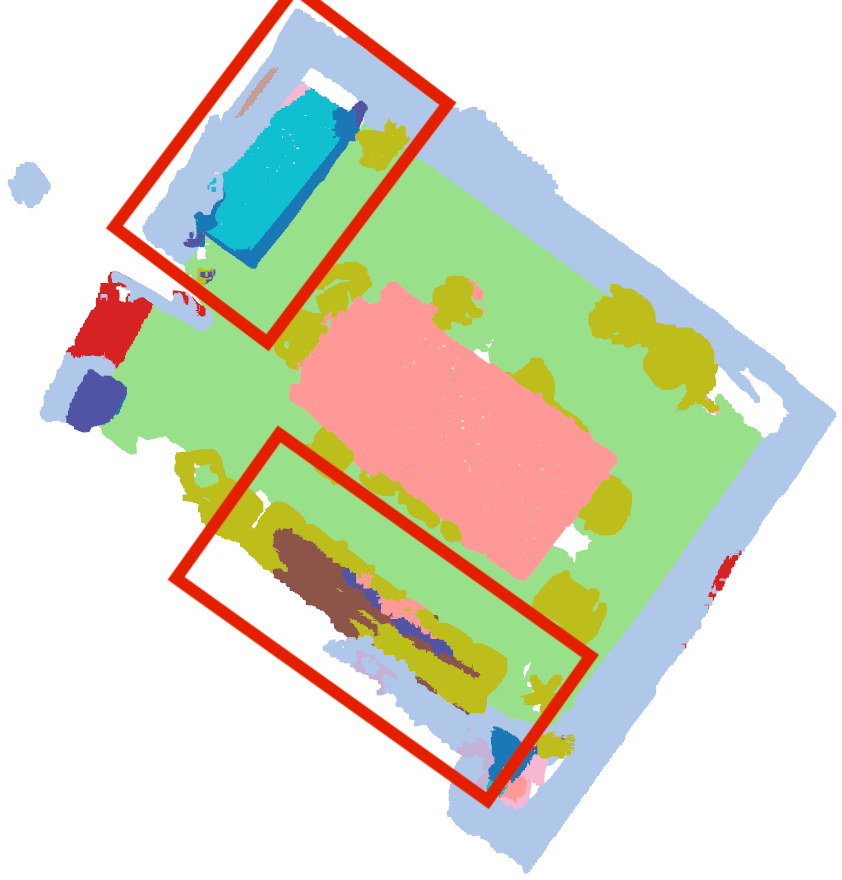}
    &\includegraphics[width=0.19\linewidth, height=0.11\textheight]{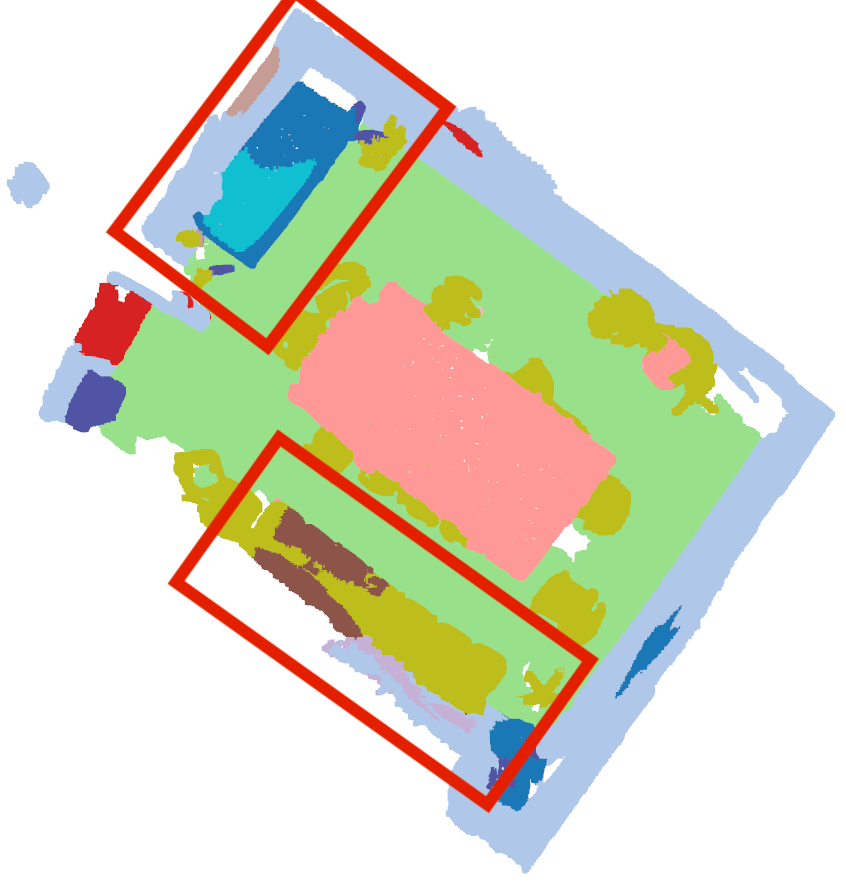}
    &\includegraphics[width=0.19\linewidth, height=0.11\textheight]{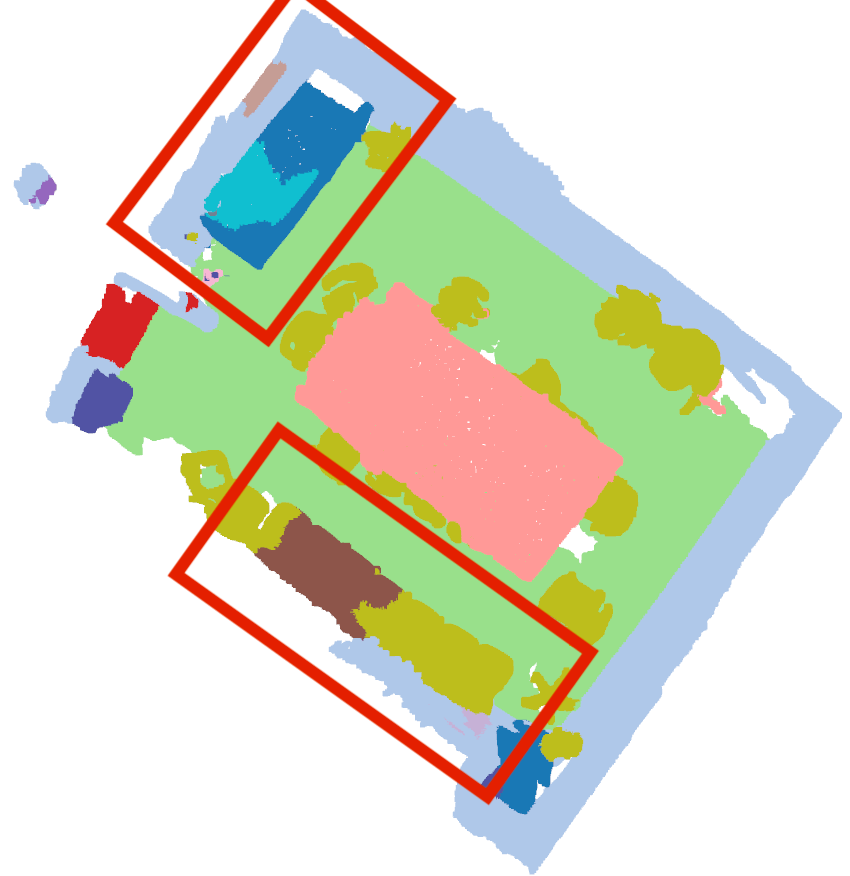}
    &\includegraphics[width=0.19\linewidth, height=0.11\textheight]{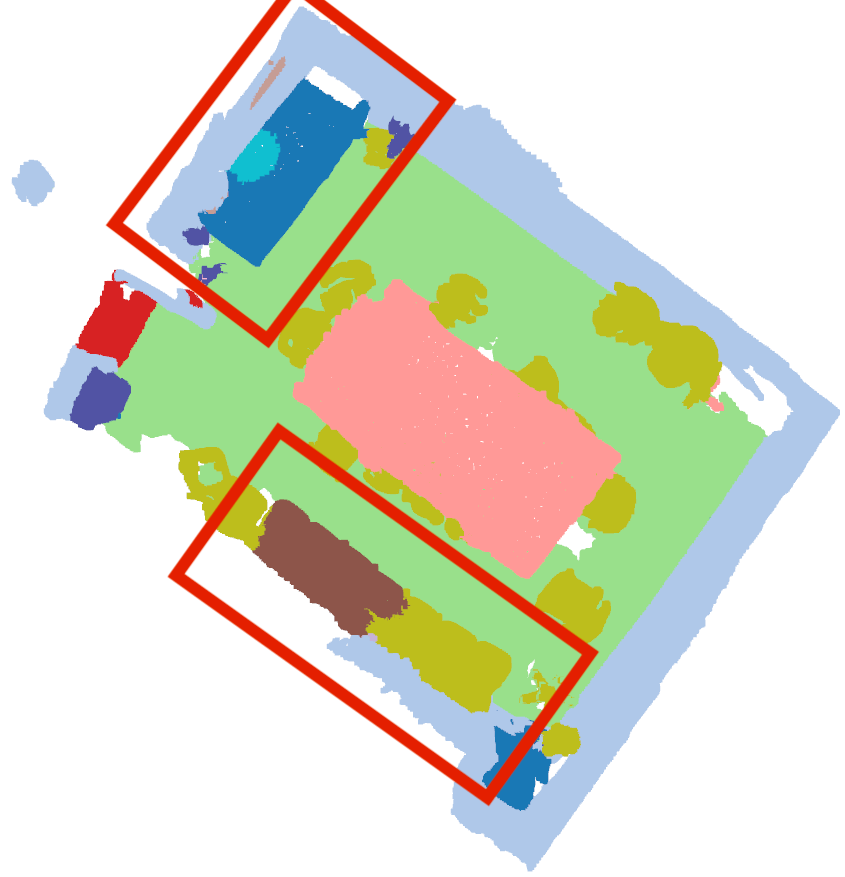} \\
    Input RGB & \multicolumn{4}{c}{Trained from scratch} \\
    \midrule
    \includegraphics[width=0.19\linewidth, height=0.11\textheight]{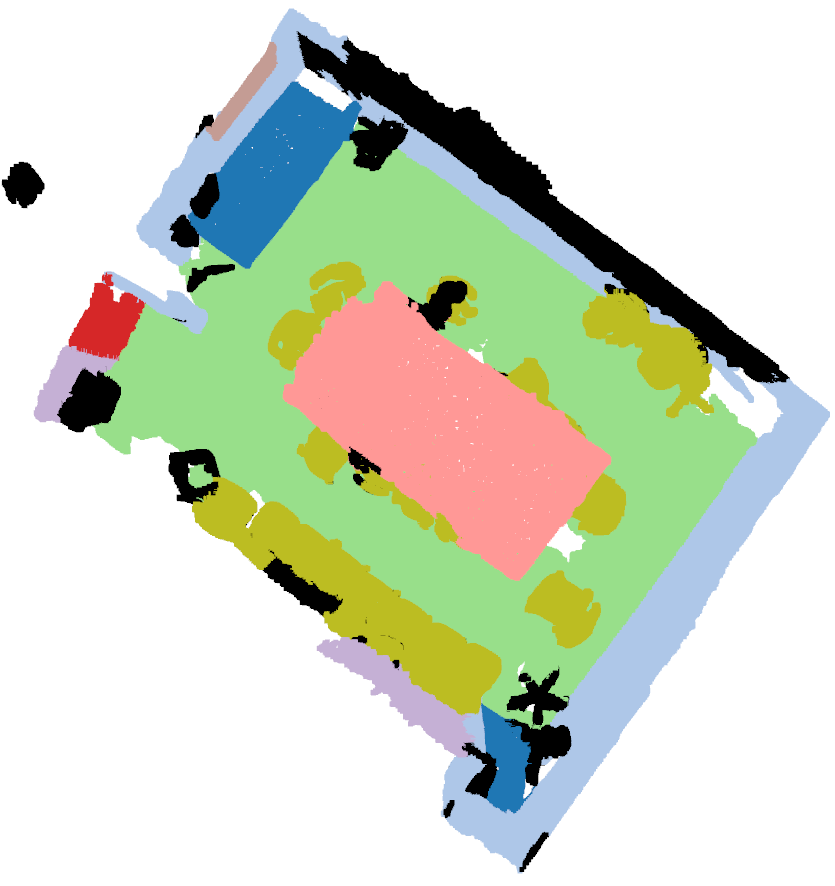}
    & \includegraphics[width=0.19\linewidth, height=0.11\textheight]{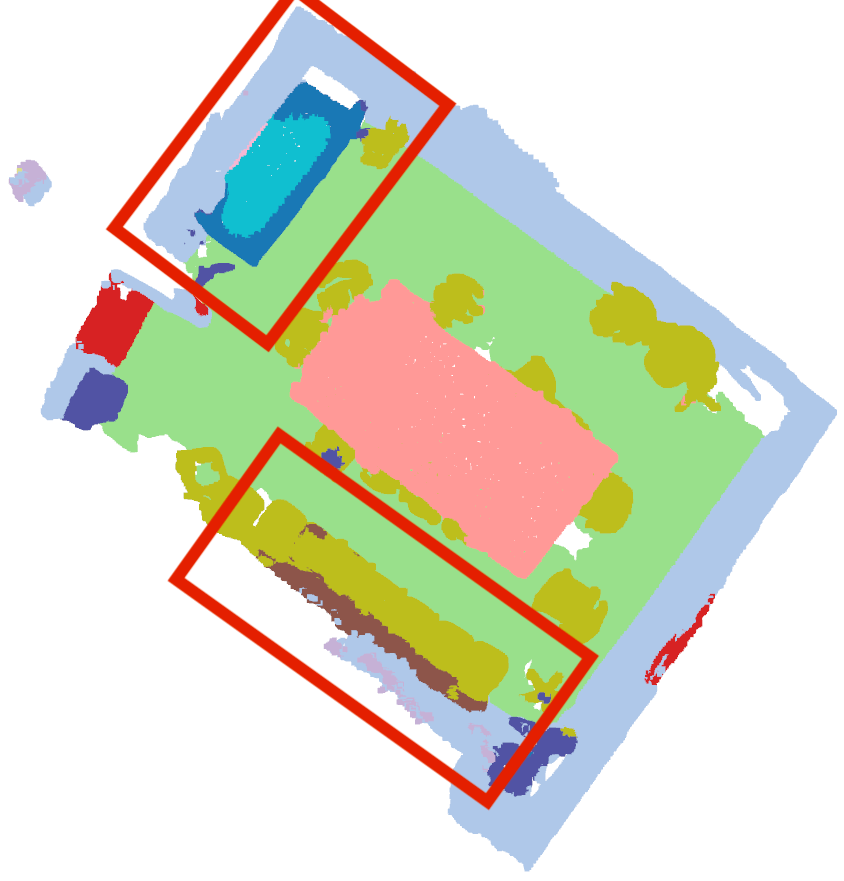}
    & \includegraphics[width=0.19\linewidth, height=0.11\textheight]{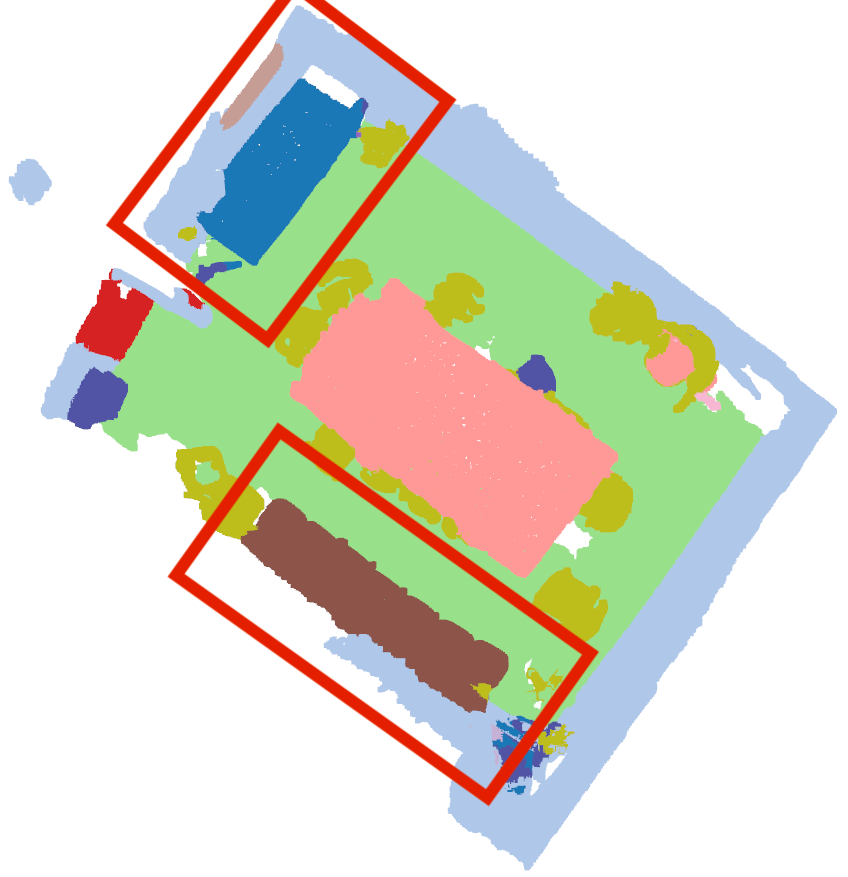}
    & \includegraphics[width=0.19\linewidth, height=0.11\textheight]{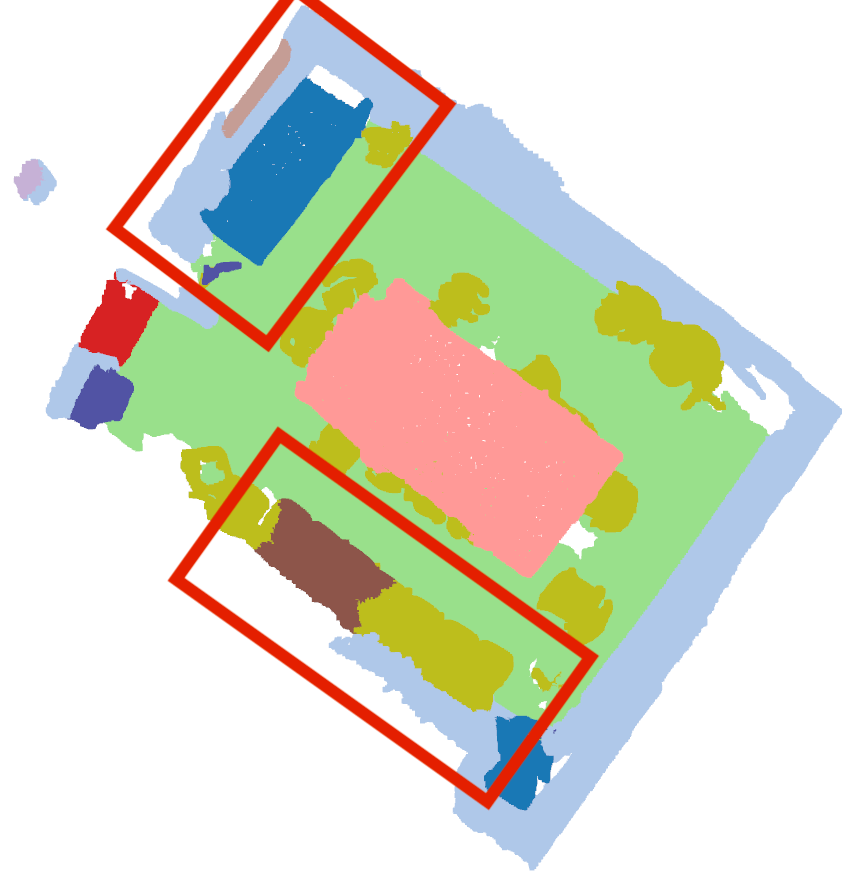} 
    & \includegraphics[width=0.19\linewidth, height=0.11\textheight]{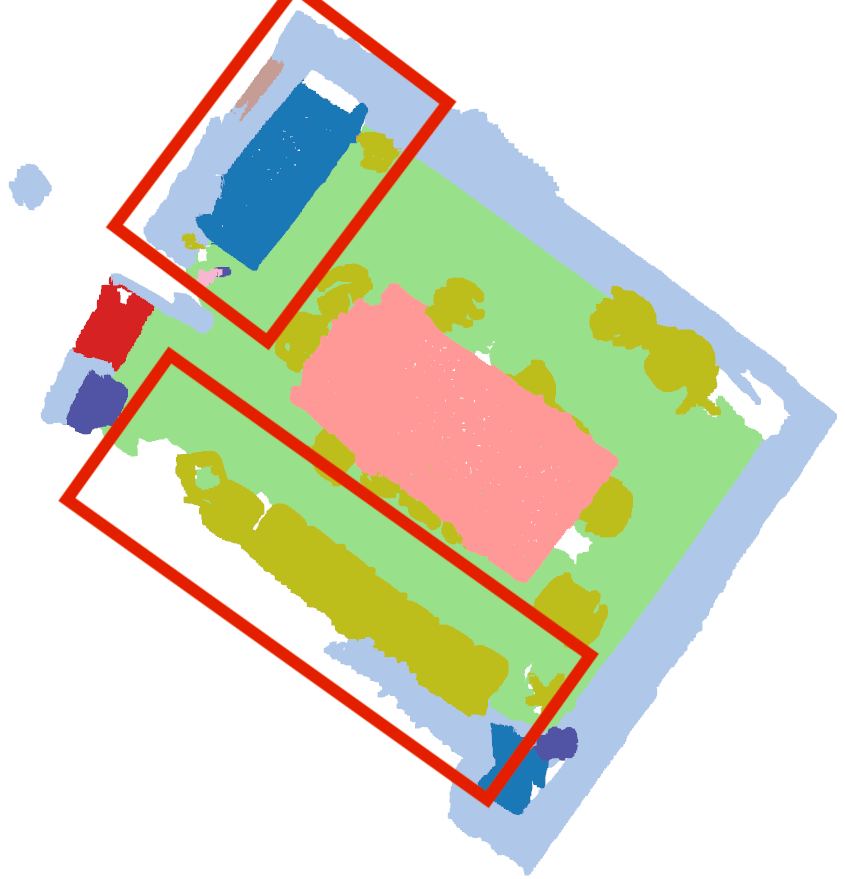}\\
    Ground-truth & \multicolumn{4}{c}{Pre-training} \\
    \bottomrule
  \end{tabular}
  }
\end{figure}

\begin{figure}
  \centering
  \caption{Qualitative results for ScanNet LA}
  \label{fig:scene234}
  \renewcommand{\arraystretch}{0.4}
  \scalebox{1}{
  \begin{tabular}{l|cccc}
    \toprule
    & \multirow{2}{*}{20} & \multirow{2}{*}{50} & \multirow{2}{*}{100} & \multirow{2}{*}{200} \\
    & \\
    \midrule
    \midrule
    \includegraphics[width=0.19\linewidth]{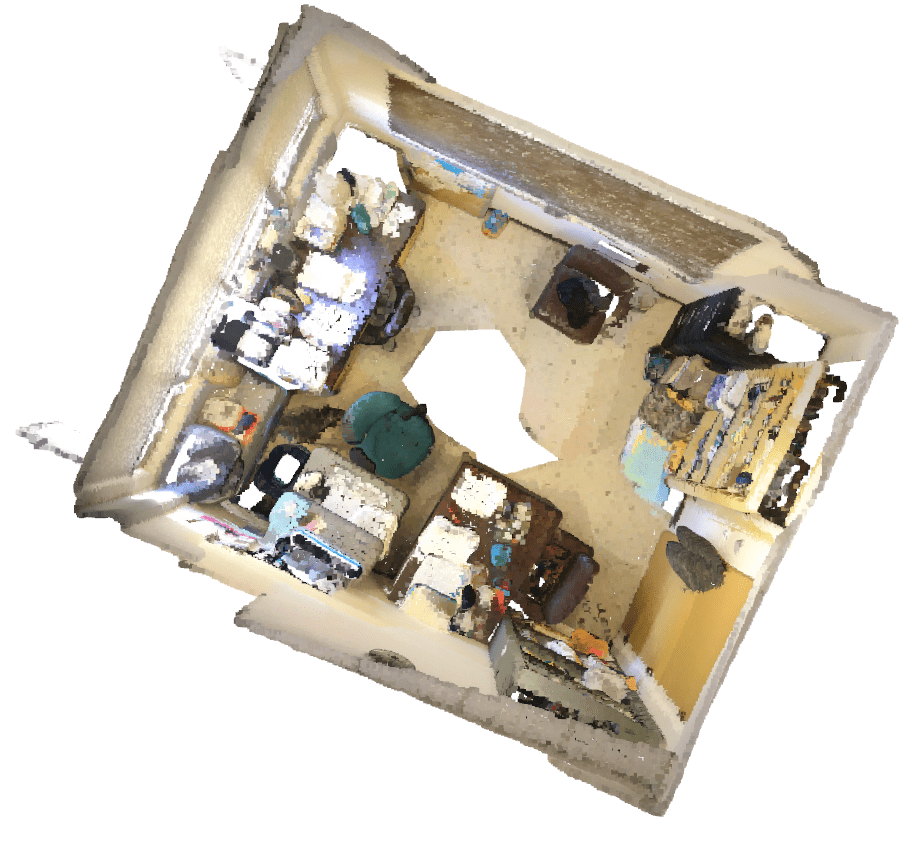}
    &\includegraphics[width=0.19\linewidth]{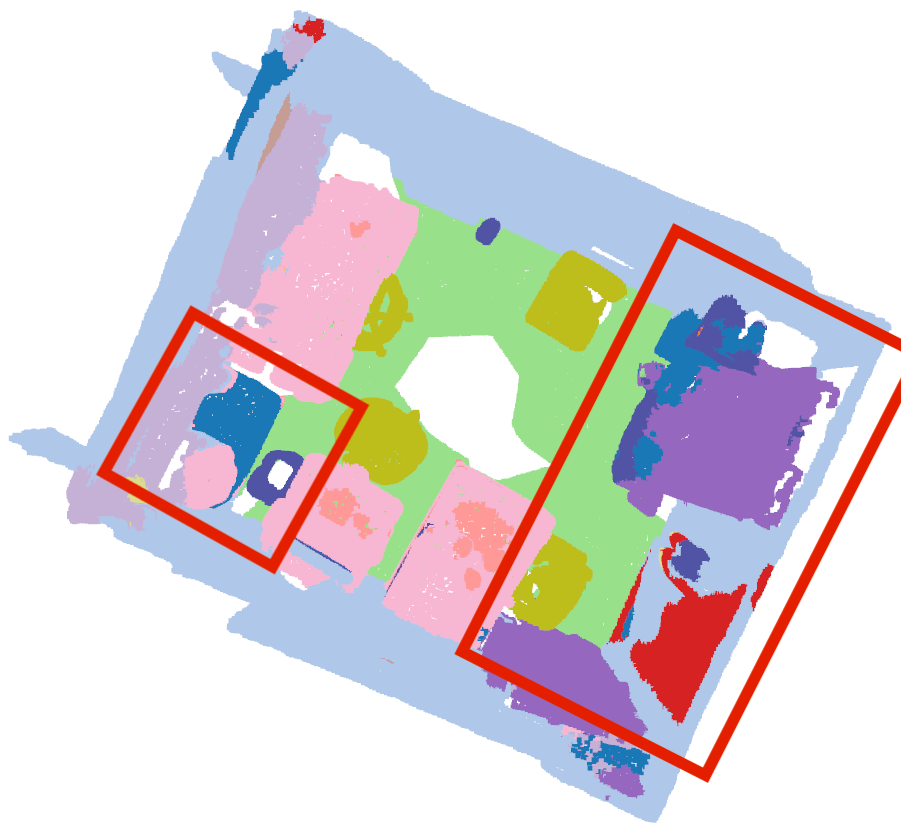}
    &\includegraphics[width=0.19\linewidth]{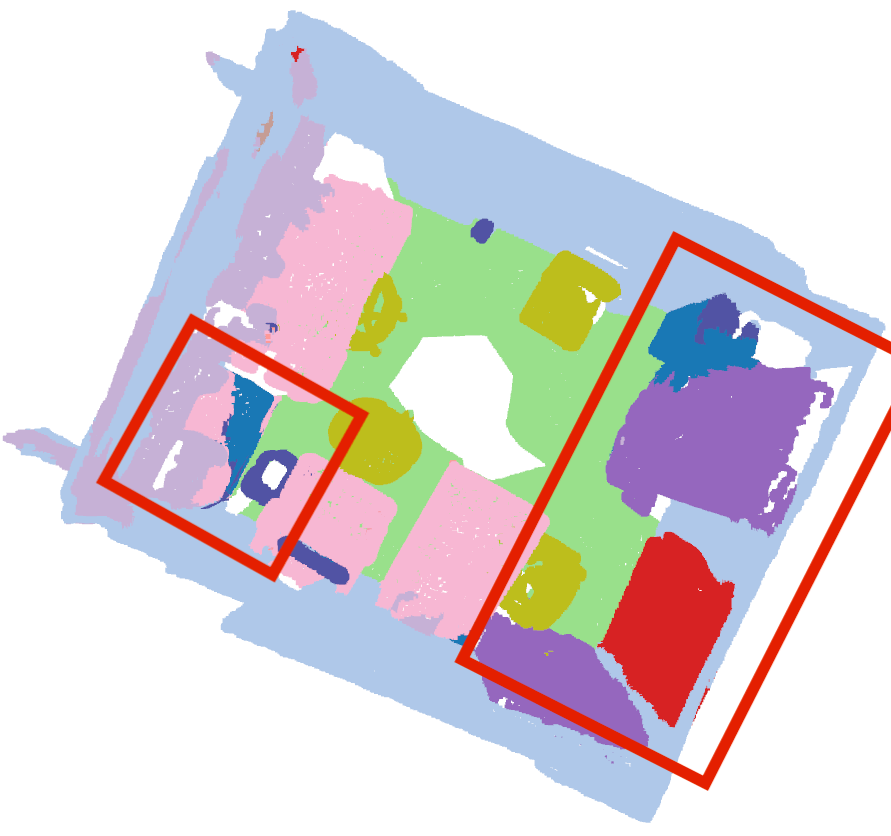}
    &\includegraphics[width=0.19\linewidth]{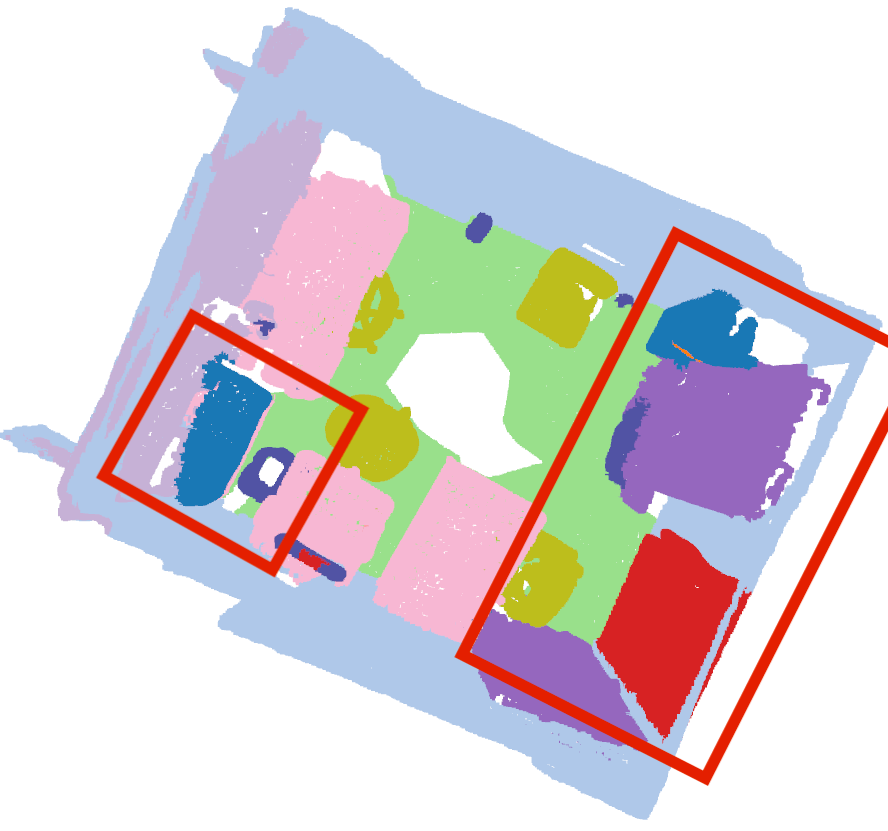}
    &\includegraphics[width=0.19\linewidth]{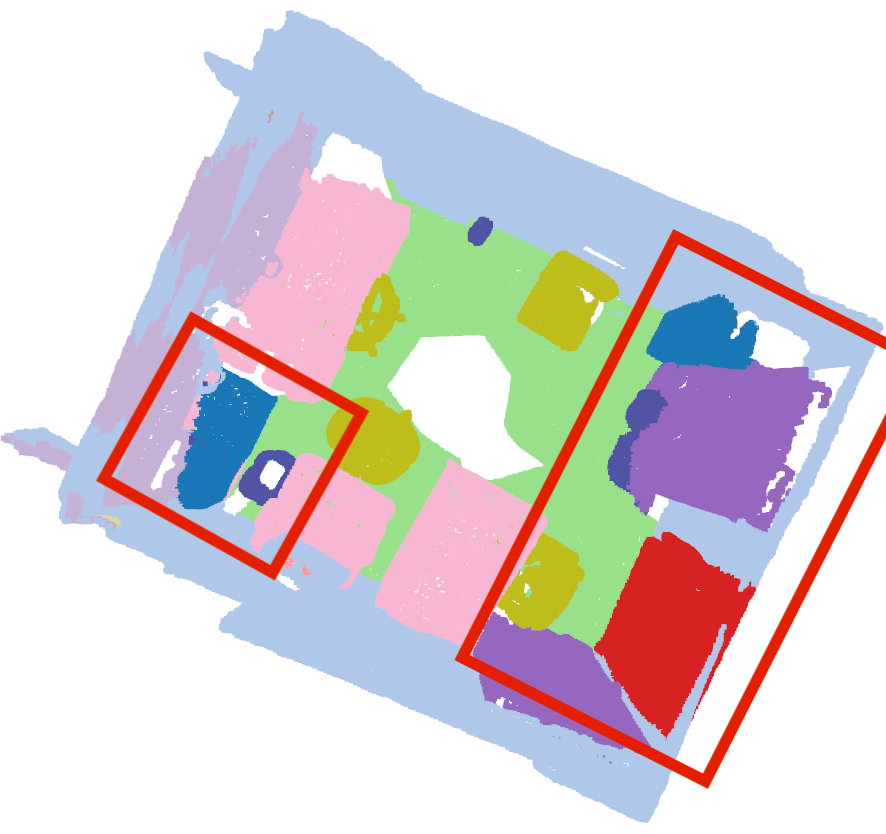} \\
    Input RGB & \multicolumn{4}{c}{Trained from scratch} \\
    \midrule
    \includegraphics[width=0.19\linewidth]{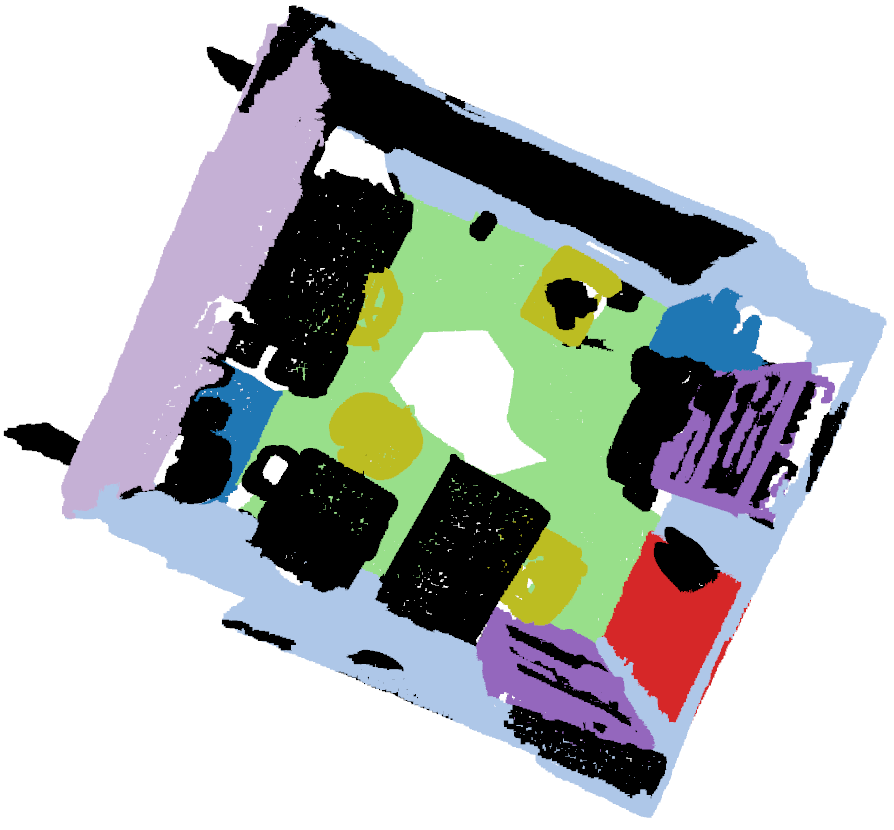}
    & \includegraphics[width=0.19\linewidth]{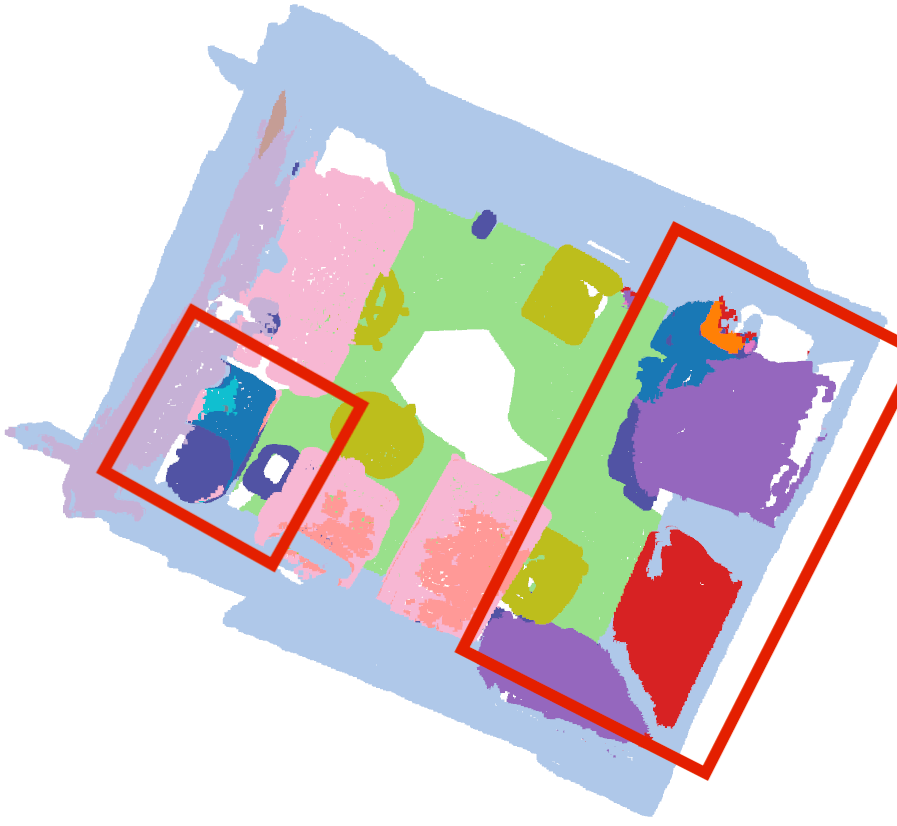}
    & \includegraphics[width=0.19\linewidth]{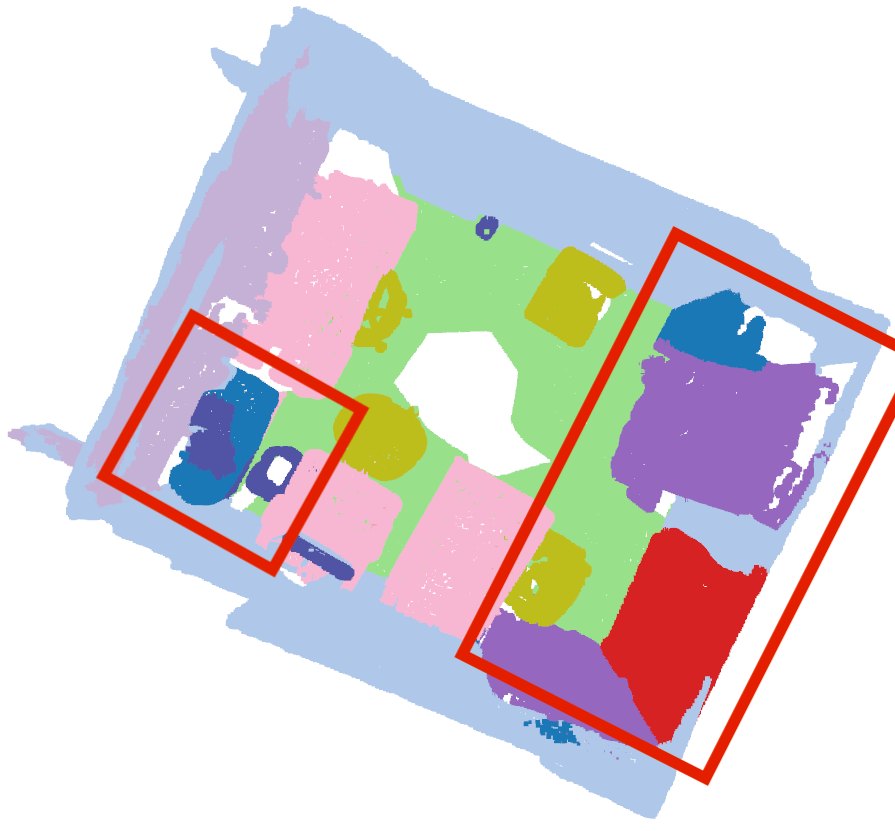}
    & \includegraphics[width=0.19\linewidth]{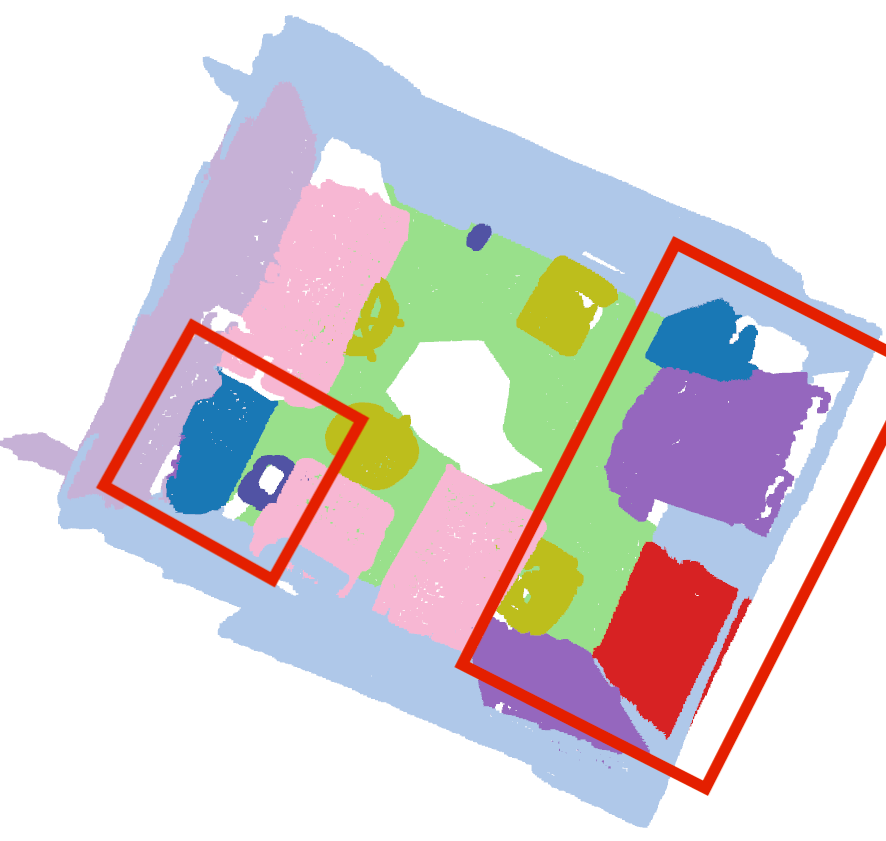} 
    & \includegraphics[width=0.19\linewidth]{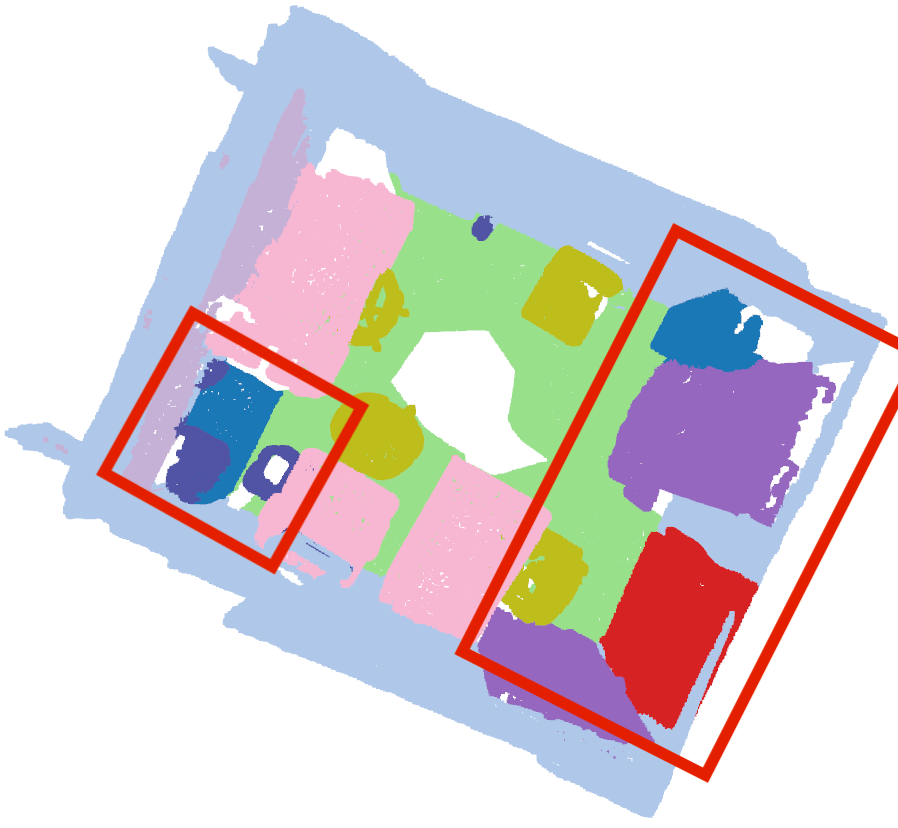}\\
    Ground-truth & \multicolumn{4}{c}{Pre-training} \\
    \midrule\noalign{\vskip .25em}
    \midrule
    \includegraphics[width=0.19\linewidth]{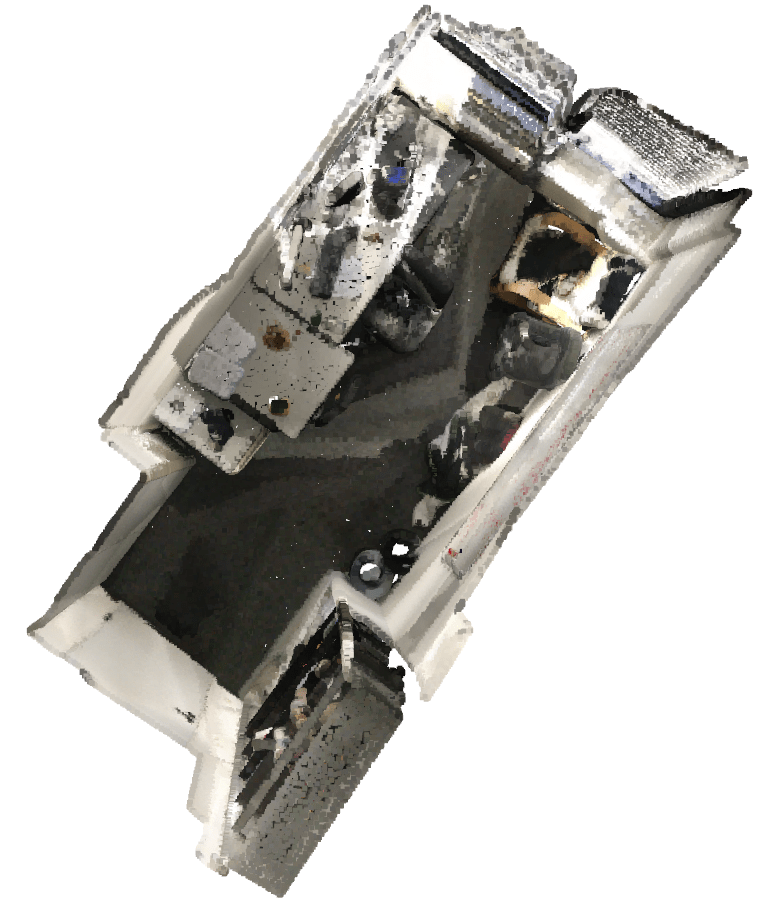}
    &\includegraphics[width=0.19\linewidth]{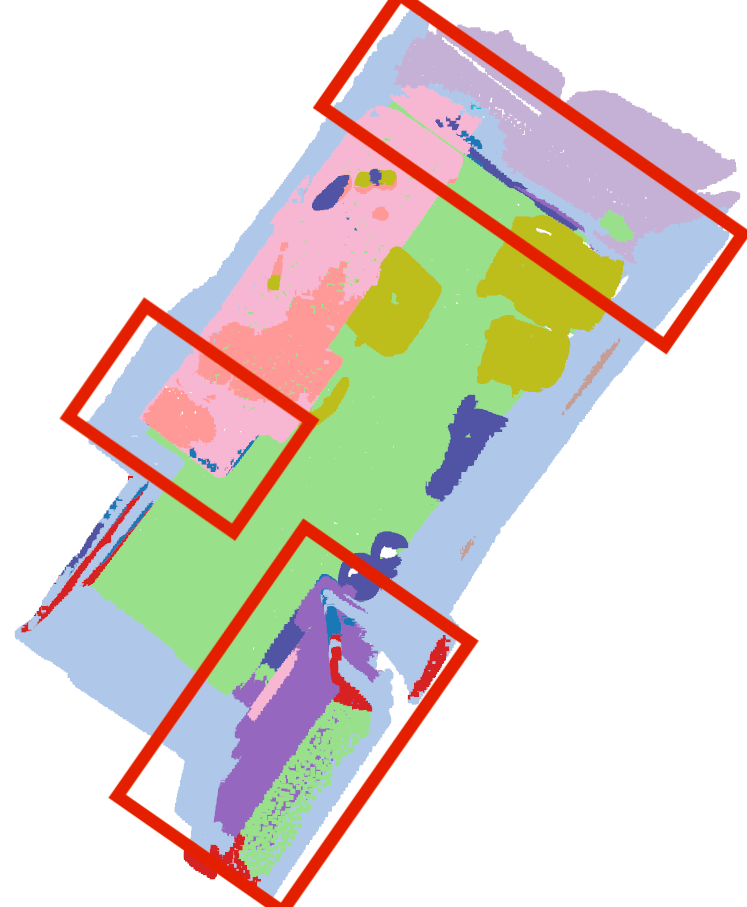}
    &\includegraphics[width=0.19\linewidth]{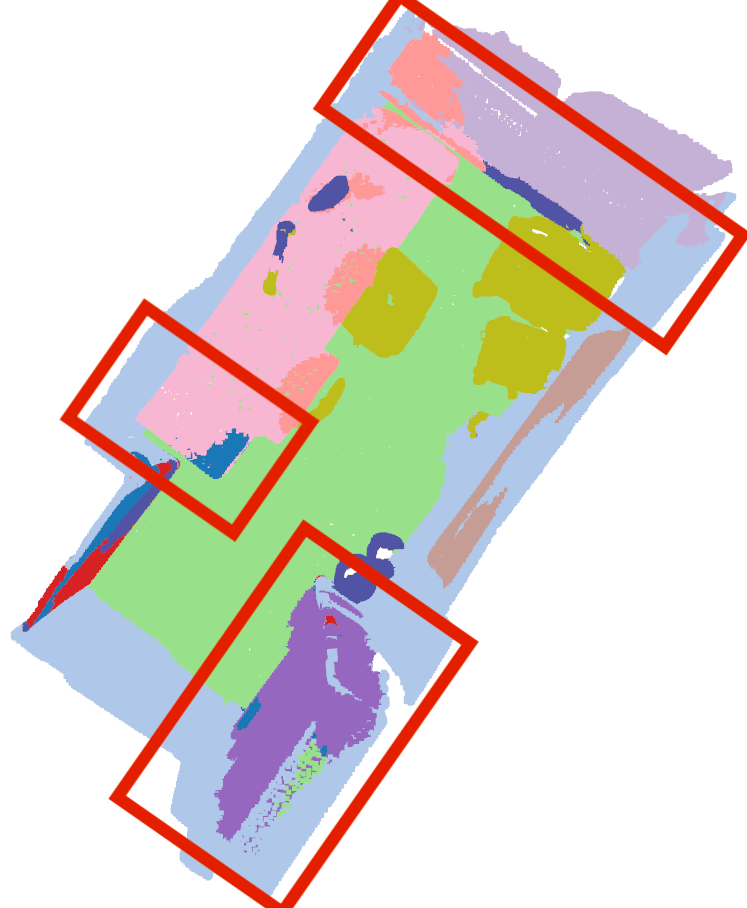}
    &\includegraphics[width=0.19\linewidth]{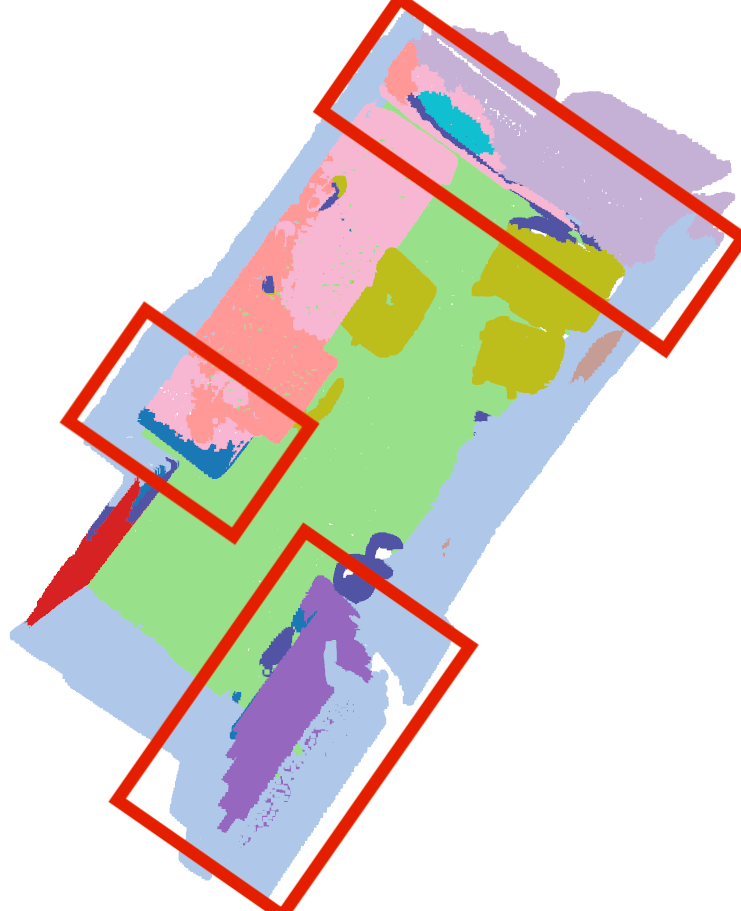}
    &\includegraphics[width=0.19\linewidth]{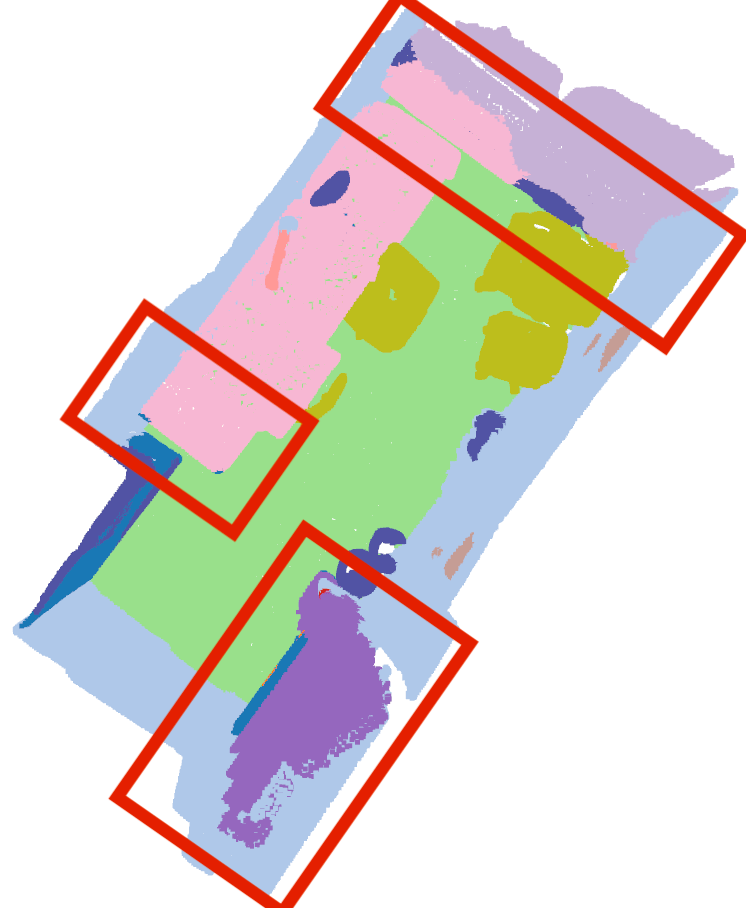} \\
    Input RGB & \multicolumn{4}{c}{Trained from scratch} \\
    \midrule
    \includegraphics[width=0.19\linewidth]{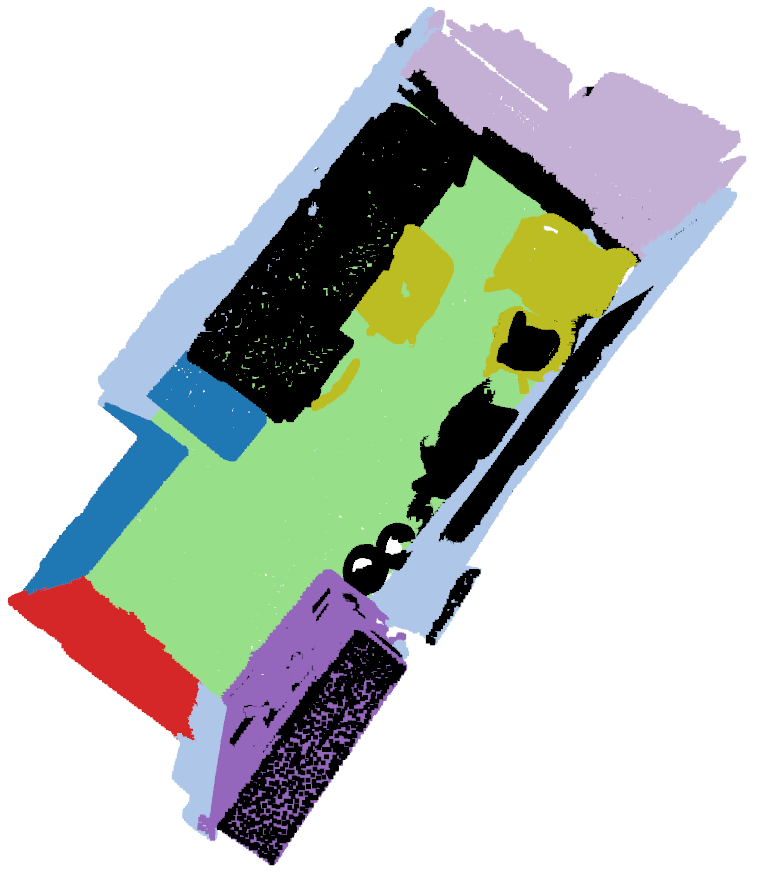}
    & \includegraphics[width=0.19\linewidth]{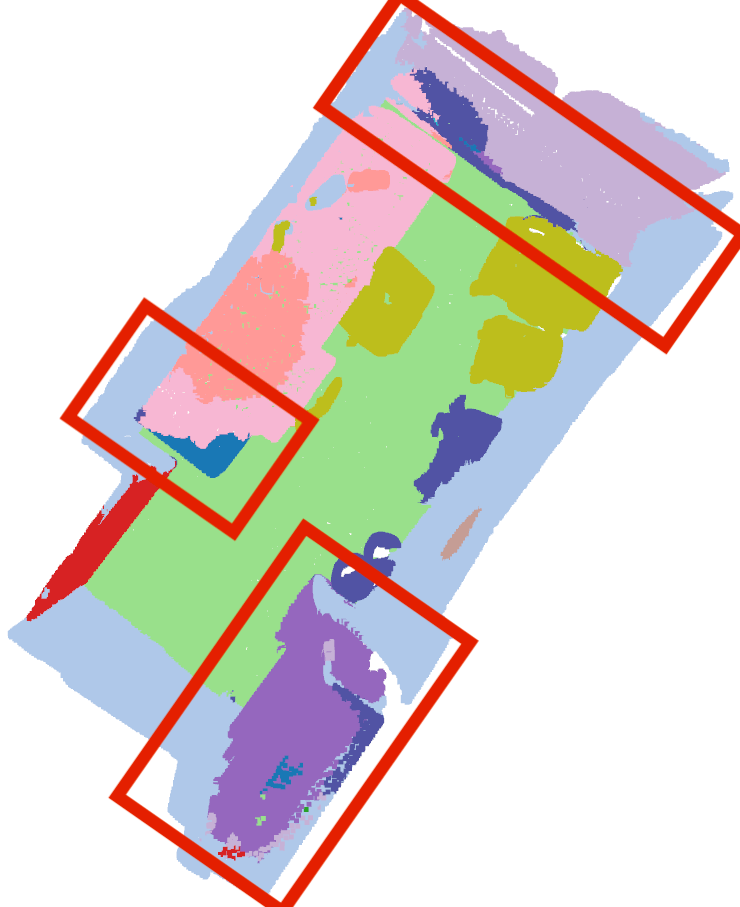}
    & \includegraphics[width=0.19\linewidth]{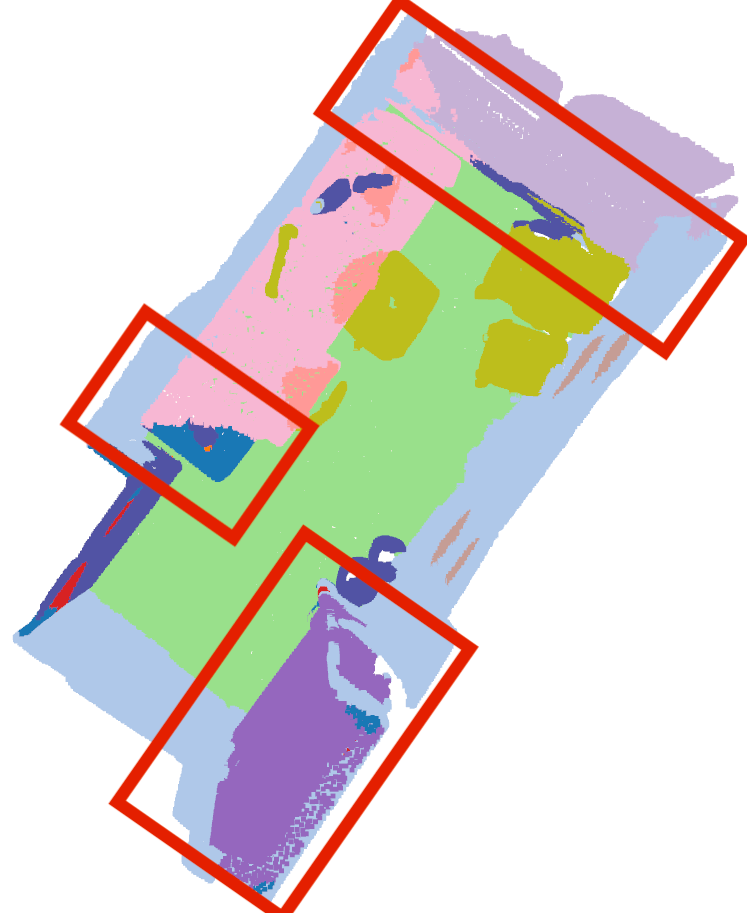}
    & \includegraphics[width=0.19\linewidth]{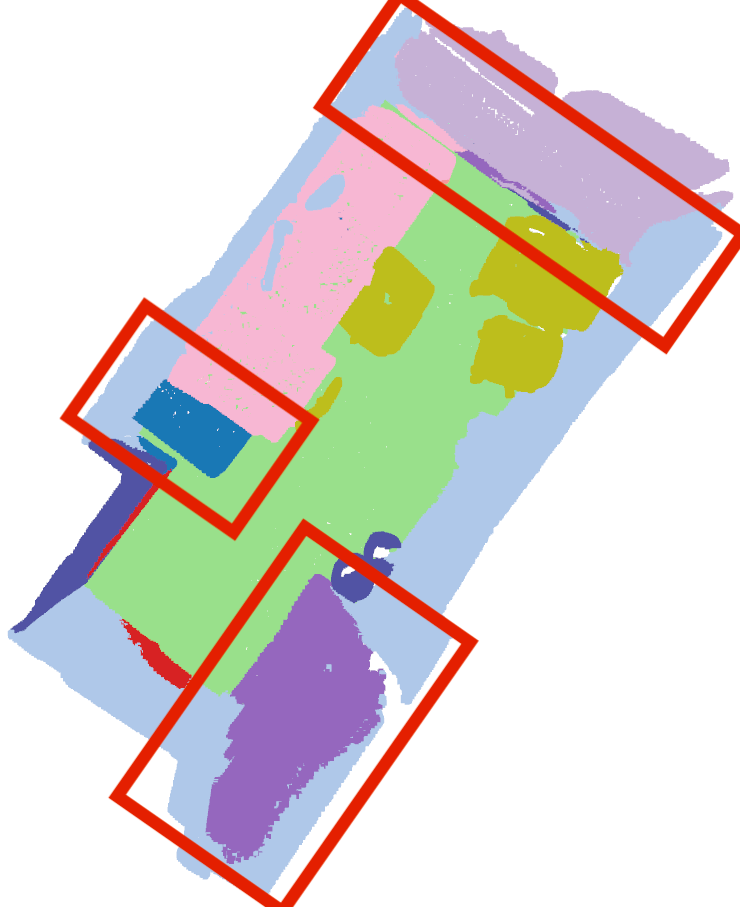} 
    & \includegraphics[width=0.19\linewidth]{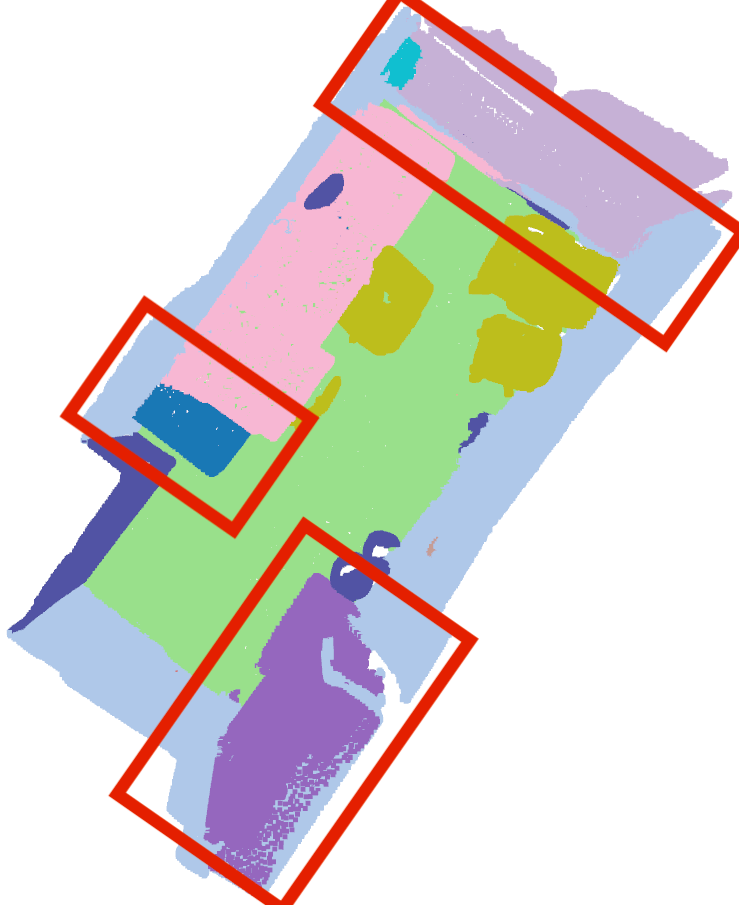}\\
    Ground-truth & \multicolumn{4}{c}{Pre-training} \\
    \midrule\noalign{\vskip .25em}
    \midrule
    \includegraphics[width=0.19\linewidth]{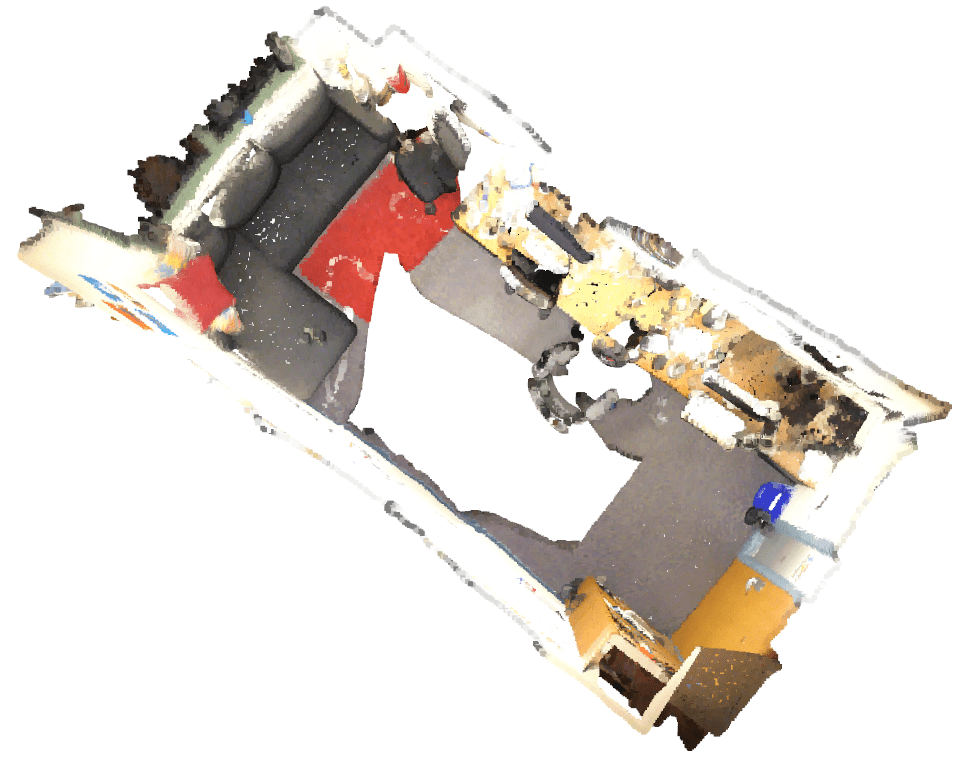}
    &\includegraphics[width=0.19\linewidth]{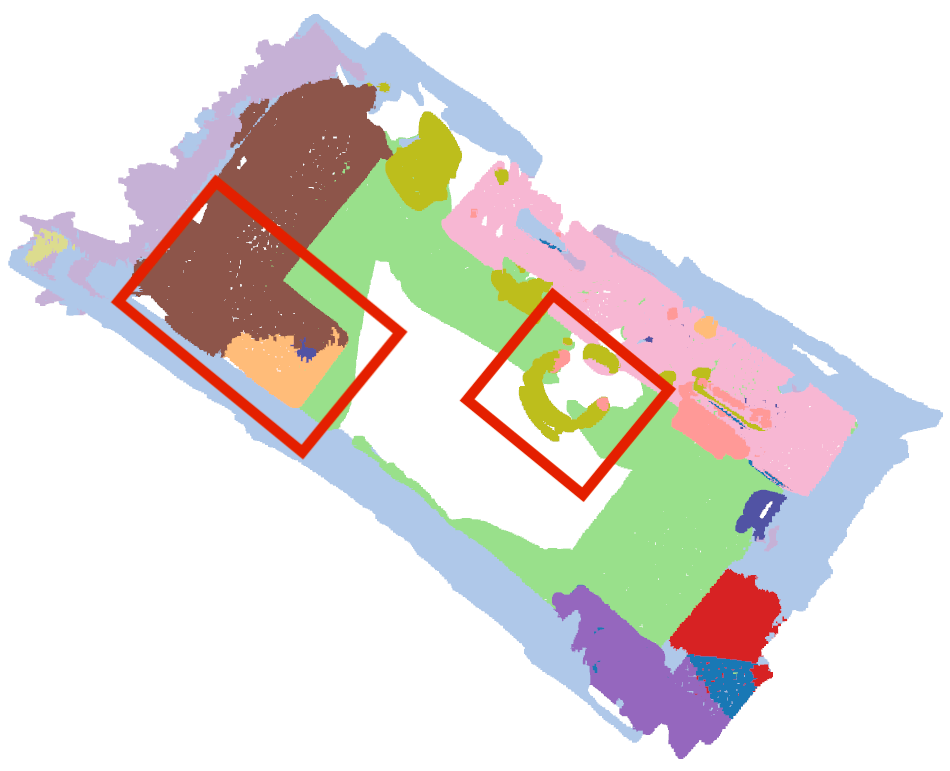}
    &\includegraphics[width=0.19\linewidth]{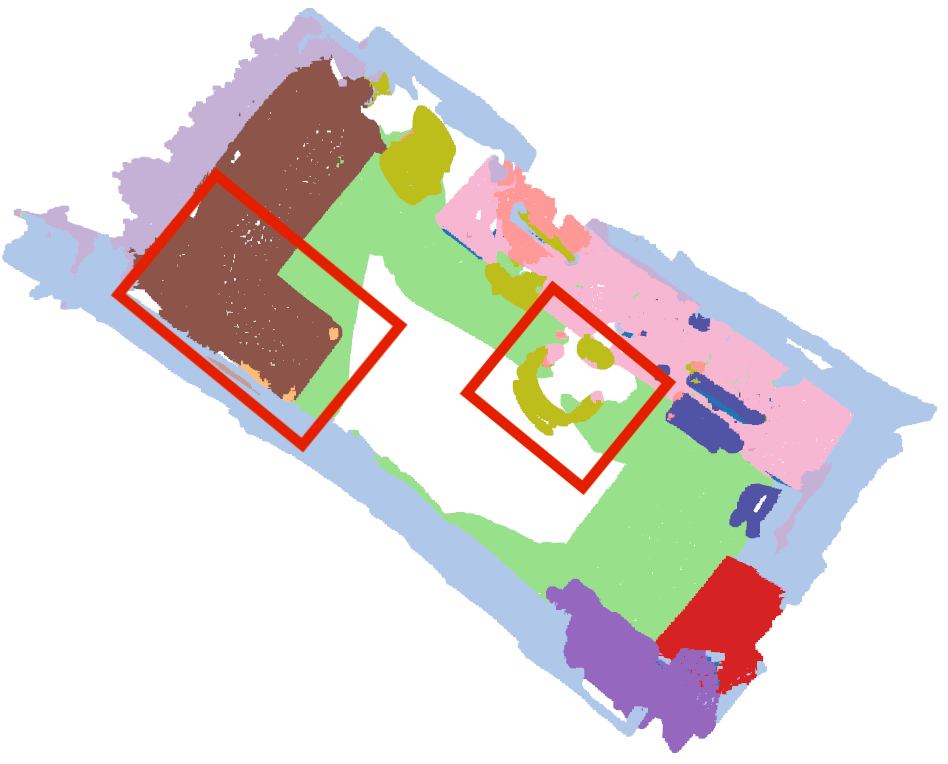}
    &\includegraphics[width=0.19\linewidth]{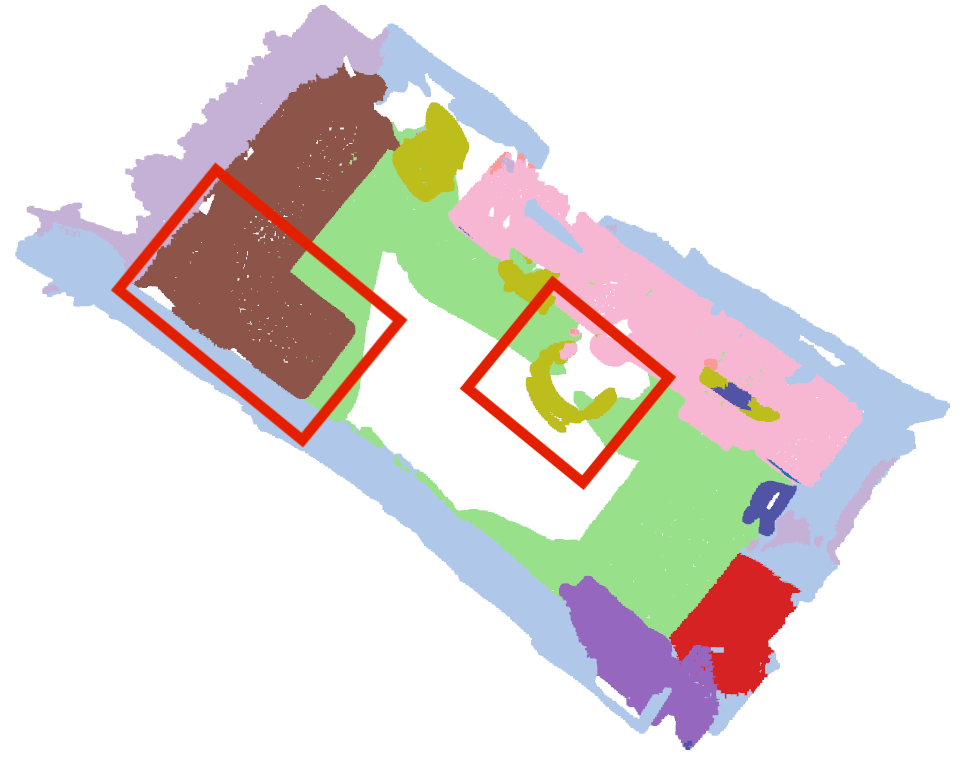}
    &\includegraphics[width=0.19\linewidth]{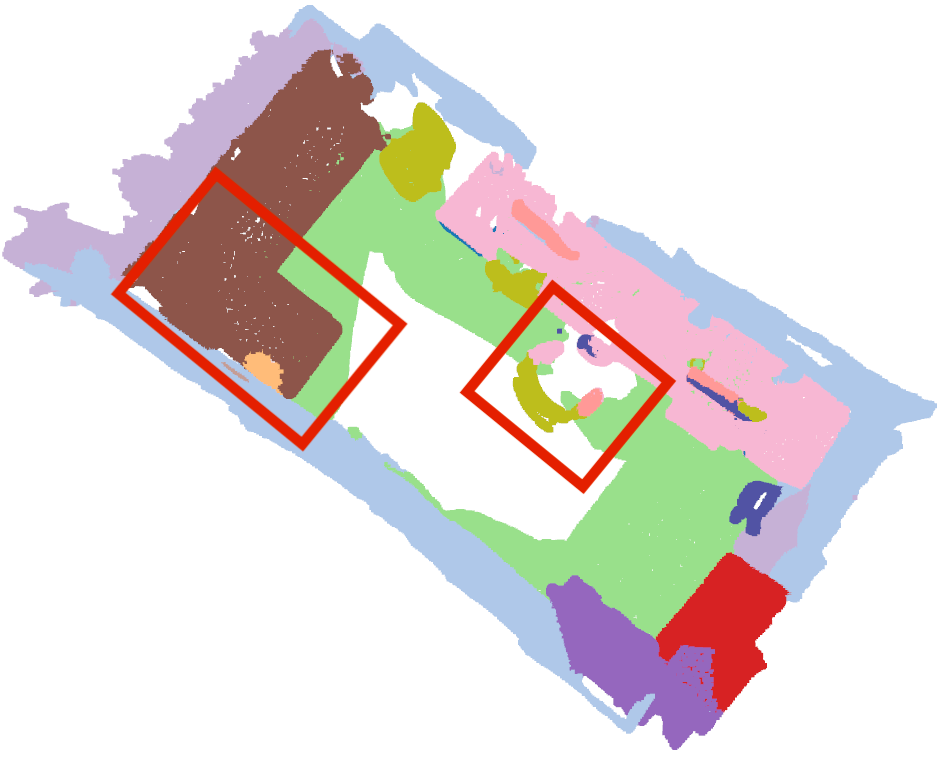} \\
    Input RGB & \multicolumn{4}{c}{Trained from scratch} \\
    \midrule
    \includegraphics[width=0.19\linewidth]{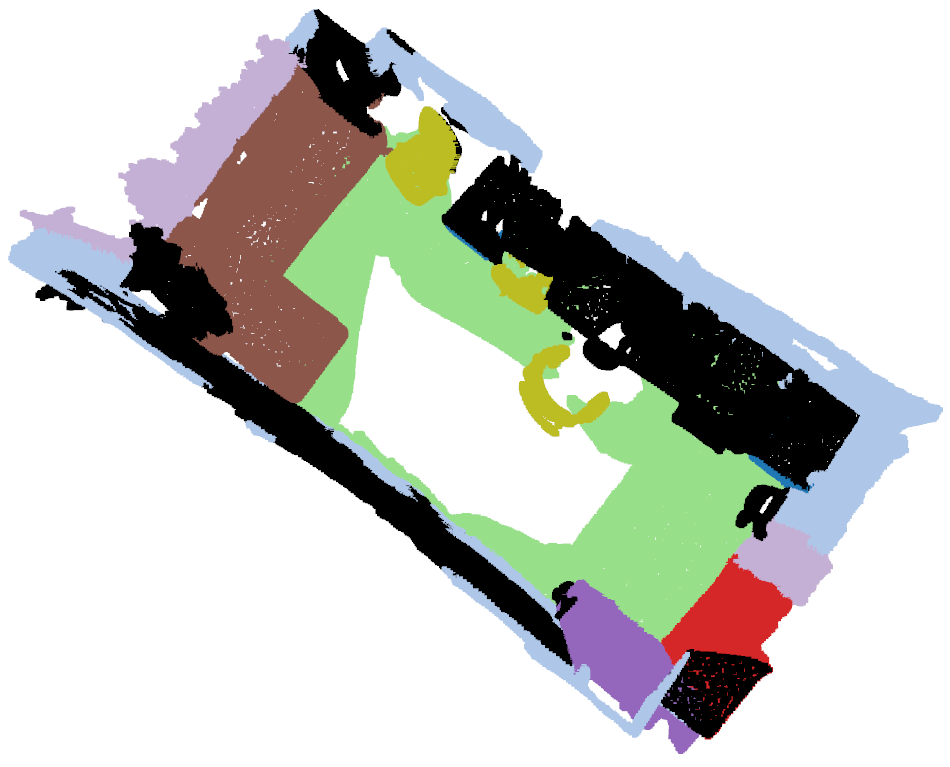}
    & \includegraphics[width=0.19\linewidth]{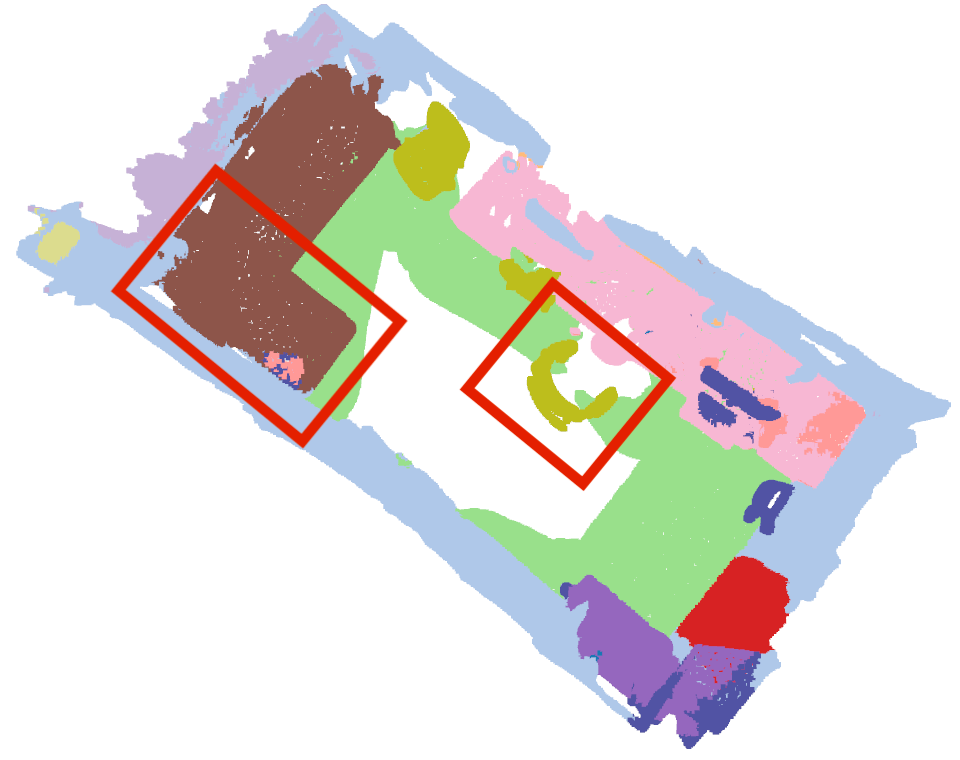}
    & \includegraphics[width=0.19\linewidth]{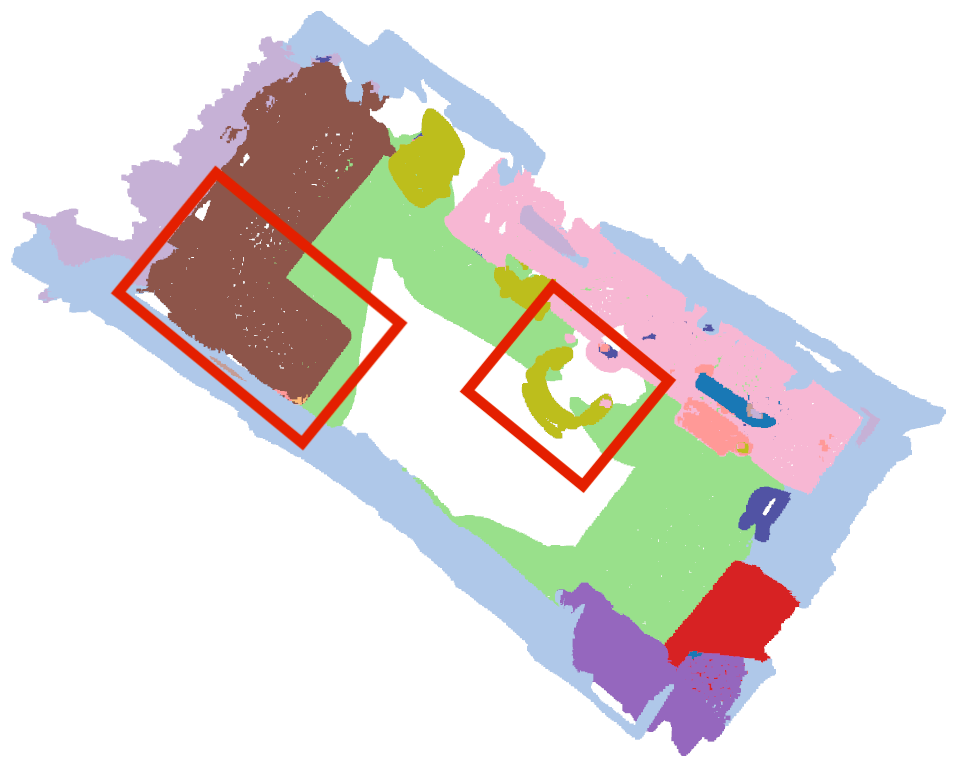}
    & \includegraphics[width=0.19\linewidth]{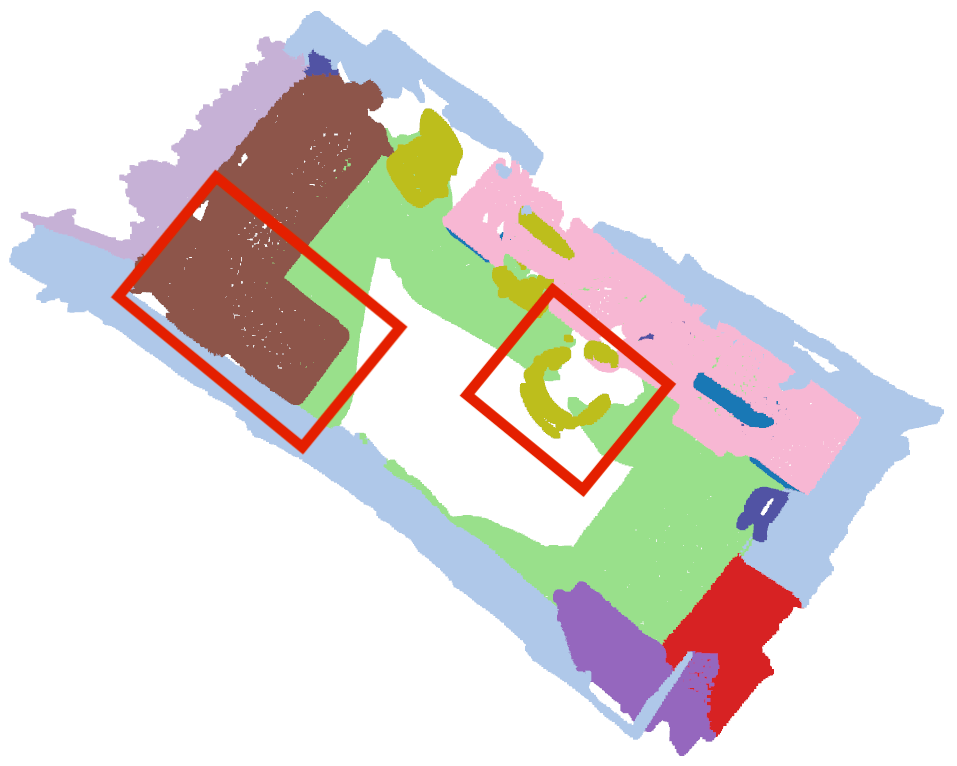} 
    & \includegraphics[width=0.19\linewidth]{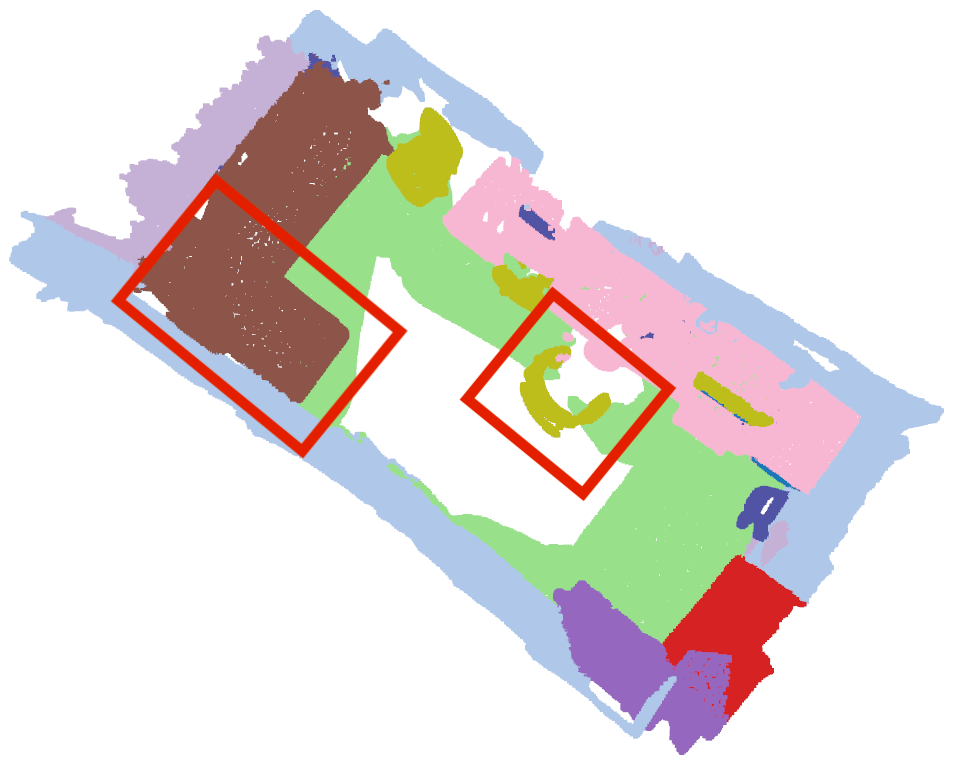}\\
    Ground-truth & \multicolumn{4}{c}{Pre-training} \\
    \bottomrule
  \end{tabular}
  }
\end{figure}

\begin{figure}
  \centering
  \caption{Qualitative results for ScanNet LR}
  \label{fig:LRscene123}
  \renewcommand{\arraystretch}{0.4}
  \scalebox{1}{
  \begin{tabular}{l|cccc}
    \toprule
    & \multirow{2}{*}{1\%} & \multirow{2}{*}{5\%} & \multirow{2}{*}{10\%} & \multirow{2}{*}{20\%} \\
    & \\
    \midrule
    \midrule
    \includegraphics[width=0.19\linewidth]{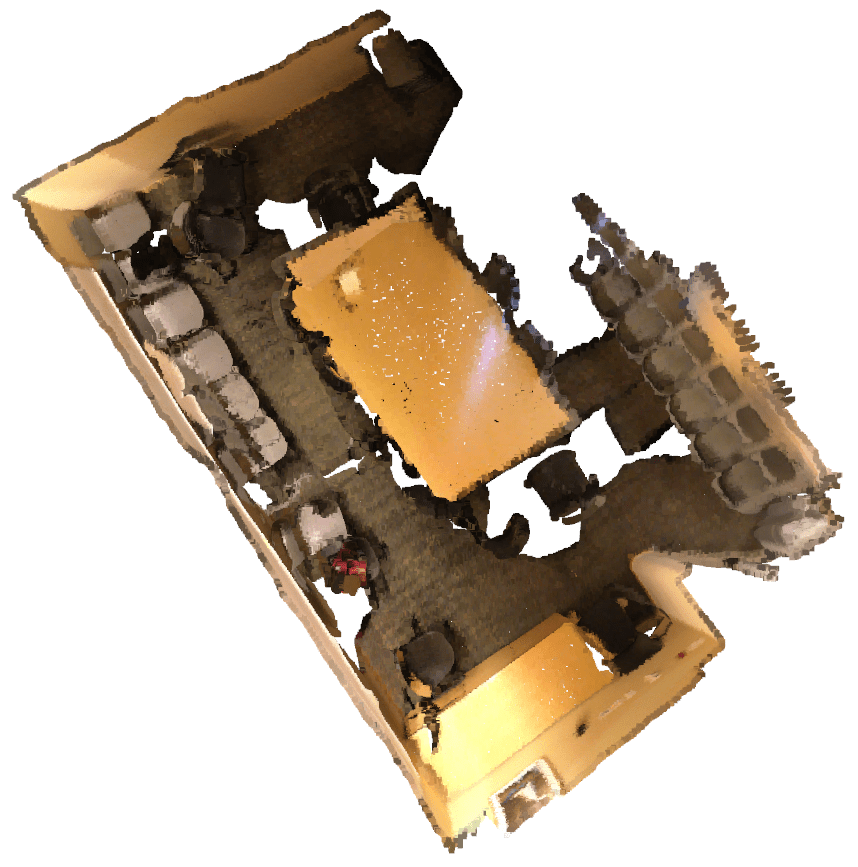}
    &\includegraphics[width=0.19\linewidth]{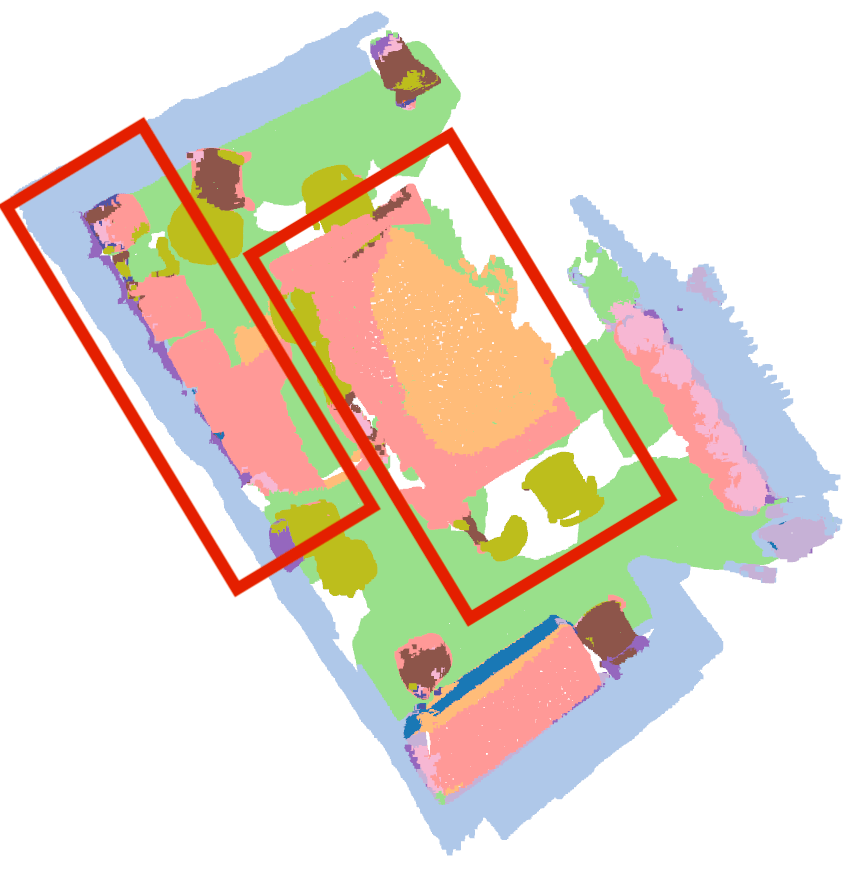}
    &\includegraphics[width=0.19\linewidth]{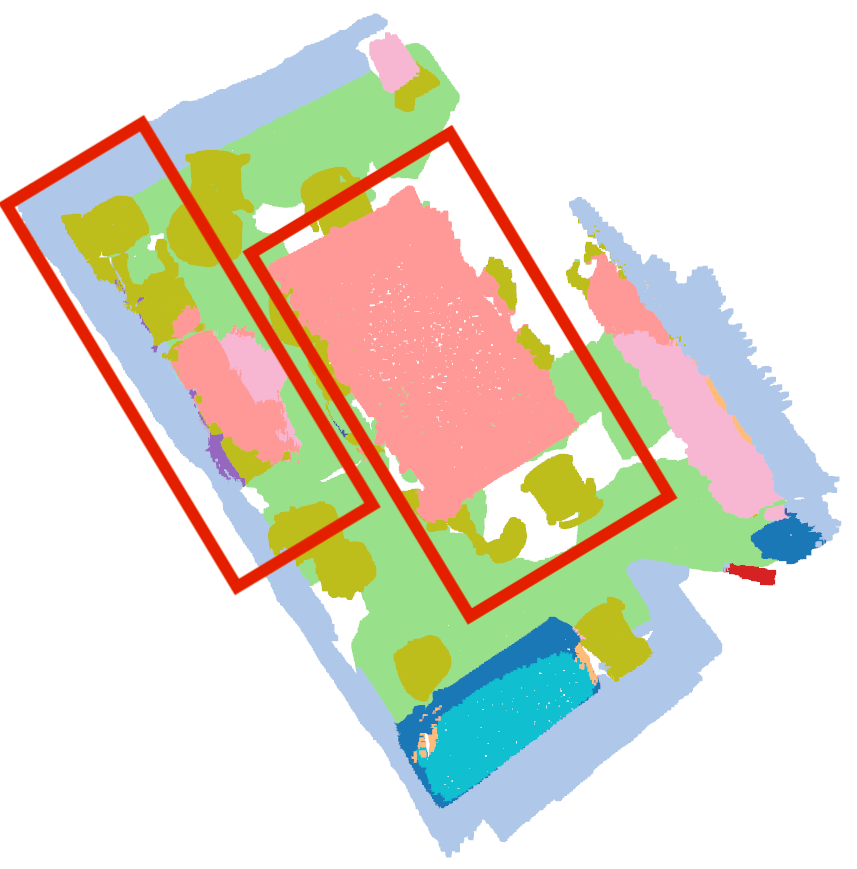}
    &\includegraphics[width=0.19\linewidth]{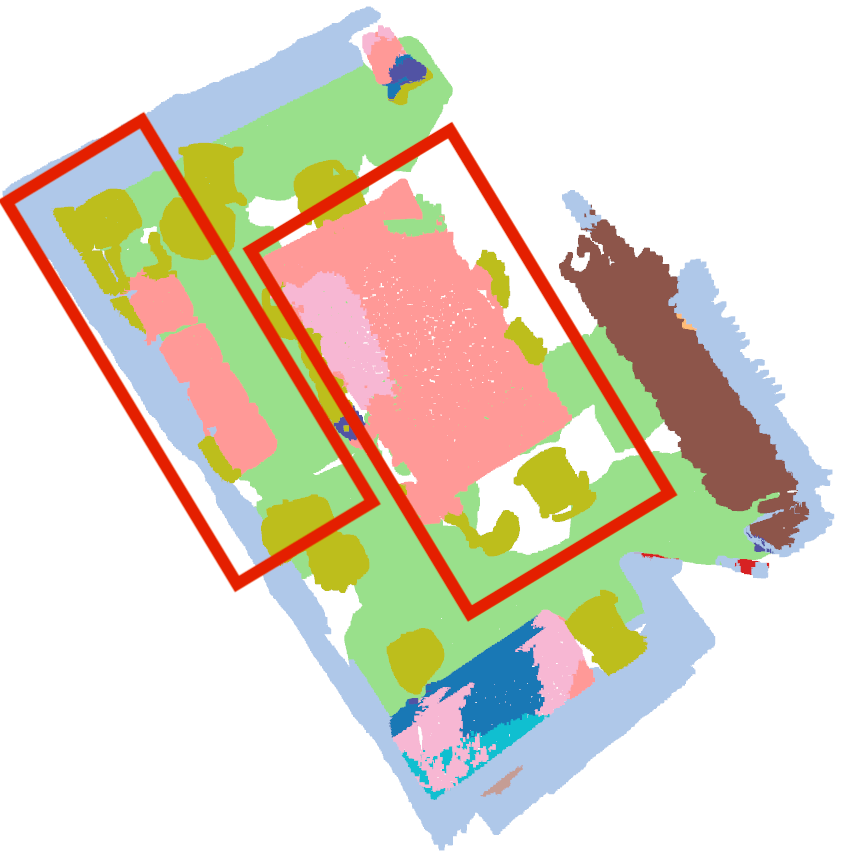}
    &\includegraphics[width=0.19\linewidth]{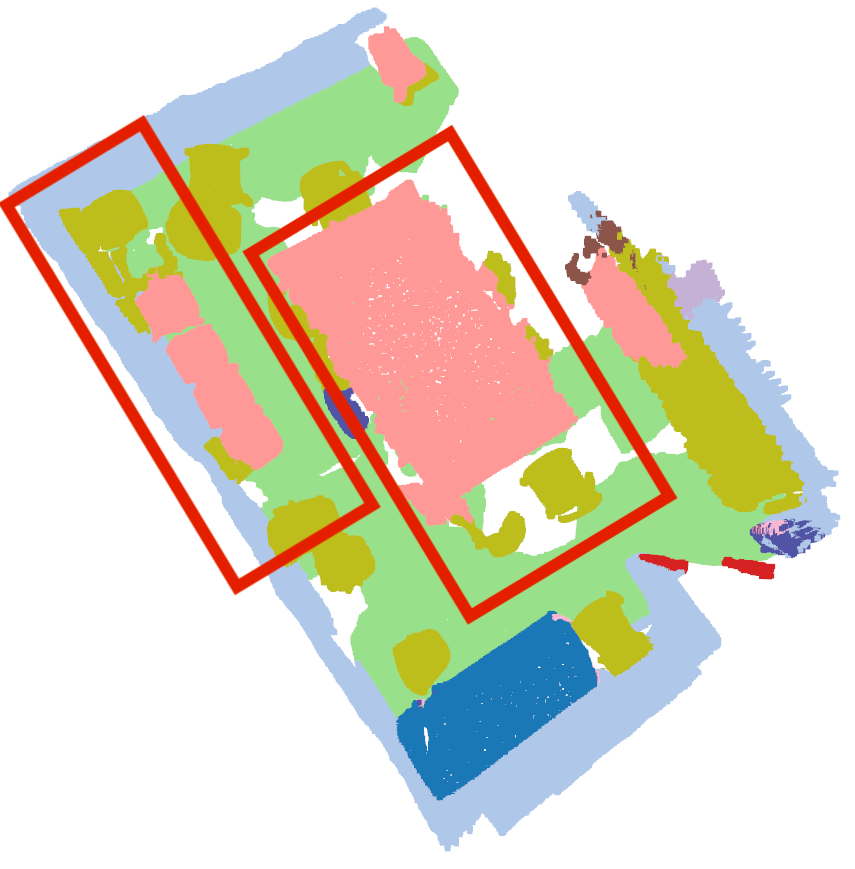} \\
    Input RGB & \multicolumn{4}{c}{Trained from scratch} \\
    \midrule
    \includegraphics[width=0.19\linewidth]{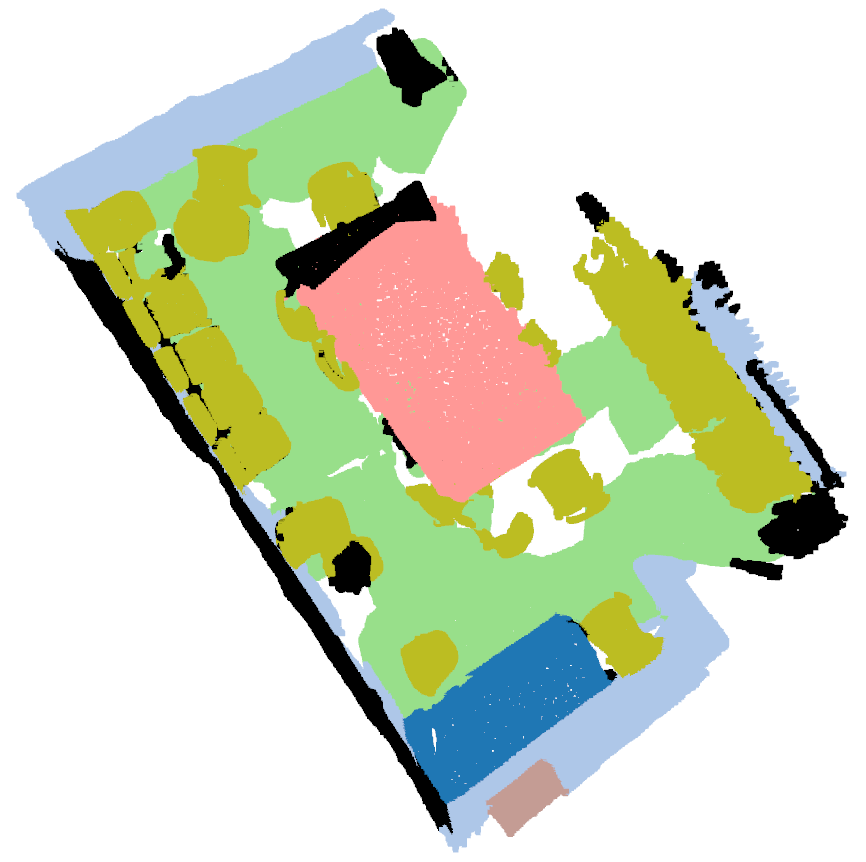}
    & \includegraphics[width=0.19\linewidth]{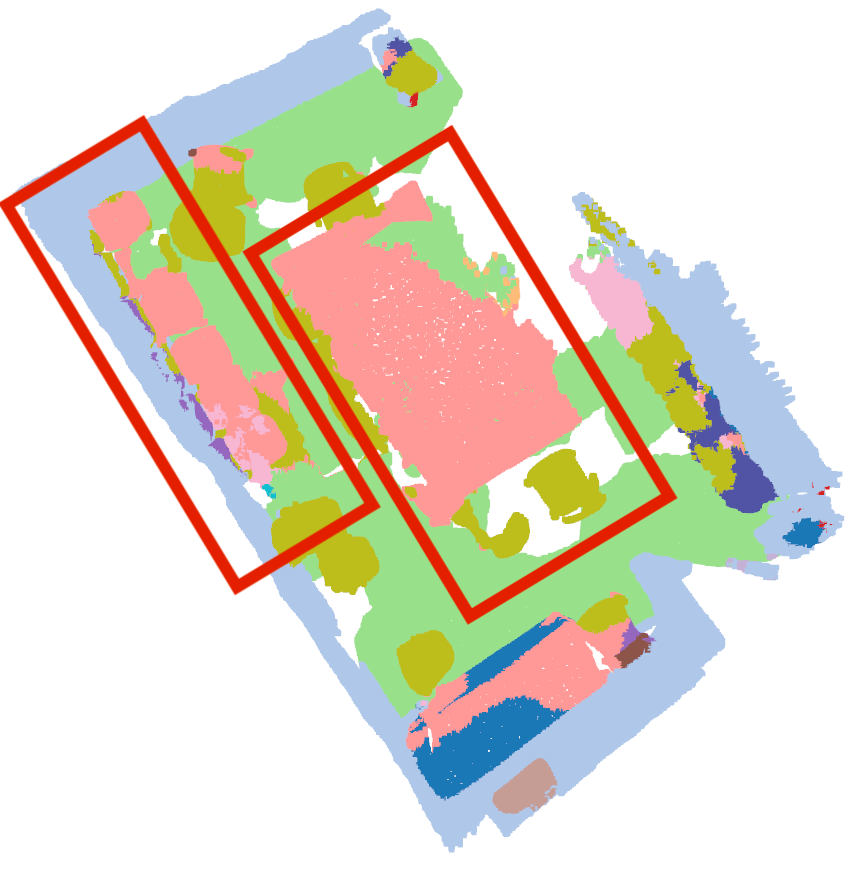}
    & \includegraphics[width=0.19\linewidth]{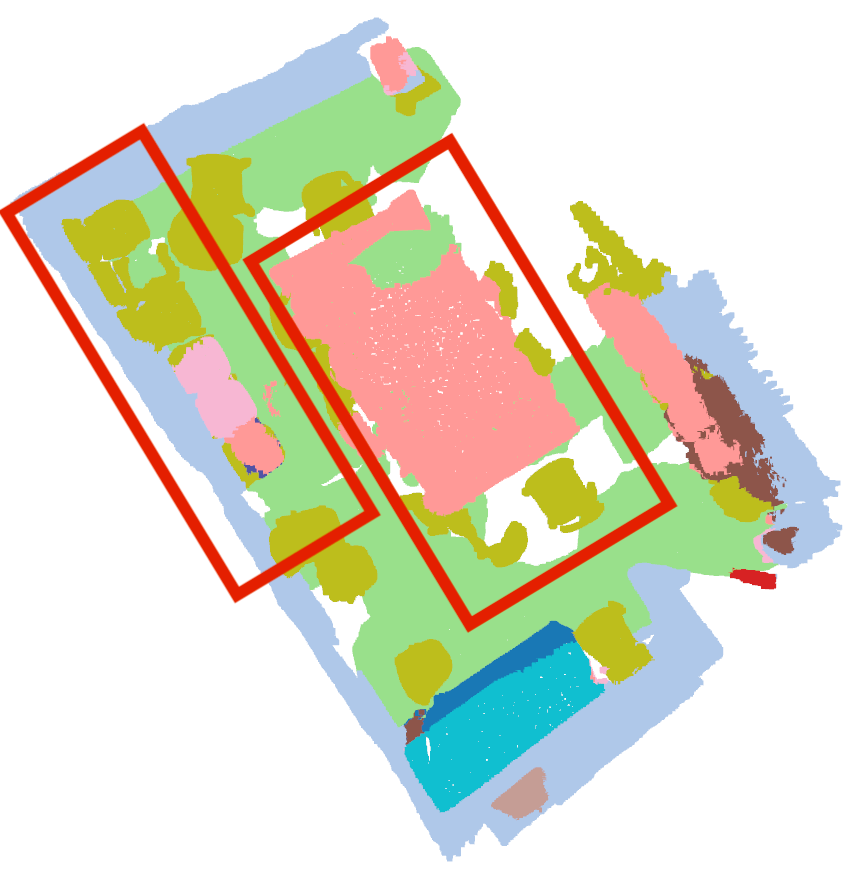}
    & \includegraphics[width=0.19\linewidth]{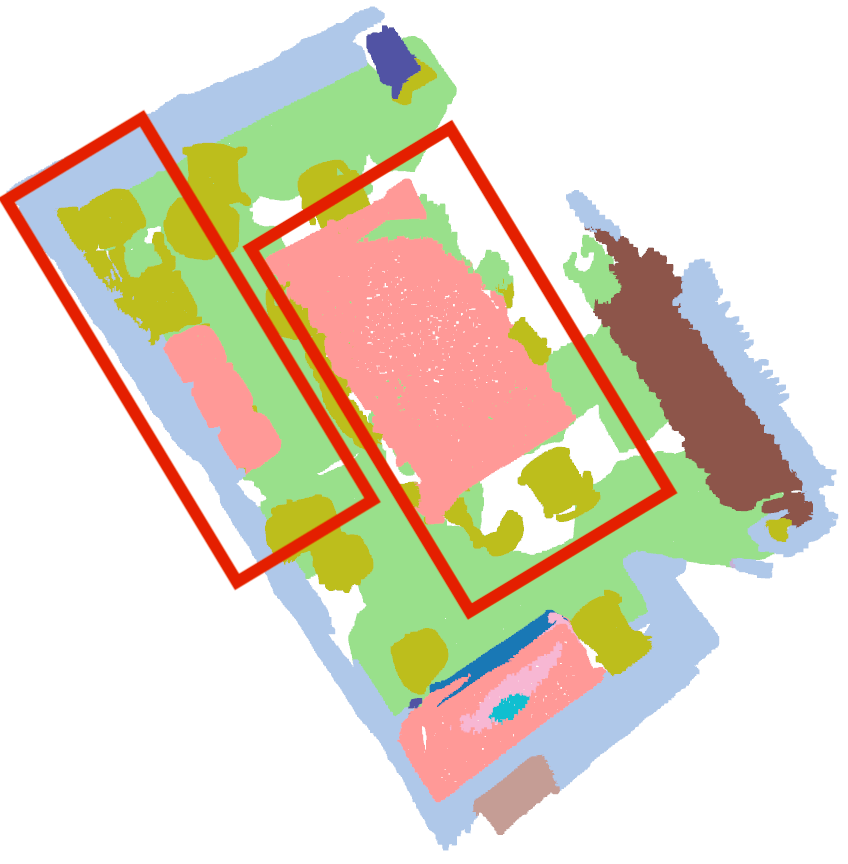} 
    & \includegraphics[width=0.19\linewidth]{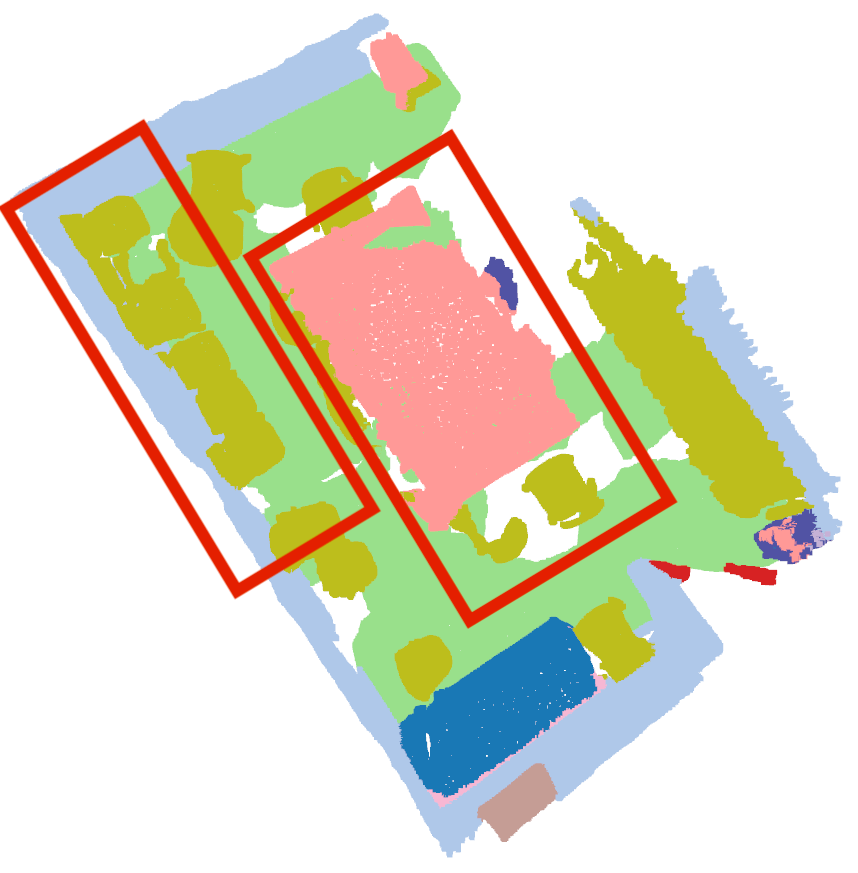}\\
    Ground-truth & \multicolumn{4}{c}{Pre-training} \\
    \midrule\noalign{\vskip .25em}
    \midrule
    \includegraphics[width=0.19\linewidth]{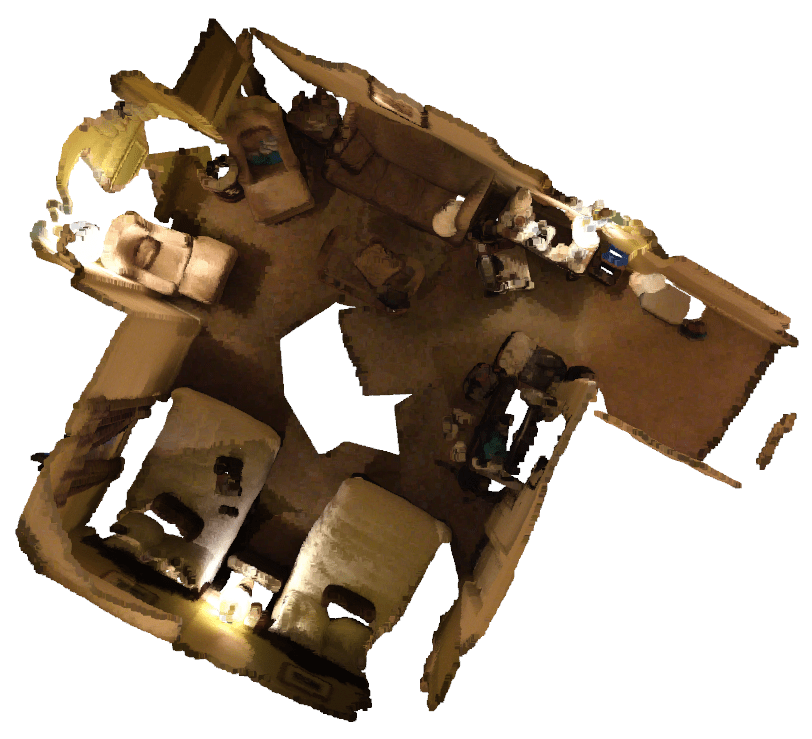}
    &\includegraphics[width=0.19\linewidth]{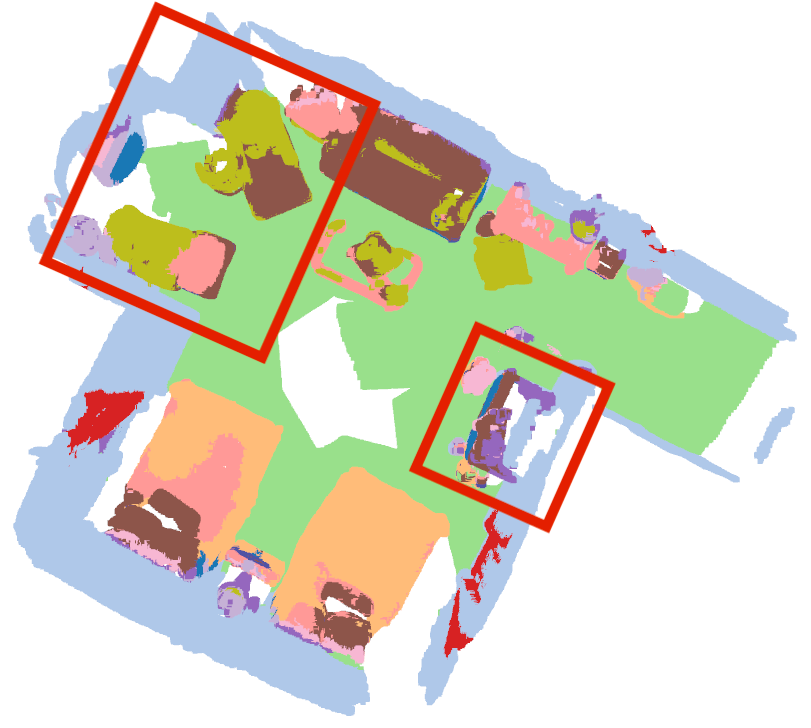}
    &\includegraphics[width=0.19\linewidth]{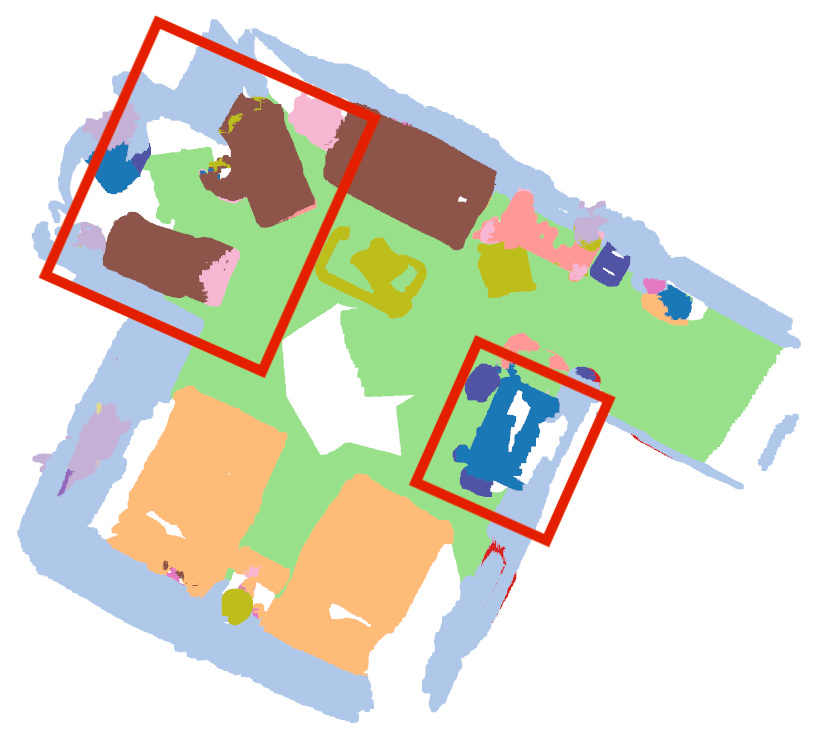}
    &\includegraphics[width=0.19\linewidth]{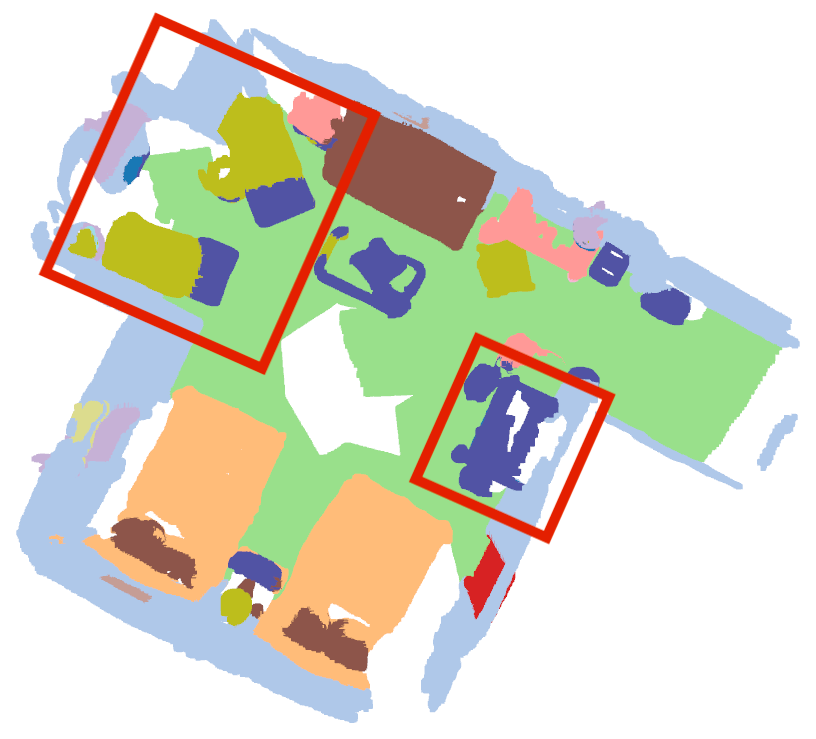}
    &\includegraphics[width=0.19\linewidth]{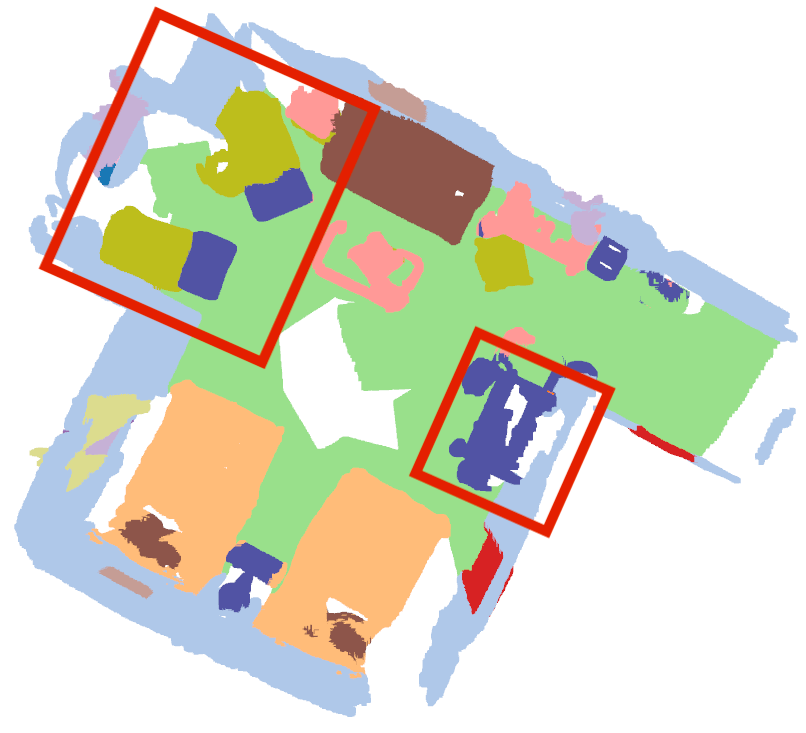} \\
    Input RGB & \multicolumn{4}{c}{Trained from scratch} \\
    \midrule
    \includegraphics[width=0.19\linewidth]{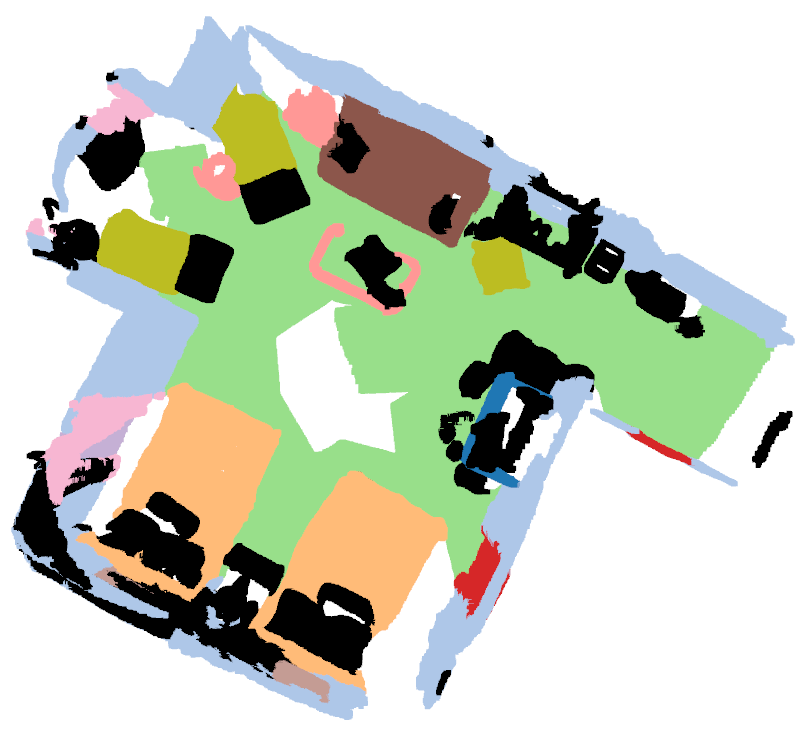}
    & \includegraphics[width=0.19\linewidth]{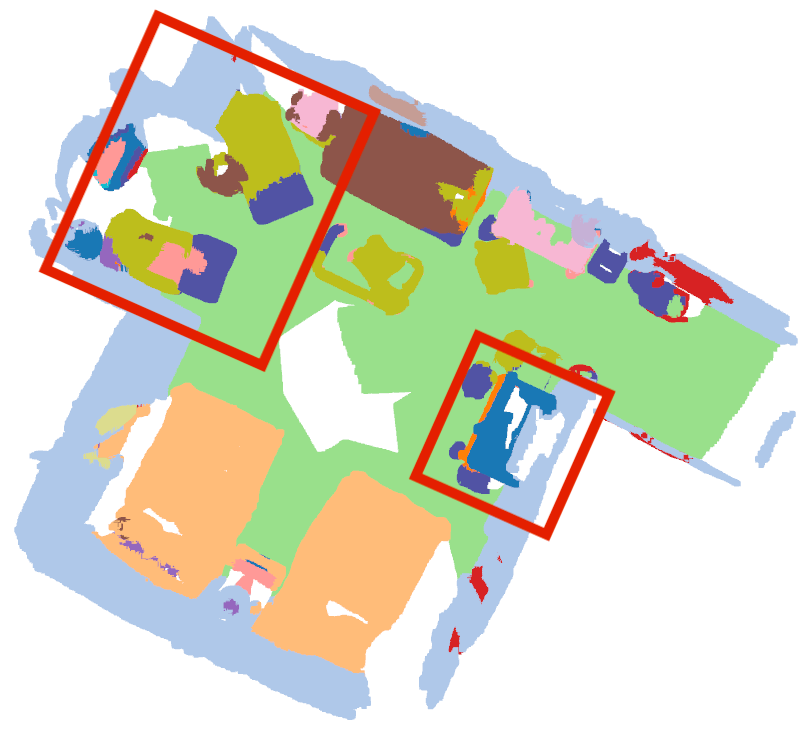}
    & \includegraphics[width=0.19\linewidth]{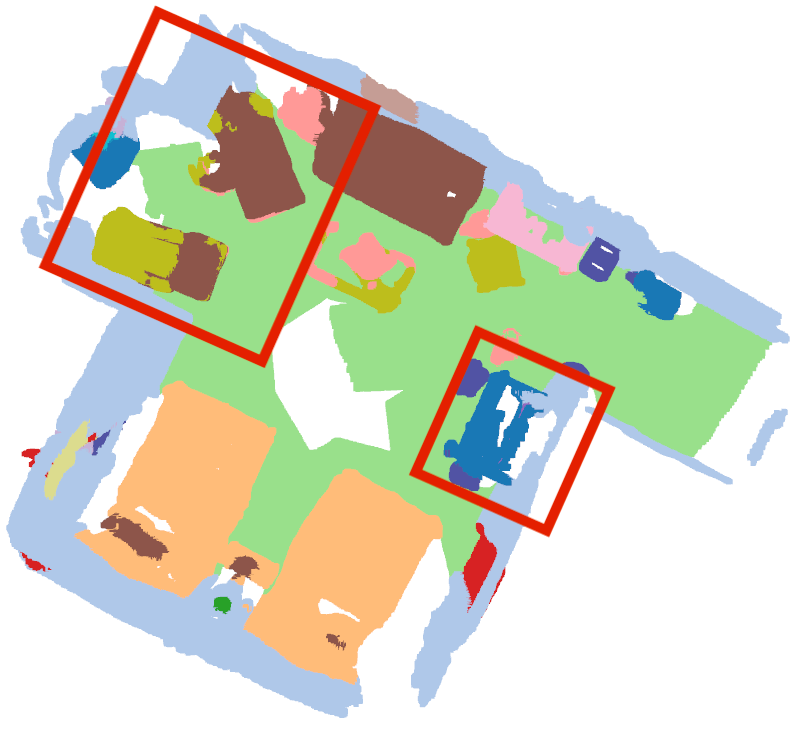}
    & \includegraphics[width=0.19\linewidth]{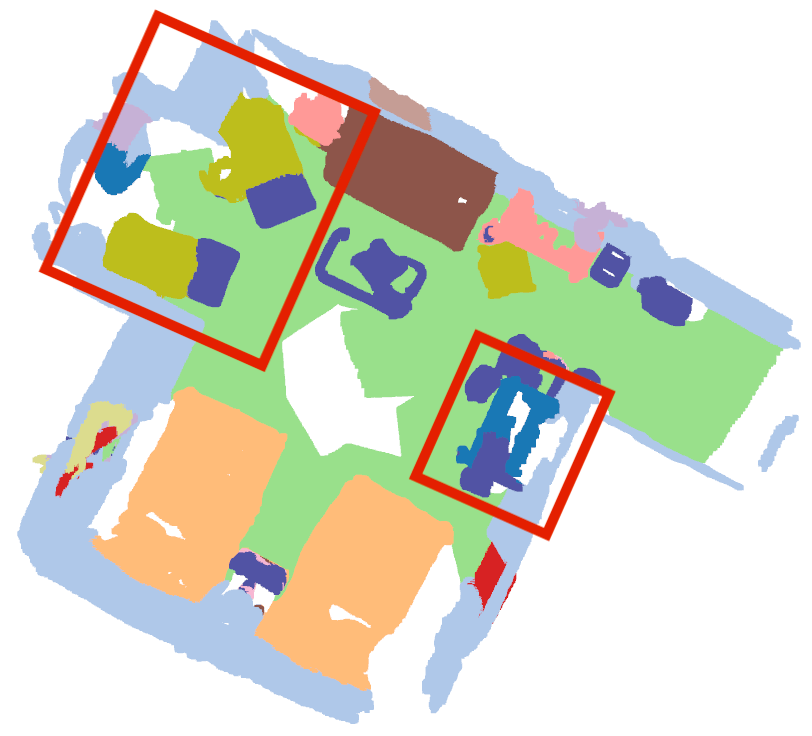} 
    & \includegraphics[width=0.19\linewidth]{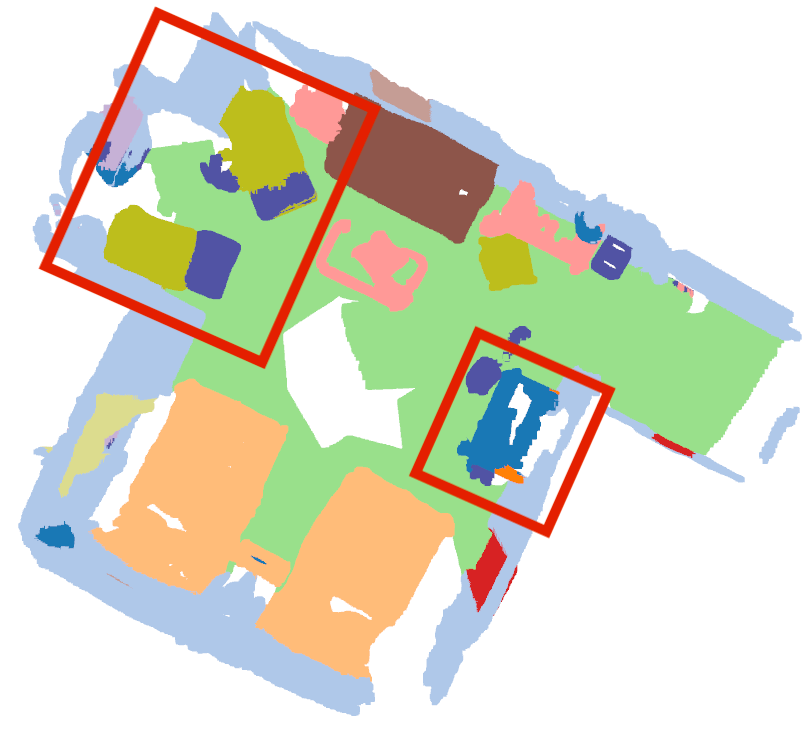}\\
    Ground-truth & \multicolumn{4}{c}{Pre-training} \\
    \midrule\noalign{\vskip .25em}
    \midrule
    \includegraphics[width=0.19\linewidth]{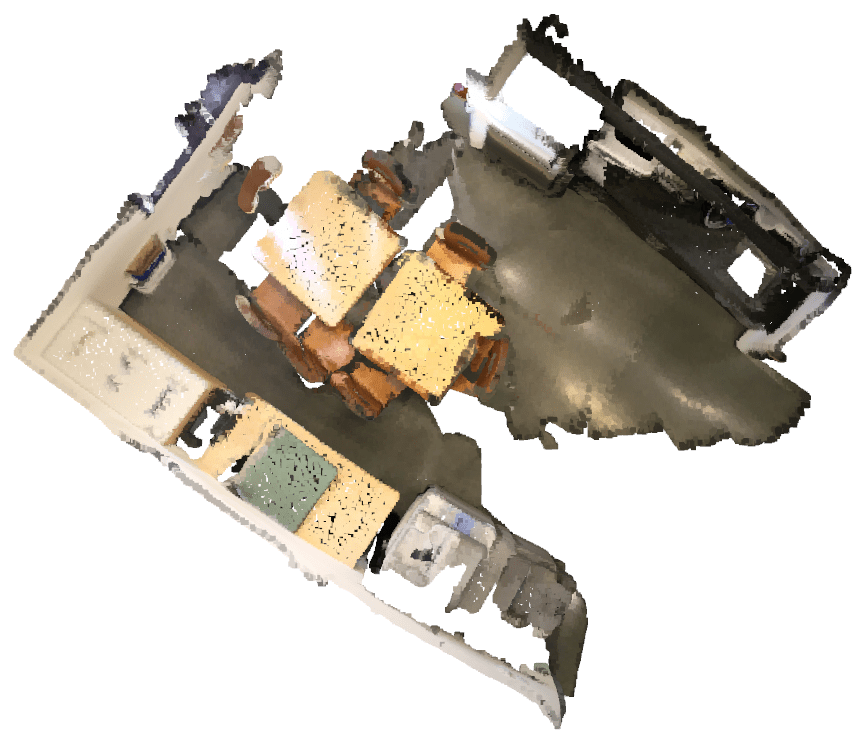}
    &\includegraphics[width=0.19\linewidth]{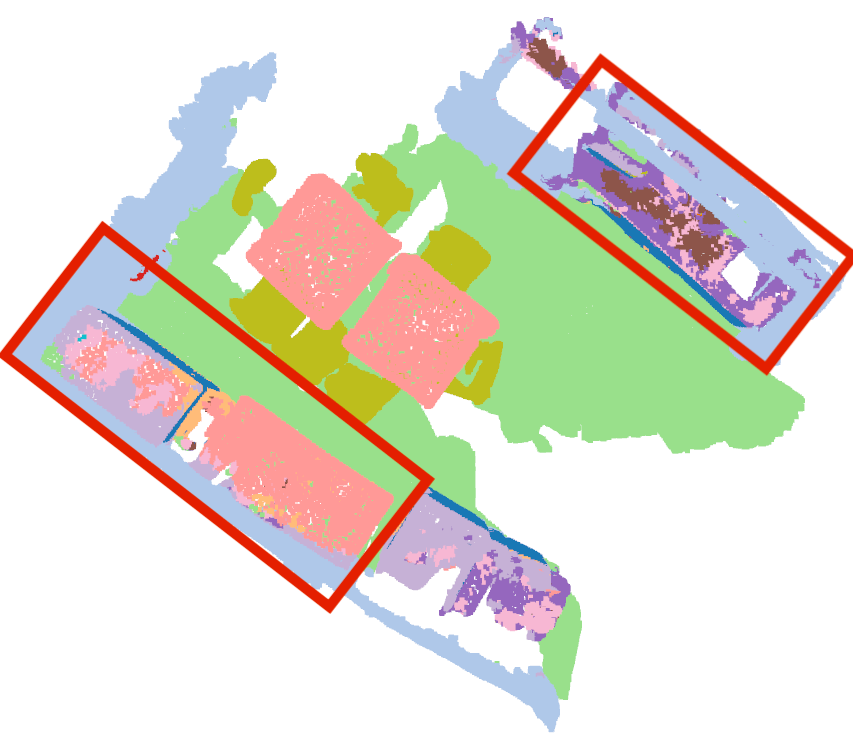}
    &\includegraphics[width=0.19\linewidth]{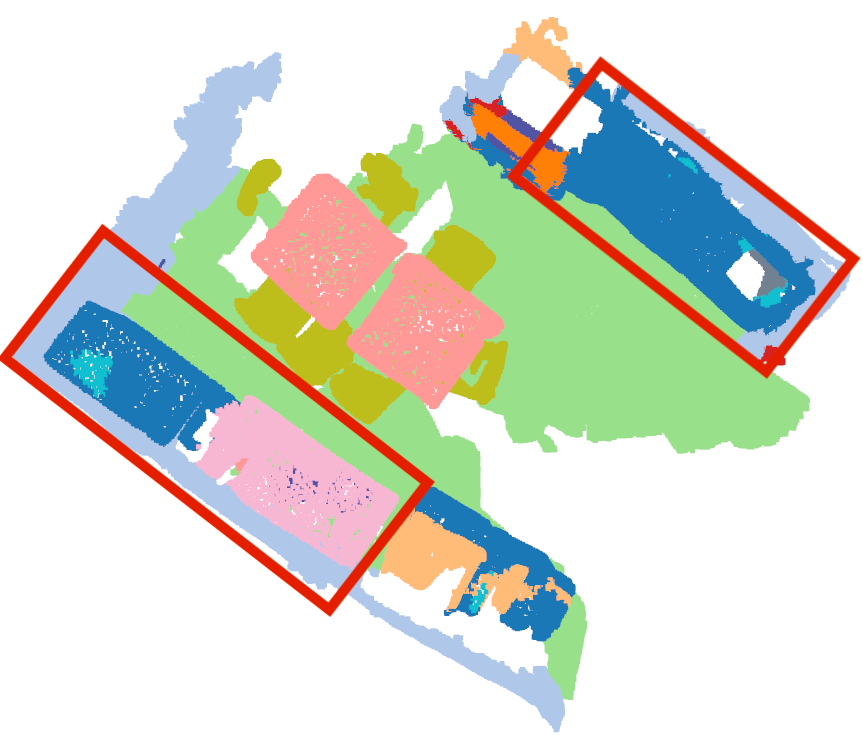}
    &\includegraphics[width=0.19\linewidth]{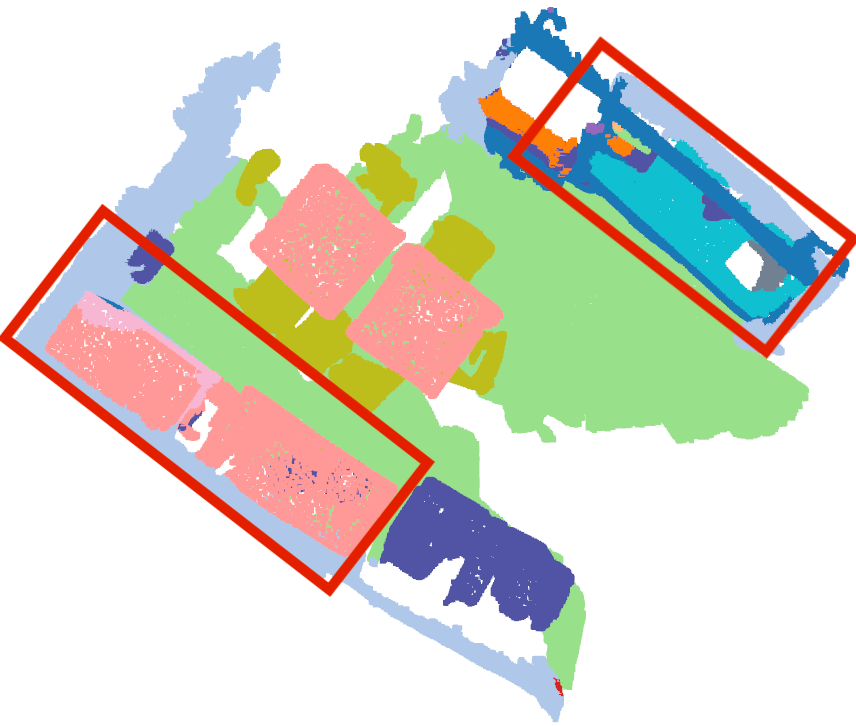}
    &\includegraphics[width=0.19\linewidth]{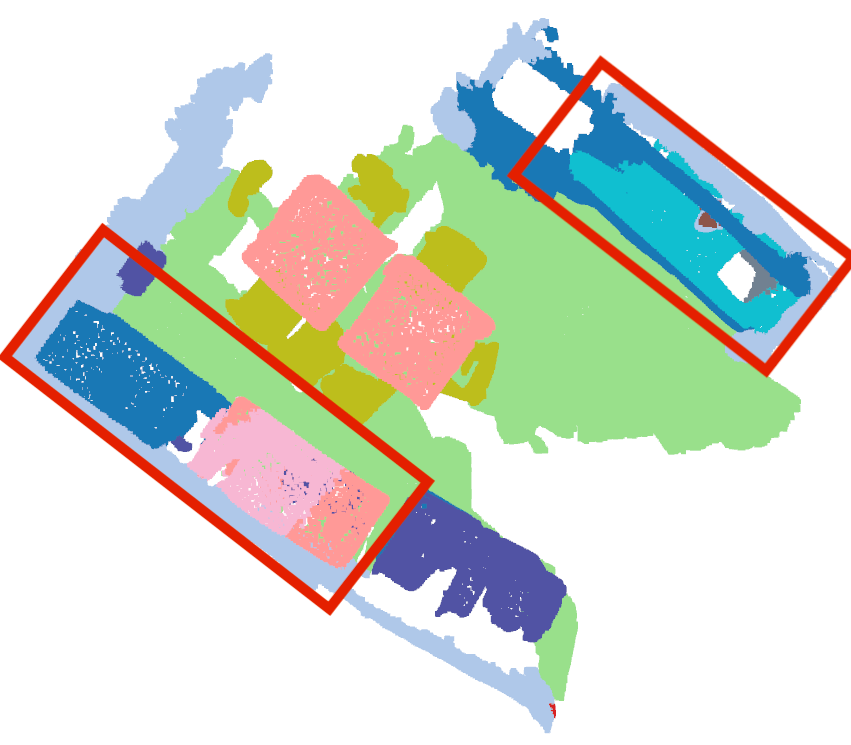} \\
    Input RGB & \multicolumn{4}{c}{Trained from scratch} \\
    \midrule
    \includegraphics[width=0.19\linewidth]{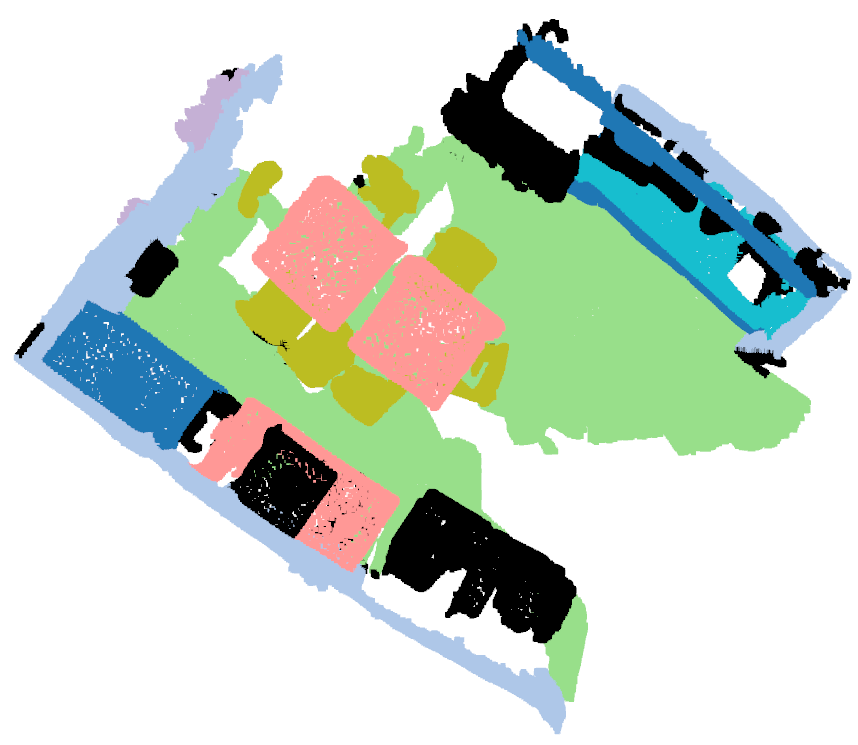}
    & \includegraphics[width=0.19\linewidth]{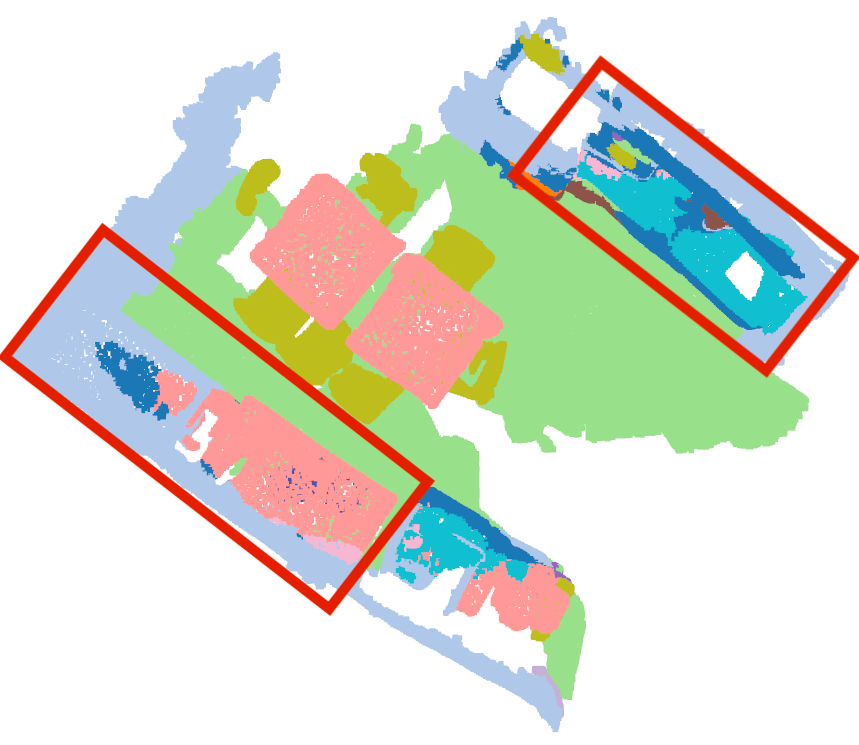}
    & \includegraphics[width=0.19\linewidth]{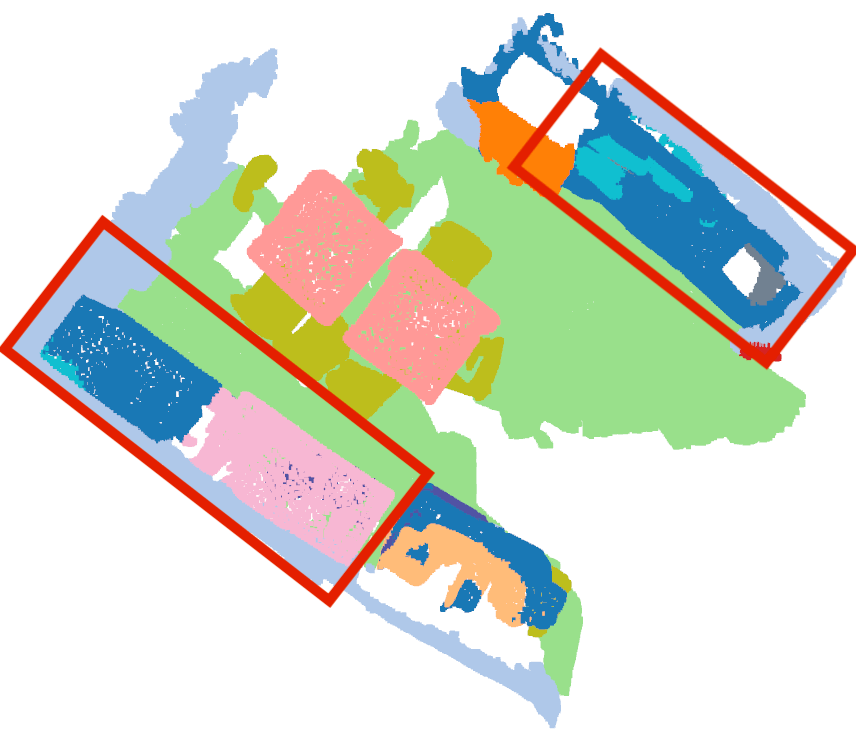}
    & \includegraphics[width=0.19\linewidth]{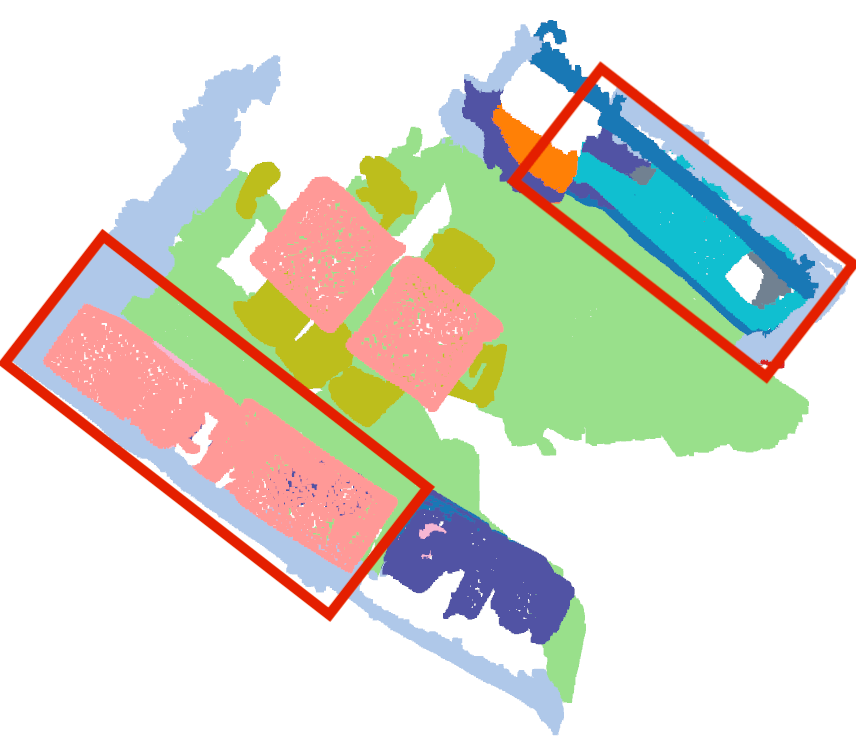} 
    & \includegraphics[width=0.19\linewidth]{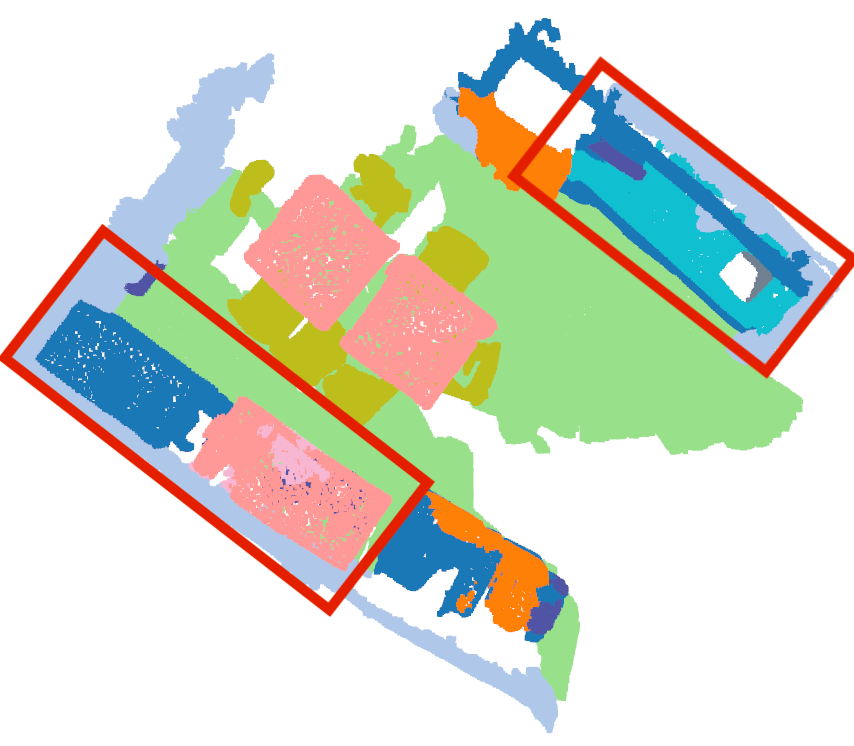}\\
    Ground-truth & \multicolumn{4}{c}{Pre-training} \\
    \bottomrule
  \end{tabular}
  }
\end{figure}

Under limited annotations scenario, \cref{fig:scene1,fig:scene234} are the results of O-CNN~\cite{WangLGST17} models supervised by limited ScanNet annotation. Then \cref{fig:LRscene123} shows the results of O-CNN models supervised under limited ScanNet reconstruction. 
We show the visual comparisons between the results of training from scratch and training with our pre-training.
We use red frames to highlight the difference, where our pre-training lead to more complete shapes and consistent prediction. The visual results echo our quantitative improvement, which shows that our pre-trained models improve the generalizability for scene understanding by the 2D transferred knowledge.  

\begin{figure}
	\centering
	\caption{Corresponding colors to the categories in ScanNet.}
	\includegraphics[width=1\textwidth]{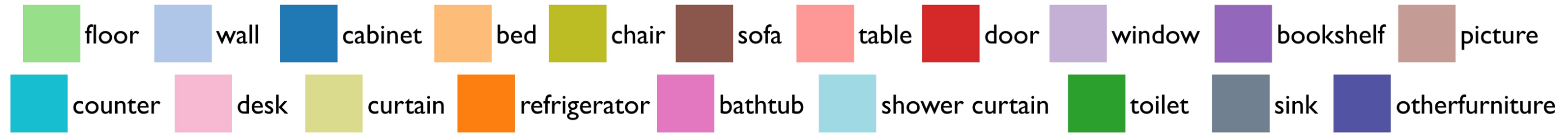}
	\label{fig:color map}
\end{figure}

\clearpage
%
%
\bibliographystyle{splncs04}
\bibliography{egbib}
\end{document}